\documentclass{article}

\usepackage{arxiv}

\usepackage{amsmath}
\usepackage{amsfonts}
\usepackage{graphicx}
\usepackage{enumerate}
\usepackage{url} 
\usepackage{textcomp}
\usepackage{xcolor}
\usepackage{lipsum}
\usepackage{caption}
\usepackage{subcaption}
\usepackage{amsthm}
\usepackage{amsmath}
\usepackage{amssymb}
\usepackage{delarray}
\usepackage{color}
\usepackage{footmisc}
\usepackage{comment}
\usepackage{float}
\usepackage{bm}
\newtheorem{rem}{Remark}

\usepackage{color}
\usepackage{comment}
\usepackage{ulem}
\usepackage{soul}

\title{A deep learning pipeline for cross-sectional and
longitudinal multiview data integration}

\author{ 
	{\hspace{1mm}Sarthak Jain} \\
	Electrical and Computer Engineering \\
	University of Minnesota Twin Cities\\
	Minneapolis, MN 55455 \\
	\texttt{sarthak1996jain17@gmail.com} \\
        \And 
        {\includegraphics[scale=0.06]{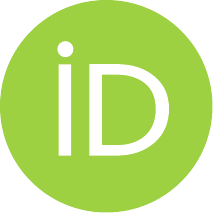}\hspace{1mm}Sandra E. Safo}\thanks{Corresponding Author: Sandra Safo, www.sandraesafo.com} \\
	Division of Biostatistics and Health Data Science\\
	University of Minnesota Twin Cities\\
	Minneapolis, MN 55455 \\
	\texttt{ssafo@umn.edu} \\
}

\begin{document}
\maketitle
\begin{abstract}
Biomedical research now commonly integrates diverse data types or views from the same individuals to better understand the pathobiology of complex diseases, but the challenge lies in meaningfully integrating these diverse views.
Existing methods often require the same type of data  from all views (cross-sectional data only or longitudinal data only) or do not consider any class outcome in the integration method, presenting limitations.
To overcome these limitations, we have developed a pipeline that harnesses the power of statistical and deep learning methods to integrate cross-sectional and longitudinal data from multiple sources. Additionally, it identifies key variables contributing to the association between views and the separation among classes, providing deeper biological insights. This pipeline includes variable selection/ranking using linear and nonlinear methods, feature extraction using functional principal component analysis and Euler characteristics, and joint integration and classification using dense feed-forward networks and recurrent neural networks.
We applied this pipeline to cross-sectional and longitudinal multi-omics data (metagenomics, transcriptomics, and metabolomics) from an inflammatory bowel disease (IBD) study and we identified microbial pathways, metabolites, and genes that discriminate by IBD status, providing information on the etiology of IBD. We conducted simulations  to compare the two feature extraction methods.\\
\textbf{Availability and Implementation:} The proposed pipeline is available from the following GitHub repository:
\url{https://github.com/lasandrall/DeepIDA-GRU}.\\
\textbf{Contact: } ssafo@umn.edu \\
\end{abstract}
\keywords{Data Integration; Multiview Dashboard; Integrative Analysis; Software; Multi-omics; Multi-modal}

\section{Introduction}
Biomedical research now commonly integrates diverse data types (e.g. genomics, metabolomics, clinical) from the same individuals to better understand complex diseases. 
These data types, whether measured at one time point (cross-sectional) or multiple time points (longitudinal), offer diverse snapshots of disease mechanisms. Integrating these complementary data types provides a comprehensive view, leading to meaningful biological insights into disease etiology and heterogeneity.

Inflammatory bowel disase (IBD) including Crohn's disease and ulcerative colitis, for instance, is a complex disease with multiple factors (including clinical, genetics, molecular, and microbial levels) contributing to the heterogeneity of the disease. IBD is an autoimmune disorder associated with the inflammation of the gastrointestinal tract (Crohn's disease) or the inner lining of the large intestine and rectum (Ulcerative Colitis), and results from imbalances and interactions between microbes and the immune system. 
To better understand the etiology of IBD, the IBD  integrated human microbiome project (iHMP) was initiated to investigate potential individual factors contributing to the heterogeneity in IBD{~\cite{IBDMBD}}. 
In that study, individuals with and without IBD from five medical centers were recruited and  followed for one year and the molecular profiles of the host (e.g. host transcriptomics, metabolomics, proteomics) and microbial activities (e.g. metagenomics, metatranscriptomics) were generated and investigated. Several statistical, temporal, dysbiosis and integrative analyses were performed on the multi-omics data. Integrative analyses used included lenient cross-measurement type temporal matching and cross-measurement type interaction testing.

Our work is motivated by the IBD iHMP study, and the many biological research studies that generate both cross-sectional and longitudinal data with an ultimate goal of rigorously integrating these different types of data to investigate individual factors that discriminate between disease groups. Several methods, both linear~\cite{safo2018,safo2018-2} and non-linear~\cite{andrew2013,kan2016,benton2019}, have been proposed in the literature to integrate data from different sources but these methods expect the same types of data (e.g. cross-sectional data only, or longitudinal data only), which limits our ability to apply these methods to our motivating data  that is a mix of cross-sectional and longitudinal data. For instance, methods for associating  two or more views (e.g. {\cite{safo2018,safo2018-2}, iDeepViewLearn~\cite{wang2023interpretable}, JIVE~\cite{Lock2013-ru}, DeepCCA~\cite{andrew2013}, DeepGCCA~\cite{Feng2022}, kernel methods~\cite{safo2023scalable}, co-inertia~\cite{Min2020}}) or for joint association and prediction (e.g. {SIDA~\cite{safo2022}, DeepIDA~\cite{safo2021}, JACA~\cite{zhang_jaca}, MOMA~\cite{Moon2022-jp}, CVR~\cite{luo2016}, randmvlearn\cite{SafoScalable2023}, BIP~\cite{Chekouo2022-du}, sJIVE~\cite{PALZER2022107547}}) are applicable to cross-sectional data only. The   Joint Principal Trend Analysis (JPTA) method proposed in \cite{Zhang2018} for integrating longitudinal data is purely unsupervised, only applicable to two longitudinal data, cannot handle missing data, and assumes equal number of time-points in both the views.  On the other hand, methods for integrating both cross-sectional and longitudinal data are {scarce in the literature. The few existing methods \cite{lee_2019_mildInt, lee_2019_alzheimer} do not maximize association between views and more importantly, when applied to our motivating data, cannot be used to identify variables discriminating between those with and without IBD. Both these methods use recurrent neural networks to extract features from the different modalities and then simply concatenate the extracted features to perform classification.}

To bridge the gap in existing literature, we build a pipeline that 1) integrates longitudinal and cross-sectional data from multiple sources  such that there is maximal separation between different classes (e.g. disease groups) and maximal association between views; and 2) identifies and ranks relevant variables contributing most to the separation of classes and association of views. Our pipeline combines the strengths in statistical methods, such as the ability to make inference, reduce dimension, and extract longitudinal trends, with the flexibility of deep learning, and consists of: 
i) variable selection/ranking; ii) feature extraction and iii) joint integration and classification (Figure \ref{fig:DeepIDA}).

In particular, for variable selection/ranking, we consider the linear methods (linear mixed models [LMM] and JPTA) and the nonlinear method (deep integrative discriminant analysis [DeepIDA] \cite{safo2021}). DeepIDA is a deep learning method for joint association and classification of cross-sectional data from multiple sources. It combines resampling techniques, specifically bootstrap, to rank variables  based on their contributions to classification estimates. Since DeepIDA is applicable to cross-sectional data only,  for longitudinal data, we combine DeepIDA with gated recurrent units (GRUs), a class of recurrent neural networks (RNN), to rank variables. We refer to this method as DeepIDA-GRU-Bootstrapping (DGB). Of note, LMM explores linear relationships between a longitudinal variable and an outcome and focuses on identifying  variables discriminating between two classes; JPTA explores linear relationship between two longitudinal views and focuses on identifying variables that maximally associate the views; and DGB models  nonlinear relationships between classes and two or more longitudinal and cross-sectional data and focuses on simultaneously maximizing within-class separation and between-view associations. For feature extraction, we explore the two methods: Euler characteristics (EC) and functional PCA (FPCA), to extract $1$-dimensional embeddings from each of the ($2$-dimensional) longitudinal views. EC and FPCA inherently focus on different characteristics of longitudinal data while extracting features and in this work, the two are compared and analysed using a simulated dataset.
Finally, for integration and classification, we combine the existing DeepIDA method (without bootstrap) with GRUs, {taking as inputs the selected variables and extracted features from each view. } Since we do not implement variable ranking at this stage,  we refer to this method as DeepIDA-GRU, to distinguish it from DBG which implements bootstrap in DeepIDA with GRUs.  DeepIDA-GRU could be used for jointly integrating a mix of longitudinal and cross-sectional view{s} and discriminating between two or more classes. We emphasize that DeepIDA-GRU combines the existing DeepIDA method (without bootstrap) with GRUs.  As such, DeepIDA-GRU can directly take longitudinal data as input,  making the feature extraction step (which could potentially lead to a loss of information) optional. Please refer to Figure \ref{fig:DeepIDA} for a visual representation of the DeepIDA-GRU framework.

In summary, we provide a pipeline that innovatively combines the strengths of existing statistical and deep learning methods to rigorously integrate cross-sectional and longitudinal data from multiple sources for deeper biological insights. Our pipeline offers four main contributions to the field of integrative analysis. First, our framework allows users to integrate a mix of cross-sectional and longitudinal data, which is appealing and could have broad utility. Second, we allow for the use of a clinical outcome in variable selection or ranking. Third, we model complex nonlinear relationships between the different views using deep learning. Fourth, our framework has the ability to accommodate missing data.


\section{MATERIALS AND METHODS}\label{sec:methods}
\subsection{Datasets and Data Preprocessing}
To evaluate the effectiveness of the proposed pipeline, we use simulations to compare the two feature extraction methods and make recommendations on when each is suitable to use. We applied the pipeline to cross-sectional (host transcriptomics data) and longitudinal (metagenomics and metabolomics) IBD data \cite{IBDMBD} from 90 subjects who had all three measurements. Before preprocessing, the metagenomics data contained path abundances of $22113$ gene pathways, the metabolomics data consisted of $103$ hilic negative factors, and the host transcriptomics data consisted of $55765$ probes. We note that for most of the participants, multiple samples of their host transcriptomics data were collected within a single week. Therefore, in this work, we consider the host transcriptomics view as a cross-sectional view, and the data for each individual were taken as the mean of all samples collected from them. Preprocessing followed established techniques in the literature \cite{Zhao2021_normalization, Maza2016-qo} and consisted of i) keeping variables that have less than $90\%$ zeros (for metagenomics) or $5\%$ zeros (for metabolomics) in all collected samples; (ii) adding a pseudo count of 1 to each data value (this ensures that all entries are nonzero and allows for taking logarithms in the next steps); (iii) normalizing using the `Trimmed Mean of M-values' method~\cite{Maza2016-qo} (for metagenomics); (iv) logarithmic transformation of the data, and (v) plotting the histogram of variances and filtering out variables (pathways) with low variance across all collected samples. After the preprocessing steps, the number of variables remaining for the metagenomics, metabolomics and transcriptomics data was $2261$, $93$, and $9726$, respectively. {More details about data preprocessing are provided in the Supplementary Material.}

\subsection{Notations and overview of proposed pipeline}
\begin{figure*}[t]
\captionsetup{justification=centering}
         \centering
    \includegraphics[width=0.99\textwidth]{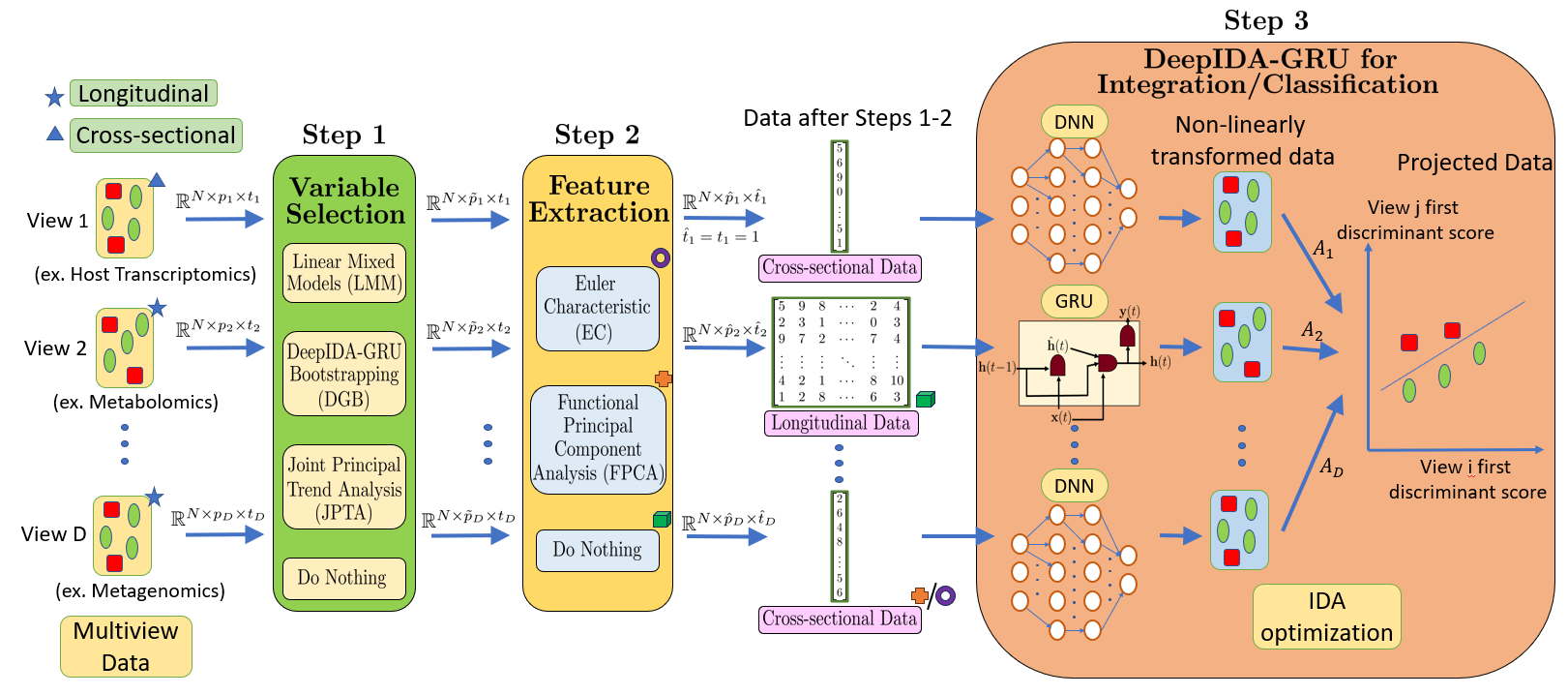}
         \caption{Pipeline for multiview data integration/classification. \textbf{Variable selection}: Every view's data $\mathbf{X}_d \in \mathbb{R}^{N \times p_d \times t_d}$ is first passed through this step to obtain $\widetilde{\mathbf{X}}_d \in \mathbb{R}^{N \times \tilde{p}_d \times t_d}$, where $\tilde{p}_d < p_d$ if either LMM, DGB or JPTA is used for variable selection and $\tilde{p}_d = p_d$ if this step is skipped. LMM focuses on selecting variables maximally separating the classes whereas JPTA focuses on those maximally associating the views. DGB focuses on identifying variables simultaneoulsy maximizing within-class separation and between-view associations. LMM considers each variable separately and lacks context from other views and other variables within the same view. JPTA and DGB leverage between-view and between-variable relationships while selecting variables. JPTA works with two longitudinal views, whereas DGB is capable of taking in any number of views. 
         \textbf{Feature Extraction}: Feature extraction is performed on longitudinal views to convert $\widetilde{\mathbf{X}}_d \in \mathbb{R}^{N \times \tilde{p}_d \times t_d}$ into $\widehat{\mathbf{X}}_d \in \mathbb{R}^{N \times \hat{p}_d \times \hat{t}_d}$, where $\hat{t}_d =1$ if either EC or FPCA is used and $\hat{t}_d = t_d$ if this step is skipped. EC and FPCA convert longitudinal data to one-dimensional form, which is especially important when one want to use the existing integration methods based on cross-sectional views only.  
         EC outperforms FPCA when there are distinct differences in the covariance structure between the classes, while FPCA is better when there are variations in time-trends between the classes. \textbf{Integration/Classification}: DeepIDA-GRU is used for integration/classification, where the longitudinal views are fed into GRUs and the cross-sectional views are fed into dense neural networks. The output of these networks are integrated using Integrative Discriminant Analysis and classified using Nearest Centroid Classifier.}
  \label{fig:DeepIDA}
\end{figure*}
Let $\mathbf{X}_d \in \mathbb{R}^{N \times p_d \times t_d}$ be a tensor representing the 
longitudinal (if $t_d > 1$) or cross-sectional (if $t_d = 1$) data corresponding to the $d$-th view (for $d \in [1:D]$), for the $N$ subjects. The subjects, variables, and time points of view $d \in [1:D]$ are indexed from $[1:N], [1:p_d]$ and $[1:t_d]$, respectively. Here, for each subject $n \in [1:N]$, the data corresponding to the {$d$-th} view {has} $p_d$ variables and each of these $p_d$ variables was measured at $t_d$ time points. Also, let ${\mathbf{X} = \{\mathbf{X}_d: d \in [1:D]\}}$ denote the collection of data from all views. $\mathbf{X}_d^{(n,\rho, \tau)} \in \mathbb{R}$ denotes the value of the variable $\rho \in [1:p_d]$ at time point $\tau \in [1:t_d]$ of the $n$-th subject (for $n \in [1:N]$) in the $d$-th view (for $d \in [1:D]$). Moreover, we use `:' to include all the data of a particular dimension, for example, ${\mathbf{X}_d^{(n,:,:)} \in \mathbb{R}^{p_d \times t_d}}$ denotes the multivariate time series data of the $n$-th subject corresponding to the $d$-th view. Note that there are a total of $K$ classes $\{1,2,\cdots, K\}$ and each subject $n \in [1:N]$ belongs to one of the $K$ classes and the class of the $n$-th subject is denoted by $\kappa(n)$. 
The proposed pipeline for integrating both cross-sectional and longitudinal views is pictorially illustrated in Figure~\ref{fig:DeepIDA} and consists of the following steps: (i) \textbf{Variable Selection or Ranking} is used to find the top variables in each view and eliminate irrelevant variables. In other words, the tensor $\mathbf{X}_d \in \mathbb{R}^{N \times p_d \times t_d}$ is converted to a smaller tensor $\widetilde{\mathbf{X}}_d \in \mathbb{R}^{N \times \tilde{p}_d \times t_d}$ with fewer variables $\tilde{p}_d < p_d$ for all $d \in [1:D]$. In this work, we use LMM, DGB and JPTA for variable selection. We describe these briefly in subsequent sections and in more detail in the Supplementary Material.  The variable selection step is optional and one could go directly to the next step {(in this case $\widetilde{\mathbf{X}}_d = \mathbf{X}_d$)}; (ii) \textbf{ Feature extraction} is used to extract important one-dimensional feature embedding from longitudinal data. This step converts the tensor ${\widetilde{\mathbf{X}}_d \in \mathbb{R}^{N \times \tilde{p}_d \times t_d}}$ to $\widehat{\mathbf{X}}_d \in \mathbb{R}^{N \times \hat{p}_d \times 1}$, where $\hat{p}_d$ is the dimension of the extracted embedding.  The two methods explored in this work for feature extraction are based on Euler curves and FPCA, described briefly in subsequent sections and in more detail in the supplementary material. This step is also optional and one could directly go to the next step {(in this case $\widehat{\mathbf{X}}_d = \widetilde{\mathbf{X}}_d$)}; 
(iii) \textbf{Integration and Classification} uses  DeepIDA-GRU  to simultaneously integrate the multiview data $\{\widehat{\mathbf{X}}_d, d \in [1:D]\}$ obtained after the first two steps and perform classification. We will describe each part of the pipeline in the following {subsections}. 



\subsection{\textbf{Step 1: Variable Selection or Ranking}}
\label{subsec:variableSelec}
Given the high-dimensionality of our data, it is reasonable to assume that some of the variables are simply noise and do not contribute to the distinction between the classes in the views or the correlation between the views. Consequently, it is essential to identify relevant or meaningful variables. We investigated three techniques for selecting variables from cross-sectional and longitudinal data: (i) linear mixed models (LMM), (ii) DeepIDA-GRU bootstrapping (DGB) and (iii) joint principal trend analysis (JPTA). LMM is a univariate method applied to each longitudinal {or cross-sectional} variable and to each view separately. LMM chooses variables that are essential in discriminating between classes in each view. JPTA is a multivariate linear dimension reduction method for integrating two longitudinal views. DGB is a multivariate nonlinear dimension reduction technique that can be used to combine two or more longitudinal and cross-sectional datasets and differentiate between classes. It is useful for choosing variables that are relevant both in discriminating between classes and in associating views. LMM is applicable to any number of longitudinal{ and cross-sectional} data. {Similarly, DGB is also applicable to any number of longitudinal and cross-sectional data.} On the other hand, JPTA can be applied to only two longitudinal views.  It is possible to omit the variable selection step and instead use the entire set of variables in the second step of the pipeline (that is, $\widetilde{\mathbf{X}}_d = {\mathbf{X}}_d$). We briefly describe the three {variable} selection methods. Please refer to the supplementary material for more details. 
\subsubsection{Linear Mixed Models (LMMs)}
LMMs are generalizations of linear models that allow the use of both fixed and random effects  to model dependencies in samples arising from repeated measurements. {LMMs} were used in \cite{lloyd2019} for differential abundance (DA) analysis of longitudinal data from the IBD study to identify important longitudinal variables discriminating between IBD status. To determine if a given variable is important to discriminate between disease groups, we construct the two models: (i) null model and (ii) full model. 
The outcome for each model is the longitudinal variable. The null model associates the outcome with a fixed variable (i.e. time) plus a random intercept, adjusting for covariates of interest (e.g. sites). The full model includes the null model plus the disease status of the sample, treated as a fixed variable. Then, the full and null model are compared using ANOVA to determine statistically significant (p-value $< 0.05$) variables that discriminate between the classes considered. While LMM use the class status in variable selection, it handles each variable separately and does not consider between-views and within-view dependencies. This could lead to a suboptimal variable selection because some variables may only be significant in the presence of other variables. 
\subsubsection{Joint Principal Trend Analysis (JPTA)}
Joint Principal Trend Analysis (JPTA) was introduced in 2018 by~\cite{Zhang2018} as a method to extract shared latent trends and identify important variables from a pair of longitudinal high-dimensional datasets. Following our notation, we let $\{\mathbf{X}_1^{(n,:,:)}: n \in [1:N]\}$ and $\{\mathbf{X}_2^{(n,:,:)}: n \in [1:N]\}$ be the longitudinal datasets for view $1$ and view $2$, respectively, for the $N$ subjects. The number of variables in view $1$ and view $2$ are $p_1$ and $p_2$, respectively, and the number of time points for the two views are $t_1=t_2=T$. Therefore, each subject's data $\mathbf{X}_i^{(n,:,:)}$ (for $i \in \{1,2\}$ and $n \in [1:N]$) is a $p_i \times T$ tensor. In JPTA, the key idea is to represent the data of the two views with the following common principal trends: 
\begin{align}
\mathbf{X}_1^{(n,:,:)} &= \mathbf{u} \mathbf{\Theta} \mathbf{B}' + \mathbf{Z}_1^{(n)}, \nonumber\\
\mathbf{X}_2^{(n,:,:)} &= \mathbf{v} \mathbf{\Theta} \mathbf{B}' + \mathbf{Z}_2^{(n)},\nonumber
\end{align}
for $n \in [1:N]$, where (i) $\mathbf{u}$ and $\mathbf{v}$ are $p_1 \times 1$ and $p_2 \times 1$ vectors of variable loadings, respectively; (ii) $\Theta$ is a $1 \times (T+2)$ vector of cubic spline coefficients; (iii) $\mathbf{B}$ is a cubic spline basis matrix of size  $T \times (T+2)$; and (iv) $\mathbf{Z}_i^{(n)}$ for $i \in \{1,2\}$ are the respective noise vectors. To obtain $(\mathbf{u},\mathbf{\Theta}, \mathbf{v})$, the following loss is minimized
\begin{align}
   & \min_{\mathbf{u},\mathbf{v},\mathbf{\Theta}} \sum_{n=1}^{N} \left(|| \mathbf{X}_1^{(n,:,:)}-\mathbf{u} \mathbf{\Theta} \mathbf{B}' ||_F^2 +|| \mathbf{X}_2^{(n,:,:)}-\mathbf{v} \mathbf{\Theta} \mathbf{B}' ||_F^2\right) \nonumber \\
   & \text{s.t. } \mathbf{\Theta} \mathbf{H} \mathbf{\Theta}' \!\leq\! c, ||\mathbf{u}||_1 \!\leq \! c_1, ||\mathbf{v}||_1 \! \leq \! c_2, ||\mathbf{u}||_2^2 \leq 1, ||\mathbf{v}||_2^2 \leq 1,\nonumber \end{align}
where $||\cdot||_F$ represents the Frobenius norm, $\mathbf{H}$ is a $(T+2)$ by $(T+2)$ matrix given by $\mathbf{H}_{i,j} = \int{B_i^{''}(t) B_j^{''}(t)} dt$ (where $[\mathbf{B}]_{t,m} = B_m(t)$) and the sparsity parameters $c_1$ and $c_2$ control the number of nonzero entries in the vectors $\mathbf{u}$ and $\mathbf{v}$, respectively. In particular, after solving the optimization problem, the variables corresponding to the entries of $\mathbf{u}$ and $\mathbf{v}$ which have high absolute values (the top $c_1$ entries from $\mathbf{u}$ and the top $c_2$ entries from $\mathbf{v}$), are the variables that we select as important, using the JPTA method. Thus, using JPTA, we select the top $c_1$ and $c_2$ {variables} for the two views, respectively, that maximize the association between the views. It is important to note that JPTA has several shortcomings relative to LMM and DGB: (i) it does not take into account information about the class labels while selecting the top variables (which makes it more suitable for data exploration and not regression and classification problems), (ii) it can only be used with two longitudinal data, and (iii) it assumes an equal number of time points for both views. 

\subsubsection{DeepIDA-GRU-Bootstrapping (DGB)} \label{sec:DGB}
DGB is a novel method we propose in this manuscript as an extension to DeepIDA~\cite{safo2021} to the scenario where there are longitudinal data in addition to cross-sectional data. DeepIDA is a multivariate dimension reduction method for learning non-linear projections of different views that simultaneously maximize separation between classes and association between views. To aid in interpretability, the authors proposed a homogeneous ensemble approach via bootstrap to rank variables according to how much they contribute to the association of views and separation of classes. 
In its original form, DeepIDA is applicable only to cross-sectional data, which is limiting. Thus, for longitudinal data,  we integrate gated recurrent units (GRUs) into the DeepIDA framework. Gated Recurrent Units (GRUs)\cite{cho-etal-2014-learning} [Supplementary material], are a class of recurrent neural networks (RNNs) that allow learning long-term dependencies in sequential data and help mitigate the problem of vanishing / exploding gradients in vanilla RNNs {\cite{cho-etal-2014-learning,lstm-0, lstm-1, understanding_LSTM}}. We refer to this modified network as DeepIDA-GRU (which is shown pictorially in Figure~\ref{fig:DeepIDA}). Specifically, each cross-sectional view is fed into a dense neural network, and each longitudinal view is fed into a GRU. The inclusion of GRUs in the DeepIDA framework enables us to extend the bootstrapping idea of~\cite{safo2021} to multiview data consisting of both longitudinal and cross-sectional views. We call this approach for variable selection DeepIDA-GRU-Bootstrapping (DGB). A detailed description of the DeepIDA bootstrap procedure can be found in~\cite{safo2021} but for  completeness sake, we enumerate the main steps as applied to DGB here.  
\begin{itemize}
    \item From the set of $N$ subjects $[1:N]$, randomly sample with replacement $N$ times, to generate each of the $M$ bootstrap sets $\{B_1, B_2, \cdots, B_M\}$. Sets $\{B_1^c, B_2^c, \cdots, B_M^c\}$ are called out-of-bag sets. 
    \item From each view, construct $M$ number of $D$-tuples: ${\{\mathcal{V}_m| m \in [1:M]\}}$ of bootstrapped variables, where each $D$-tuple $\mathcal{V}_m$, consists of $D$ sets, denoted by $\mathcal{V}_m = (V_{1,m}, V_{2,m}, \cdots, V_{D,m})$. Here, the $d$-th set $V_{d,m}$ consists of randomly selected $80$ percent variables from the $d$-th view (where $d \in [1:D]$). This gives us the set of $M$ bootstrapped variable sets: $\{\mathcal{V}_1 = (V_{1,1} , V_{2,1}, \cdots, V_{D,1}), \mathcal{V}_2 = (V_{1,2} , V_{2,2}, \cdots, V_{D,2}), \cdots,\mathcal{V}_M=  (V_{1,M} , V_{2,M}, \cdots, V_{D,M})\}$. 
    \item Pair the bootstrapped subject sets with the bootstrapped variable sets. Let the bootstrapped pairs be given by $(B_1, \mathcal{V}_1),(B_2, \mathcal{V}_2), \allowbreak\cdots,  (B_M, \mathcal{V}_M)$ and the out-of-bag pairs be given by $(B_1^c, \mathcal{V}_1), (B_2^c, \mathcal{V}_2),\allowbreak \cdots, (B_M^c, \mathcal{V}_M)$.
    \item For every variable $v$ in every view, initialize its score as ${S_v=0}$. For each bootstrapped pair $(B_i,\mathcal{V}_i)$ and the out-of-bag pair $(B_i^c,\mathcal{V}_i)$ (where $i \in [1:M]$), 
    \begin{itemize}
        \item First train the DeepIDA-GRU network using bootstrapped pair $(B_i, \mathcal{V}_i)$ and then test the network on the out-of-bag pair $(B_i^c, \mathcal{V}_i)$. This gives us a baseline accuracy for the $i$-th pair and the corresponding model is the baseline model for the $i$-th bootstrapped pair. 
        \item For each variable $u \in \mathcal{V}_i$, randomly permute the value of this variable among the different subjects (while keeping the other variables intact). Test the learned baseline model on the permuted data. If there is a decrease in accuracy (compared to the baseline accuracy), then it means that the variable $u$ was likely important in achieving the baseline accuracy. Therefore, in such a scenario, increase the score of variable $u$ by $1$, that is, $S_u=S_u+1$.
    \end{itemize}
\end{itemize}
The overall importance of any variable $u$ is then computed by 
\begin{equation} \label{eq:eff_prop}
    \text{eff\_prop}(u) \!=\! \frac{S_u}{\text{Total bootstrapped pairs containing } u}.
\end{equation}

Of note the Integrative Discriminant Analysis (IDA) objective enables DGB to select variables that are important in simultaneously separating the classes and associating the views. However, compared to LMM and JPTA, DGB can be computationally expensive. The bootstrapping process, however, is parallelizable, which can significantly improve run time. There exist variants of GRUs that can handle missing data as well~\cite{Che2018-nd} and replacing GRU with such variants would allow DGB to handle missing data.

\subsection{\textbf{Step 2: Feature Extraction}}
\label{subsec:featureExtraction}
The feature extraction methods extract important one-dimensional features from longitudinal data. We investigate two methods: (i) Euler Curves (EC) and (ii) Functional Principal Component Analysis (FPCA) for feature extraction. Note that this step is optional because DeepIDA-GRU can accept longitudinal data directly. 
\subsubsection{Euler Curves}
\begin{figure*}
\captionsetup{justification=centering}
            \centering
            \includegraphics[width = 0.85\textwidth]{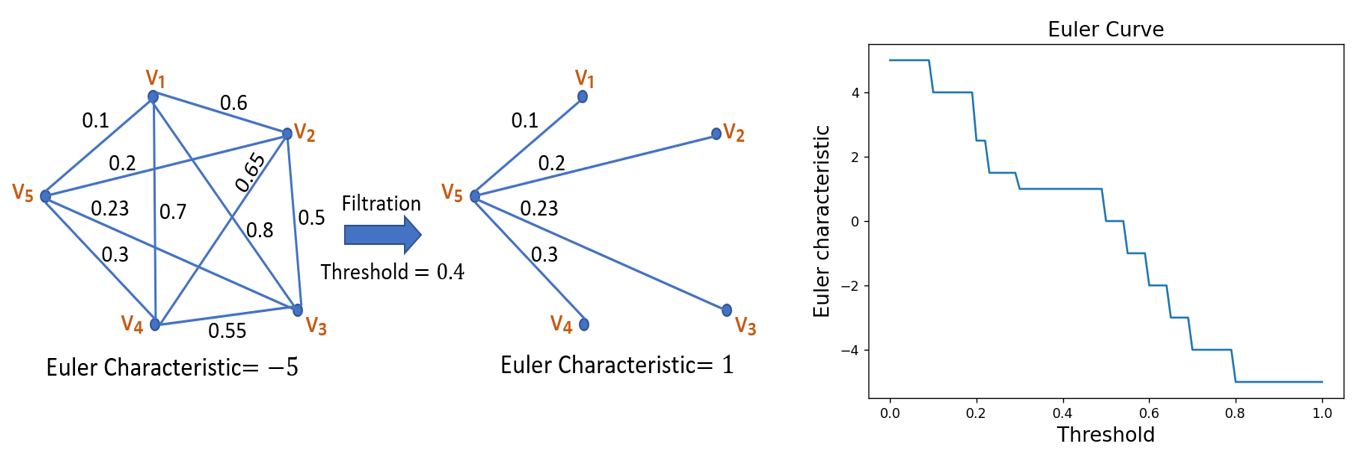}
            \caption{Pictorial representation of the process of constructing an EC curve of a graph: (i) filter the graph using a given threshold and compute the corresponding Euler characteristic, (ii) plot the Euler characteristics for a sequence of increasing thresholds.}
            \vspace{8mm}
            \label{fig:ECProcess}
\end{figure*}

The Euler Characteristic (EC) was first proposed by Euler in 1758 in the context of polyhedra. Recently, Zavala et al.~\cite{zavala2021} explored the potential of EC as a topological descriptor for complex objects such as graphs and images. EC curves, which are low-dimensional descriptors, were created to capture the essential geometrical features of these objects. The construction of EC curves is as follows.
An edge weighted undirected graph $(\mathsf{V},\mathsf{E}, \mathsf{W})$ with $|\mathsf{V}|$ vertices, $|\mathsf{E}|$ edges, and set of weights $\mathsf{W} = \{w(e)| e \in \mathsf{E}\}$, can be represented using a symmetric $|\mathsf{V}|$ by $|\mathsf{V}|$ matrix $M$, where $M_{i,j} = w({e_{i,j}})$ is the weight associated with the edge $e_{i,j}$ between the vertices $v_i$ and $v_j$. For example, the leftmost graph in Figure~\ref{fig:ECProcess} can be represented by the $5$ by $5$ matrix $M$, given by
\begin{equation*}
    M = \begin{bmatrix}
1 & 0.6 & 0.8 & 0.7 & 0.1\\
0.6 & 1 & 0.5 & 0.65 & 0.2 \\
0.8 & 0.5 & 1 & 0.55 & 0.23 \\
0.7 & 0.65 & 0.55 & 1 & 0.3\\
0.1 & 0.2 & 0.23 & 0.3 & 1
\end{bmatrix}.
\end{equation*}
\noindent The Euler characteristic (EC), denoted by $\eta$, of a graph is defined by the difference in the number of vertices and the number of edges:
\begin{equation*}
    \eta = |V|-|E|
\end{equation*}
For instance, the EC of the leftmost graph in Figure~\ref{fig:ECProcess} is {$\eta = 5-10 = -5$}.\par

In order to obtain a low-dimensional descriptor for complex objects (such as graphs, images, matrices, fields, etc.), EC is often combined with a process known as filtration to generate an EC curve, which can be used to quantify the topology of the complex object. Given an edge-weighted graph $(\mathsf{V},\mathsf{E}, \mathsf{W})$ and a threshold $\ell$, the filtered graph for this threshold $\ell$, which we denote by $(\mathsf{V},\mathsf{E}, \mathsf{W})_{\ell}$, is obtained by removing all the edges $e \in \mathsf{E}$ such that $w(e) > \ell$. This filtration step is illustrated in Figure~\ref{fig:ECProcess} for $\ell = 0.4$. For a threshold $\ell$, we denote the EC of the corresponding filtered graph $(\mathsf{V},\mathsf{E}, \mathsf{W})_\ell$ by $\eta_\ell$. Note that for the filtered graph of Fig~\ref{fig:ECProcess}, the EC is given by $\eta_{0.4} = 5-4 = 1$. The EC curve is a plot between $\eta_{\ell}$ and $\ell$ for a series of increasing thresholds $\ell$. The filtration process can be stopped once the threshold is equal to the largest weight of the original graph, at which point the filtered graph is the same as the original graph. The EC curve of the leftmost graph in Figure~\ref{fig:ECProcess} is the rightmost plot in that Figure. It has been demonstrated in~\cite{zavala2021} that EC curves retain important characteristics of the graph and are therefore useful representations of 2D graphs using 1D vectors. \par
To represent a multivariate time series $\widetilde{\mathbf{X}}_d^{(n,:,:)} \in \mathbb{R}^{\tilde{p}_d \times t_d}$ of subject $n \in [1:N]$ using an EC curve, we first find the $\tilde{p}_d$ by $\tilde{p}_d$ precision matrix, or the $\tilde{p}_d$ by $\tilde{p}_d$  correlation matrix or the $\tilde{p}_d$ by $\tilde{p}_d$ covariance matrix from $\widetilde{\mathbf{X}}_d^{(n,:,:)}$ (by treating the multiple time points in the time series as different samples of a given variable), and denote this matrix by $M \in \mathbb{R}^{\tilde{p}_d \times \tilde{p}_d}$. Since $M$ is a symmetric matrix, it represents an edge-weighted graph. The matrix $M$ is then subjected to a sequence of increasing thresholds to obtain an EC curve using the filtration process described above. The resulting EC curve is a $1$D representation of the time series $\widetilde{\mathbf{X}}_d^{(n,:,:)}$ that can then be used as input to the integration and classification step. 
If the number of thresholds used during the filtration process is $x$, then the EC method converts $\widetilde{\mathbf{X}}_d \in \mathbb{R}^{N \times \tilde{p}_d \times t_d}$ to $\widehat{\mathbf{X}}_d \in \mathbb{R}^{N \times x \times 1}$, which is low-dimensional. 


\subsubsection{Functional Principal Component Analysis (FPCA)}
FPCA is a dimension reduction method similar to PCA, which can be used for functional or time-series data. Here, we use FPCA to convert longitudinal data $\widetilde{\mathbf{X}}_d \in \mathbb{R}^{N \times \tilde{p}_d \times t_d}$ into a one-dimensional form $\widehat{\mathbf{X}}_d \in \mathbb{R}^{N \times (xp_d) \times 1}$, by calculating $x$-dimensional scores for each of the $p_d$ variables, where $x$ is the number of functional principal components considered for each variable. Specifically, for any given variable $j$ of view $d$, $\widetilde{\mathbf{X}}_{d}^{:,j,:} \in \mathbb{R}^{N \times t_d}$ is the collection of univariate time series of all $N$ subjects for that variable $j$ (where $d \in [1:D]$ and $j \in [1:\tilde{p}_d]$). FPCA first finds the top $x$ functional principle components (FPCs): $f_1(t), f_2(t), \dots, f_x(t) \in \mathbb{R}^{1 \times t_d}$ of the $N$ time series in $\widetilde{\mathbf{X}}_d^{:,j,:}$. These $x$ FPCs represent the top $x$ principal modes in the $N$ univariate time series and are obtained using basis functions such as B-splines and wavelets. 
Each of the $N$ univariate time series in $\widetilde{\mathbf{X}}_d^{:,j,:}$ is then projected on each of the $x$ FPCs to get an $x$-dimensional score for each subject $n \in [1:N]$, corresponding to this variable. 
The scores of all $\tilde{p}_d$ variables are stacked together to obtain a $\tilde{p}_d x$-dimensional vector for that subject. Thus, with the FPCA method, we convert longitudinal data $\widetilde{\mathbf{X}}_d \in \mathbb{R}^{N \times \tilde{p}_d \times t_d}$ to  cross-sectional  data $\widehat{\mathbf{X}}_d \in \mathbb{R}^{N \times (x p_d) \times 1}$.

{In the Synthetic Analysis of EC and FPCA Section}, we compare EC and FPCA using simulations. We demonstrate that when the covariance structure between the classes differs, the EC curves are particularly better at feature extraction than the functional principal components. However, EC curves are not as effective as FPCA in distinguishing between longitudinal data of different classes when the classes have a similar covariance structure and only differ in their temporal trends.

\subsection{\textbf{Step 3: Integration and Classification}}
\label{sec:Integration and Classification}
In this step, we describe our approach for integrating the output data from any of the first two steps or the original input data, if the first two steps are skipped. 
Denote by $\{\widehat{\mathbf{X}}_d \in \mathbb{R}^{N \times \hat{p}_d \times \hat{t}_d}, d \in [1:D]\}$ the data obtained after the first two steps: selection of variables and extraction of features (where both these steps are optional). 
Data from the $D$ views are integrated using DeepIDA combined with GRUs (i.e., DeepIDA-GRU) as described in the variable selection section. As noted, in the DeepIDA-GRU network, each cross-sectional view is fed into a dense neural network, and each longitudinal view is fed into a GRU. The role of the neural networks and the GRUs is to nonlinearly transform each view. The output of these networks is then entered into the IDA optimization problem (Figure \ref{fig:DeepIDA}). By minimizing the IDA objective, we learn discriminant vectors such that the projection of the non-linearly transformed data onto these vectors result in maximum association between the views and maximum separation between classes.  If the feature extraction step is skipped, then each longitudinal view, $\widehat{\mathbf{X}}_d \in \mathbb{R}^{N \times \hat{p}_d \times \hat{t}_d}$ with $t_d>1$, is fed into its respective GRU in the DeepIDA-GRU network. If the feature extraction step is not skipped, then
each cross-sectional ($\widetilde{\mathbf{X}}_d \in \mathbb{R}^{N \times \tilde{p}_d}$) and longitudinal ($\widehat{\mathbf{X}}_d \in \mathbb{R}^{N \times \hat{p}_d \times \hat{t}_d}$ with $\hat{t}_d=1$) view after the first step is fed into its respective dense neural network in DeepIDA-GRU. DeepIDA-GRU performs integration and classification such that the between-class separation and between-view associations are simultaneously maximized.  Similarly to the DeepIDA network \cite{safo2021}, DeepIDA-GRU also uses the nearest centroid classifier for classification. The classification performance is compared using average accuracy, precision, recall, and F1 scores.

\section{RESULTS}

\subsection{Overview of the pipeline}

We investigate the effectiveness of the proposed pipeline on the longitudinal (metagenomics and metabolomics) and cross-sectional (host transcriptomics) multiview data pertaining to IBD.
The preprocessed host transcriptomics, metagenomics and metabolomics datasets are  represented using $3$-dimensional real-valued tensors of sizes $ \mathbb{R}^{90 \times 9726 \times 1}$, $ \mathbb{R}^{90 \times 2261 \times 10}$ and $ \mathbb{R}^{90 \times 93 \times 10}$, respectively, and passed as inputs in the pipeline.
In the first step, the 
variable selection/ranking methods LMM, DGB and JPTA are used to identify key genes, microbial pathways and metabolites that are relevant in discriminating IBD status and/or associating the views. The top $200$ and $50$ variables from the metagenomics and metabolomics data, respectively, are retained {using} each method. For the host transcriptomics data, LMM and DGB are used to select the top $1000$ genes. Since JPTA is only applicable to longitudinal data, no variable selection is performed on the host transcriptomics data using JPTA. The resulting datasets are then passed through the feature extraction and integration/classification steps. We investigate the performance of our feature extraction, and integration and classification steps {by considering the following three options.} 
{\begin{itemize}
    \item 
\textbf{Method 1 - DeepIDA-GRU with no Feature Extraction: } In this case, there is no feature extraction. 
For integration and classification with DeepIDA-GRU, the cross-sectional host transcriptomics dataset 
is fed into a fully-connected neural network (with $3$ layers and $200,100,20$ neurons in these three layers), while the metagenomics 
and metabolomics data 
are fed into their respective GRUs (both consisting of $2$ layers and $50$ dimensional hidden unit).
 \item  \textbf{Method 2 - DeepIDA-GRU with EC for Metagenomics and Mean for Metabolomics: } In this case, the two longitudinal views are each converted into cross-sectional form. In particular, EC (with $100$ threshold values) is used to reduce the metagenomics data to size $\mathbb{R}^{90 \times 100 \times 1}$. The metabolomics data is reduced to size $\mathbb{R}^{90 \times 50 \times 1}$ by computing mean across the time dimension. Of note, when we visualized the EC curves of the metabolomics data, we did not find any differences between the EC curves for those with and without IBD so we simply used the mean across time. The host transcriptomics data remain unchanged. 
%
The  host transcriptomics, metabolomics and metagenomics data were then each fed into a $3$-layered dense neural networks with structures $[200,100,20]$, $[20,100,20]$ and $[50,100,20]$, respectively, for integration and classification with DeepIDA-GRU (which in this case is equivalent to the traditional DeepIDA network). 
\item 
\textbf{Method 3 - DeepIDA-GRU with FPCA for both the Metabolomics and Metagenomics Views: } 
In this case, FPCA (with $x=3$ FPCs for each variable) is used to reduce the longitudinal metabolomics and metagenomics data into cross-sectional data of sizes $ \mathbb{R}^{90 \times 150 \times 1}$ and $\mathbb{R}^{90 \times 600 \times 1}$, respectively. The host transcriptomics data remain unchanged. 
The host-transcriptomics, metabolomics and metagenomics data 
were each fed into a $3$-layered dense neural networks with structures $[200,100,20]$, $[20,100,20]$ and $[50,100,20]$, respectively, for integration and classification using DeepIDA.
\end{itemize}}
We train and test the $9$ possible combinations of the $3$ variable selection methods: LMM, JPTA and DGB and the $3$ feature extraction plus integration/classification approaches: Method 1, Method 2 and Method 3.
\begin{figure}[t]\captionsetup{justification=centering}
         \centering
    \includegraphics[width=0.7\columnwidth]{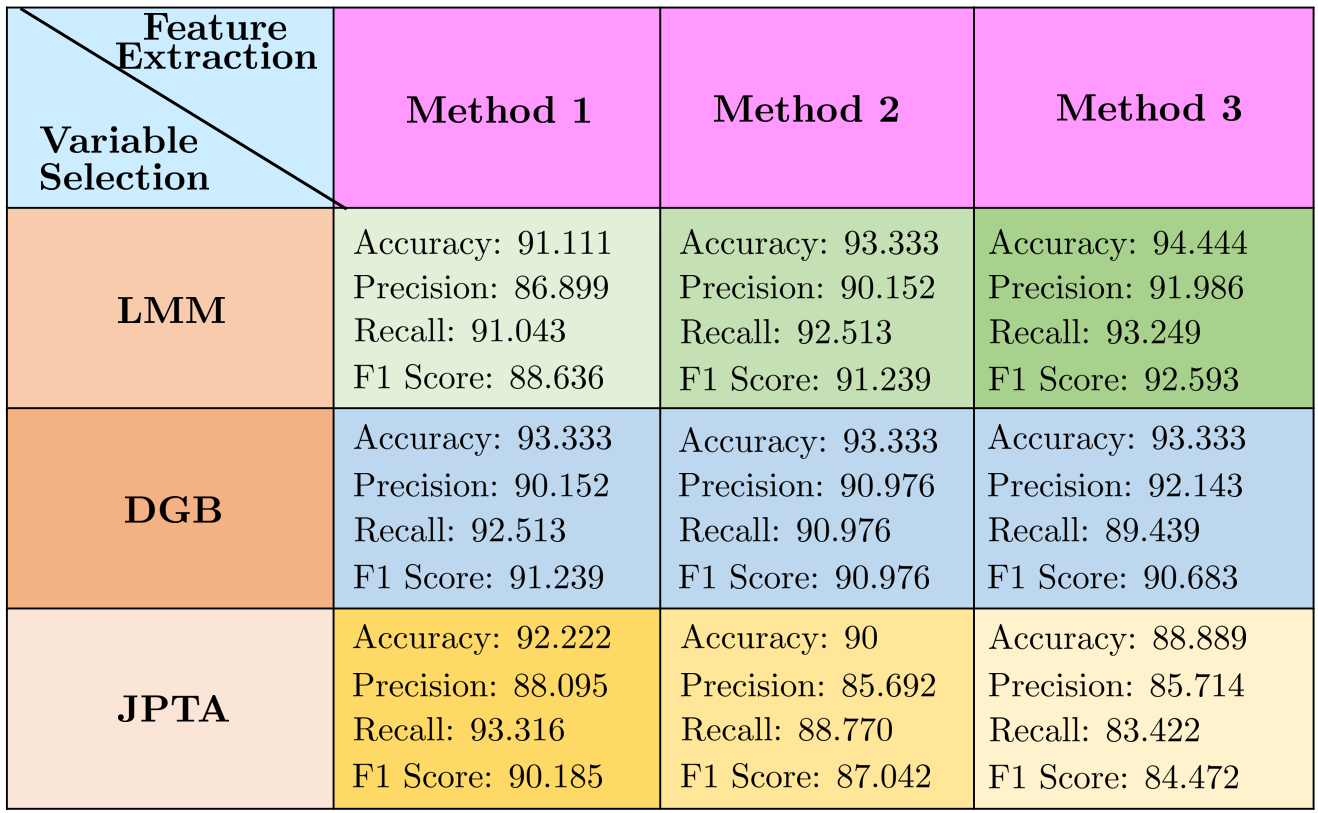}
         \caption{Performance Metrics of nine different combinations of the three feature extraction options: Method 1 (no feature extraction), Method 2 (EC based feature extraction) and Method 3 (FPCA based feature extraction); and the three variable selection methods: LMM, DGB and JPTA. For blocks with the same color, darker shade signifies better overall accuracy.}
  \label{fig:Results}
\end{figure}
\begin{figure} \captionsetup{justification=centering}
         \centering   \includegraphics[width=0.7\columnwidth]{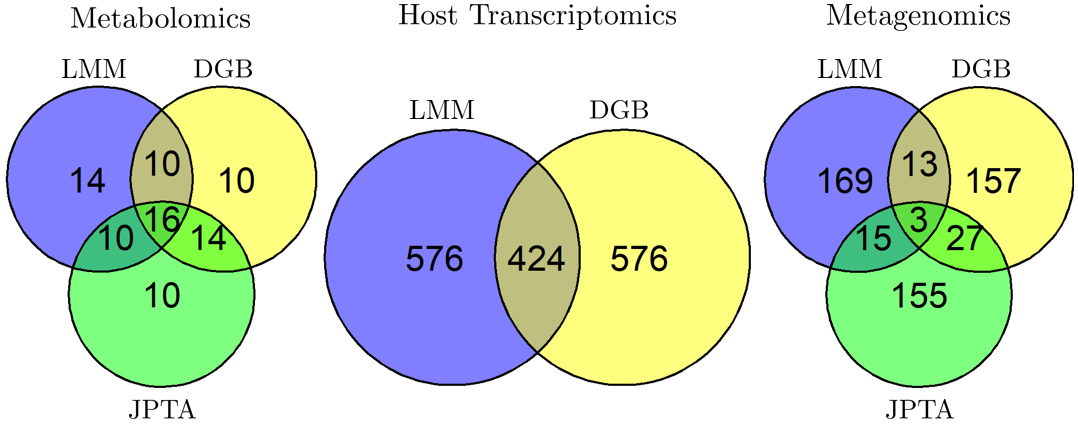}
         \caption{Venn diagrams showing the overlap between the top $1000, 50$ and $200$ variables selected from the host transcriptomics, metabolomics and metagenomics views, respectively, by the three variable selection methods LMM, DGB and JPTA (except JPTA with host transcriptomics view because JPTA does not work with cross-sectional views).}
  \label{fig:venn}
\end{figure}
Owing to the limited sample size, $N$-fold cross-validation is used to evaluate and compare the performance of these $9$ combinations. In particular, the model is trained on $N-1$ subjects (where $N=90$) and tested on the remaining $1$ subject.  This procedure is repeated $N$ times (hence $N$-folds). Average accuracy, macro precision, macro recall and macro F1-scores are the metrics used for comparison. These performance metrics have been summarized in Figure \ref{fig:Results}.
The entire procedure of $N$-fold cross validation is repeated for three arbitrarily selected seeds: $0,10000$ and $50000$. Each of the $9$ blocks in Figure \ref{fig:Results} report the performance of the best model among the three seeds. Note that since LMM and DGB leverage information about the output labels 
    while selecting variables, both these methods only use data from the $N-1$ subjects in the training split of each fold.
    In $N=90$ folds, since there are $90$ different train-test splits, LMM and DGB methods are repeated $90$ times (once for each fold). 
    Unlike LMM and DGB, JPTA does not use the output labels
    during variable selection, and hence
    it is run once on the entire dataset.

\subsection{Classification performance of the  proposed pipeline on the IBD longitudinal and cross-sectional data}
Examining the rows in Figure \ref{fig:Results}, we observe that the classification results based on variables selected by JPTA are  slightly worse than LMM and DGB. The lower performance of JPTA could be because (i) JPTA is a purely unsupervised method and does not account for class membership in variable selection and is therefore not as effective for classification tasks as LMM or DGB; (ii) no variable selection was performed on the host transcriptomics data since JPTA is applicable to longitudinal data only. The classification results based on variables selected by LMM and DGB are comparable with small variations that depended on which feature extraction method was used before integration and classification. The classification results of the feature extraction method FPCA (Method 3) applied on variables selected by LMM  are slightly  better than the feature extraction method EC (Method 2) and the direct DeepIDA-GRU application with no feature extraction (Method 1). Meanwhile, the feature extraction methods EC and FPCA, and DeepIDA-GRU (no feature extraction) yield comparable classification results when applied to  variables selected by DGB. Examining the columns in Figure \ref{fig:Results}, it is evident that the three methods (Method $1$, Method $2$ and Method $3$) have comparable results on the IBD application. The direct DeepIDA-GRU based approach (Method 1) performs best with DGB; the EC (Method 2) and the FPCA (Method 3) approaches work best with LMM and DGB.

\subsection{Variables Identified by LMM, JPTA, and DGB} \label{sec:Top5}
We compare and analyse the top variables selected by LMM, JPTA and DGB. As discussed earlier, both LMM and DGB are performed $90$ times (once for each fold). Each method generates $90$ distinct sets of selected variables. For LMM, the variables in each set are ranked according to their corresponding p-values, whereas, for DGB, the variables in each set are ranked according to their eff\_prop  scores (equation~\eqref{eq:eff_prop}). For LMM, an overall rank/score is associated to every variable using Fisher's approach for combining p-values (Supplementary Material). 
Note that if the Fisher combined p-value is equal to zero for multiple variables, these variables will be assigned the same score.
For the DGB method, the average eff\_prop value is computed to combine the $90$ scores of each variable. Lastly, JPTA is performed once and we choose variables with nonzero coefficients.

\begin{figure*}[ht]
\captionsetup{justification=centering}
     \centering
     \begin{minipage}[t]{.48\textwidth}
     \centering
     \begin{subfigure}[p]{0.73\textwidth}
         \centering
         \includegraphics[width=\textwidth]{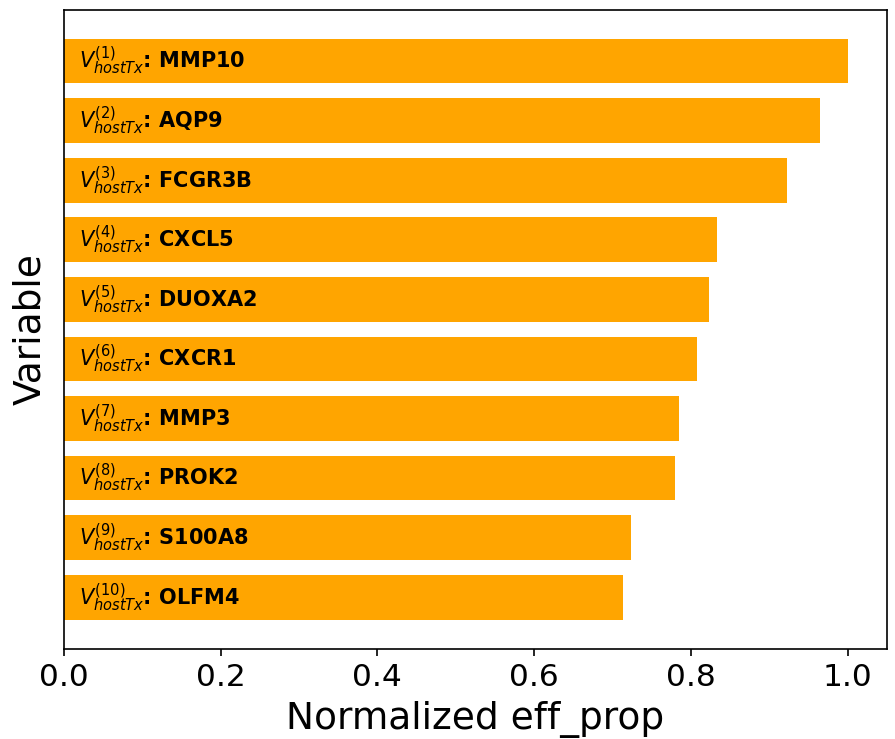}
         \caption{}       
         \label{fig:DGBtop10_hosttx}
     \end{subfigure}
     \vfil
     \vspace{0.9mm}
     \begin{subfigure}[p]{0.73\textwidth}
         \centering
         \includegraphics[width=\textwidth]{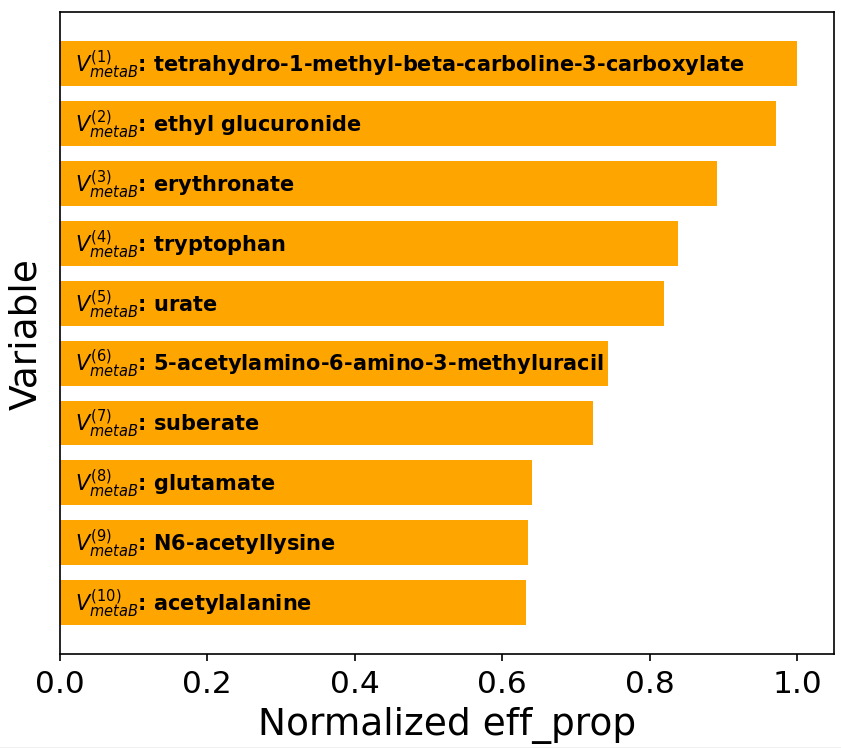}
         \caption{}
         \label{fig:DGBtop10_metabolom}
     \end{subfigure}
     \vfil
     \vspace{0.9mm}
     \begin{subfigure}[p]{0.73\columnwidth}
         \centering
         \includegraphics[width=\textwidth]{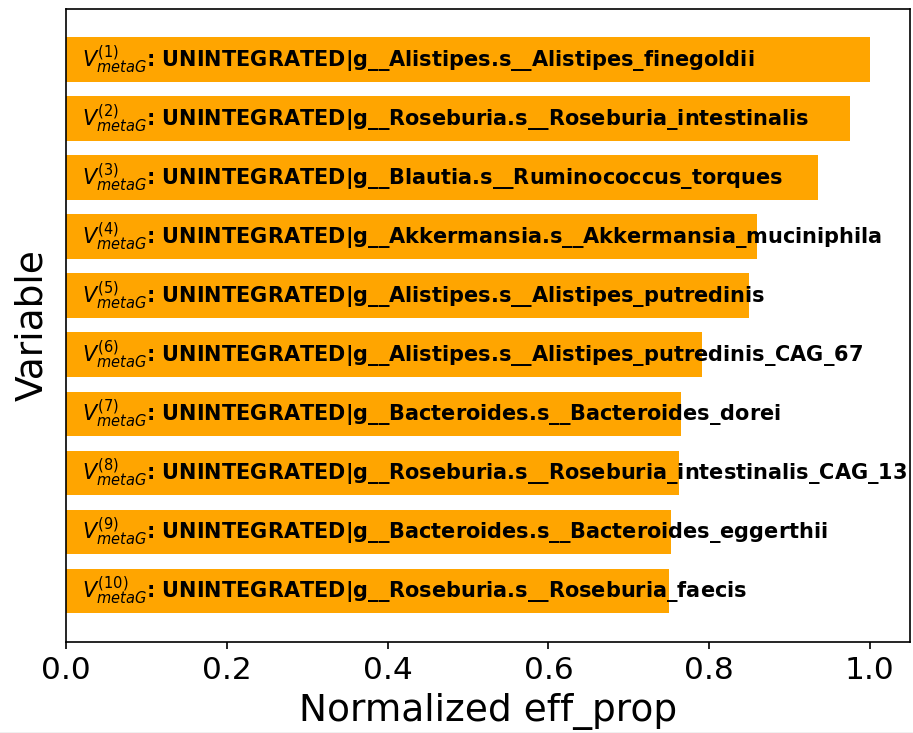}
         \caption{}
         \label{fig:DGBtop10_metagenom}
     \end{subfigure}
     \end{minipage}
     \hfil
     \begin{minipage}[t]{.5\textwidth}
     \centering
     \begin{subfigure}[p]{0.98\columnwidth}
         \centering
         \includegraphics[width=\textwidth]{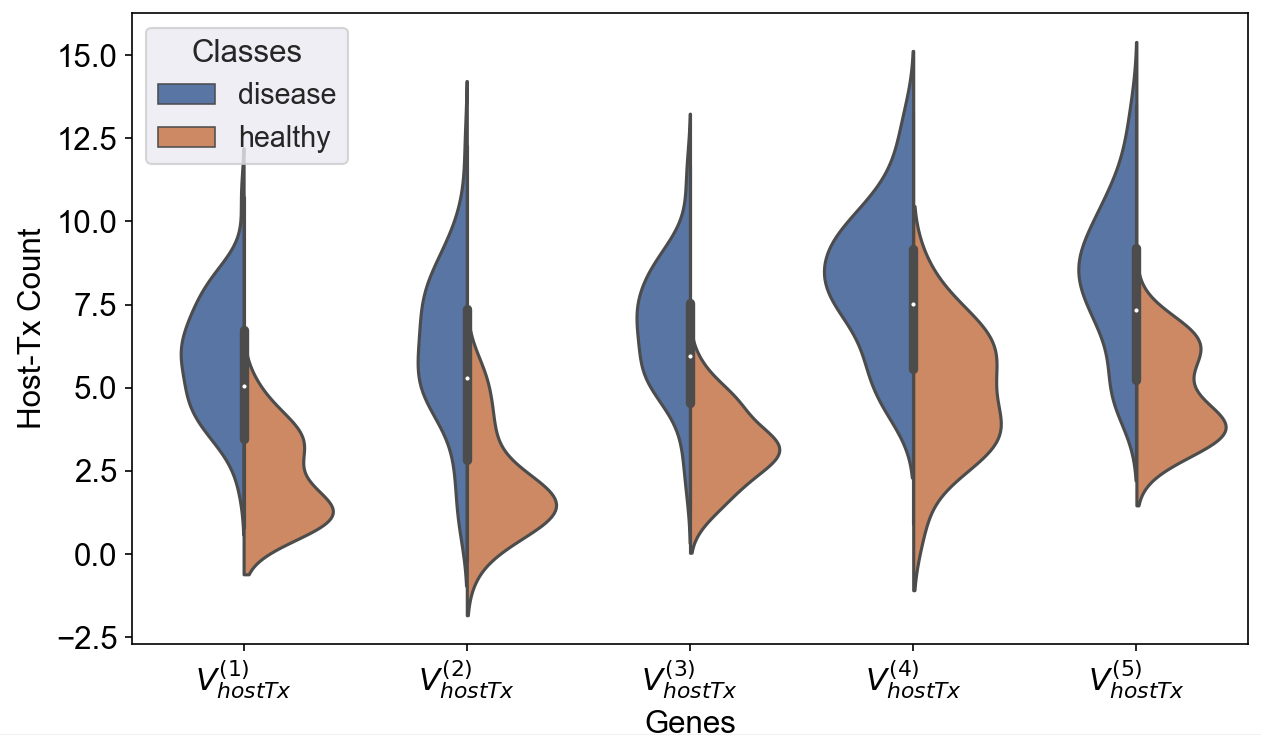}
         \caption{}
         \label{fig:DGBtop5_hosttx}
     \end{subfigure}
     \vfil

     \begin{subfigure}[p]{0.96\columnwidth}
         \centering
         \includegraphics[width=\textwidth]{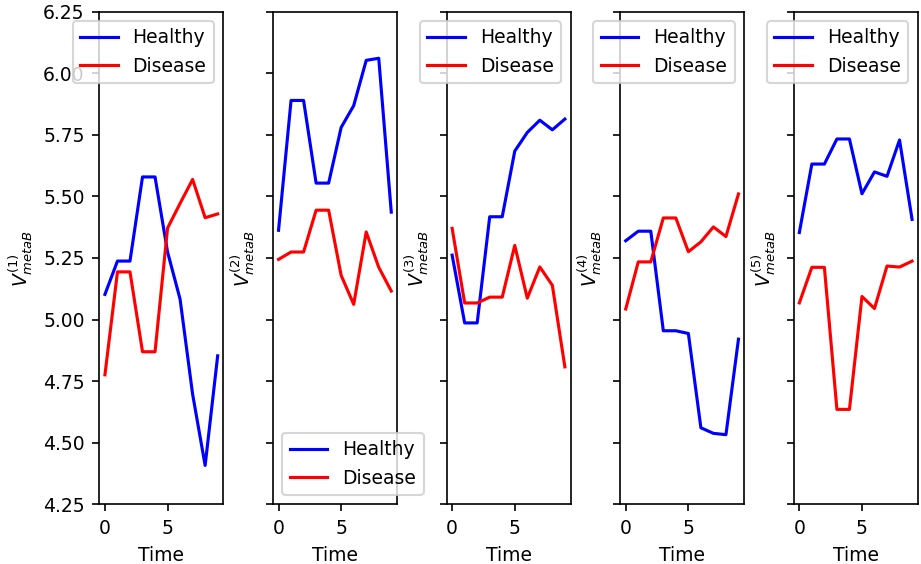}
         \caption{}
         \label{fig:DGBtop5_metabolom}
     \end{subfigure}
     \vfil

     \begin{subfigure}[p]{1\columnwidth}
         \centering
         \includegraphics[width=\textwidth]{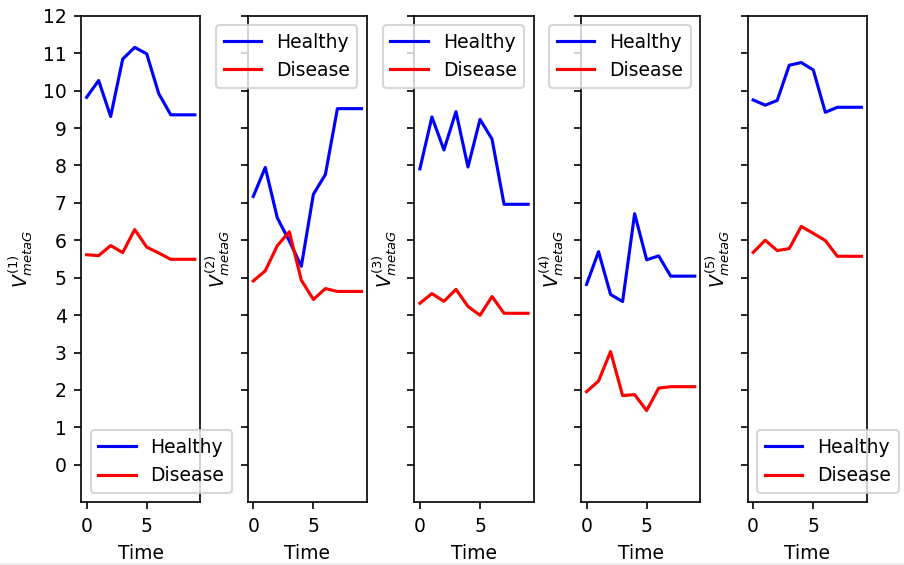}
         \caption{}
         \label{fig:DGBtop5_metagenom}
     \end{subfigure}
     \end{minipage}
        \caption{Top $10$ variables selected by DGB from the $\rm{(a)}$ host-transcriptomics, $\rm{(b)}$ metabolomics and $\rm{(c)}$ metagenomics views, along with their combined and normalized eff$\_$prop scores. In $\rm{(d)}$, the top $5$ host-transcriptomics genes  are statistically compared between the two classes using violin plots. In $\rm{(e)}$ and $\rm{(f)}$ respectively, the top $5$ metabolites and metagenomic pathways are compared between the two classes using mean time-series plots.}
        \label{fig:DGB_onepage}
\end{figure*}

Figure \ref{fig:venn} shows the intersection between the sets of variables selected by LMM, DGB and JPTA for the three views. Figures \ref{fig:DGB_onepage} (a)-(c) show the top $10$ variables selected by DGB from each view.
In Figure \ref{fig:DGBtop5_hosttx}, we use violin plots to show the distribution of the  top $5$ host-transcriptomic genes selected by DGB. The median expression of the genes are different between the two classes. 
Furthermore, in Figures \ref{fig:DGBtop5_metabolom} and \ref{fig:DGBtop5_metagenom}, we show the mean time-series curves for the metagenomics and metabolomics views. In these figures, the average of the univariate time-series of all the participants in the IBD and non-IBD classes is used to calculate two mean curves for the top five variables. Figure \ref{fig:DGB_onepage} is exclusive to the DGB approach. Similar analyses for the LMM and JPTA methods are provided in the Supplementary material, with corresponding figures.

\subsubsection{Literature Analysis of Top Variables} \label{sec:TopDGB}
There is evidence in the literature to support an association between many of the highly-ranked variables and IBD status. 
We first consider a few host transcriptomics genes selected by LMM or DGB.  The IFITM genes have been associated with the pathogenesis of gastro-intestinal tract~\cite{Alteber2018-dr}. LIPG has been observed to have altered level in Ulcerative Colitis (UC) tissue~\cite{Penrose2021-hy}. AQP9 has been shown to have predictive value in Crohn's disease~\cite{yu2021}. CXCL5 has been observed to have significantly increased levels in IBD patients~\cite{Singh2015-rw}. FCGR3B is associated with Ulcerative Colitis (UC) susceptibility~\cite{asano2013}. The MMPs (matrix metalloproteinases) like MMP3 and MMP10 have been shown to be upregulated in IBD~\cite{OSullivan2015-at, Fonseca-Camarillo2021-ip}.
DUOXA2 has been substantiated as an IBD risk gene~\cite{Grasberger2021-au}. The genes S100A8 and S100A9 have been linked with colitis-associated carcinogenesis~\cite{Zhang2017-au}. LILRA3 has been observed to be increased in IBD patients~\cite{Lan2020-yr}.
\par
We next consider some of the top metabolites selected by LMM, DGB or JPTA. 
Uridine has been identified as a therapeutic modulator of inflammation and has been studied in the context of providing protective effects against induced colitis in mice~\cite{Jeengar2017-rp}. Suberate is one of the metabolites significantly affected by neoagarotetraose supplementation (which is a hydrolytic product of agar used to alleviate intestinal inflammation)~\cite{Liu2020-ng}. 
The authors of~\cite{Qin2012-ky} suggest saccharin to be a potential key causative factor for IBD. Docosapentaenoic Acid (DPA) has been shown to alleviate UC~\cite{Dong2022-hb}. Decrease of pantothenic acid in the gut has been remarked as a potential symptom of IBD-related dysbiosis~\cite{Pratt2021}. Valerate has been observed to be altered in UC patients~\cite{Zhuang2019-yc}. It has been suggested that uracil production in bacteria could cause inflammation in the gut~\cite{Lee2013-cc}. Thymine is a pyrimidine that binds to adenine, and adenine has been suggested as a nutraceutical for the prevention of intestinal inflammation~\cite{Fukuda2016}. Ethyl glucuronide is used as a biomarker to diagnose alcohol abuse~\cite{Morini2006-aj}, and alcohol consumption is common in IBD patients~\cite{Piovezani_Ramos2021-vh}.
 \par
We next consider the metagenomics pathways selected by LMM, DGB or JPTA. We find that 
the genus Alistepis~\cite{Parker2020-th}, Roseburia~\cite{nie2021}, Blautia~\cite{Liu2021-vo} and Akkermansia~\cite{Zheng2023-hu} have been often linked to IBD and gut health. Several unintegrated pathways involving these genus have been identified by the DGB method as significant. Butanol has been identified as statistically significant (using univariate analysis) in IBD and non-IBD groups~\cite{ahmed2016}, and the pathway PWY-7003 selected by LMM is associated with glycerol degradation to butanol. Thiazole has been linked with anti-inflammatory properties against induced colitis in mice and pathway PWY-6892 (selected by LMM) is associated with thiazole biosynthesis. Tryptophan has been shown to have a role in intestinal inflammation and IBD~\cite{Li2021-hr} and the pathway TRPSYN-PWY associated with L-tryptophan biosynthesis is one of the key pathways selected by LMM. Increased levels of L-arginine is correlated with the disease severity of UC~\cite{Hong2010-cr} and pathway PWY-7400 linked with L-arginine biosynthesis has been selected by JPTA. Thiamine is associated with symptoms of fatigue in IBD~\cite{Costantini2013-jb} and pathway PWY-7357 associated with thiamine formation is selected as important by JPTA.

As evidenced by these examples from the literature, we illustrate that many genes, metabolites and pathways selected by the three methods have been linked to IBD.
There are some selected variables, however, that may not have been directly examined in the context of IBD. For instance, we could not find a direct link of the metagenomic pathways: PWY-7388 (selected in top $25$ by JPTA) to IBD. However, this pathway has been linked with psoriasis, and it has been observed that patients with psoriasis have increased susceptibilities to IBD~\cite{Chang2022-yw}.
Thus, the unstudied genes/metabolites/pathways that the three variable selection methods have discovered may potentially be novel variables linked to IBD.

\section{Synthetic Analysis of EC and FPCA}
FPCA and Euler curves (EC) provide important one-dimensional representations of longitudinal data. These methods hold particular significance because the extracted features can be used with a broad spectrum of existing integration methods that only allow cross-sectional views. Using synthetic simulations, we unravel key properties of EC and FPCA. These simulations demonstrate that Euler curves are better at distinguishing between classes when the covariance structure of the variables differs from one class to another. FPCA performs better when the time-trend of the variables differs between the classes. In addition to comparing EC and FPCA, we also illustrate the performance of the direct DeepIDA-GRU approach, where the feature extraction step is skipped and the longitudinal views are directly fed into DeepIDA-GRU. GRUs have the ability to distinguish certain complicated time-trend differences that pose challenges for FPCA and EC methods. However, FPCA and EC are computationally faster compared to GRUs. Moreover, as demonstrated by these simulations, training a GRU can be challenging and one needs to closely monitor problems like overfitting, diminishing gradients, hyperparameter tuning etc and their out-of-the-box performance may be even worse than the simpler methods like FPCA and EC. 

To compare EC, FPCA and direct DeepIDA-GRU, we use the following three approaches on synthetically generated multiview longitudinal data: (i) \textbf{DeepIDA-EC:} EC is used to extract one-dimensional features from the longitudinal views and the extracted features are fed into the DeepIDA network for integration and classification, (ii) \textbf{DeepIDA-FPC:} FPCA is used to extract features from the longitudinal views and the extracted features are fed into the DeepIDA network for integration and classification and (iii) \textbf{DeepIDA-GRU:} No one dimensional features are extracted from the longitudinal views and the data from the longitudinal views are directly fed into the DeepIDA-GRU network for integration and classification. The synthetic datasets we generate are balanced between classes and we use the classification accuracy as a metric for this comparison. \par
Here, we consider $K=2$ classes, $D=2$ views and $N = 500$ subjects. Each view $d$ (for $d \in [1:2]$) consists of $p_d = 250$ variables and $T=20$ time points. We denote by $C_1$ and $C_2$ the noise covariance matrices corresponding to classes $1$ and $2$, respectively. These covariance matrices are constructed as follows
\begin{align}
    C_1 &= C_{\text{unif}}' \cdot C_{\text{unif}},  \nonumber \\
    C_2 &= (1-\epsilon) C_1 + \epsilon \left(C_{\text{power}}' \cdot C_{\text{power}} \right),
\end{align}
where both $C_{\text{unif}}$ and $C_{\text{power}}$ are $p_1+p_2$ by $p_1+p_2$ matrices whose entries are identically and independently generated from the Uniform$(0,1)$ and Power$(10)$ distributions, respectively. Here, Power$(10)$ is the power distribution (inverse of Pareto distribution) with parameter $a=10$, whose probability density function is given by $f(x;a) = a x^{a-1}, x \in [0,1], {a \in (0,\infty)}$. Moreover, $\epsilon$ is a parameter that we manipulate to vary the amount of structural difference between $C_1$ and $C_2$. In particular, when $\epsilon=0$, $C_1 = C_2$ and the two classes have the same covariance structure. When $\epsilon = 1$, the entries of $C_1$ and $C_2$ have completely different and uncorrelated distributions. \par
We let $\phi_{d,k}$ and $\delta_{d,k}$ be the auto-regressive (AR) and moving-average (MA) parameters, respectively, for the ARMA(1,1) process, corresponding to the $d$-th view ($d \in \{1,2\}$) and the $k$-th class ($k \in \{1,2\}$). For these simulations, these ARMA parameters for the two classes are chosen to be
\begin{align}
    \phi_{1,1} &= 0.5, \phi_{2,1} = 0.7, \phi_{1,2} = 0.5-\eta, \phi_{2,2} = 0.7-\eta, \\
    \delta_{1,1} &= 0.4, \delta_{2,1} = 0.6, \delta_{1,2} = 0.4-\eta, \delta_{2,2} = 0.6-\eta,
\end{align}
where $\eta$ is another parameter that is varied to control the amount of difference between the ARMA parameters of the two classes. 
\par
The synthetic longitudinal data of subject ${n \in [1:N]}$ for view ${d \in [1:2]}$ is given by a collection of $T$ vectors: $\mathbf{X}_{d}^{(n,:,:)}= \{\mathbf{X}_{d}^{(n,:,1)}, \mathbf{X}_{d}^{(n,:,2)}, \cdots, \mathbf{X}_{d}^{(n,:,T)}\}$ (where ${\mathbf{X}_{d}^{(n,:,t)} \in \mathbb{R}^{p_d}}$). Let $\mathbf{X}^{(n,:,t)} = \left[\mathbf{X}_{1}^{(n,:,t)}, \mathbf{X}_{2}^{(n,:,t)}\right]'$, $\mathbf{w}^{(n,t)} = \left[\mathbf{w}_{1}^{(n,t)}, \mathbf{w}_{2}^{(n,t)}\right]'$, $\bm{\phi}_{:,\kappa(n)} = \left[{\phi}_{1,\kappa(n)},{\phi}_{2,\kappa(n)}\right]'$ and $\bm{\delta}_{:,\kappa(n)} = \left[{\delta}_{1,\kappa(n)},{\delta}_{2,\kappa(n)}\right]'$, where $\kappa(n)$ is the class to which the $n$-th subject belongs to and the vectors $\mathbf{w}_{1}^{(n,t)}$ and $\mathbf{w}_{2}^{(n,t)}$ are jointly distributed as $\mathbf{w}^{(n,t)} \sim \mathcal{N}(\mathbf{0}, C_{\kappa(n)})$ for all $t \in [1:T]$. Then the multiview data $\mathbf{X}^{(n,:,t)}$ of subject $n$ at time $t$ is generated according to an ARMA$(1,1)$ process with AR and MA parameters given by $\bm{\phi}_{:,\kappa(n)}$ and $\bm{\delta}_{:,\kappa(n)}$, respectively, and noise covariance matrix $C_{\kappa(n)}$ as follows.
\begin{align} \label{eq:arma}
    \mathbf{X}^{(n,:,t)} \!\!= \bm{\phi}_{:,\kappa(n)} \!\circ\! \mathbf{X}^{(n,:,t\!-\!1)} \!\!+ 
     \mathbf{w}^{(n,:,t)} \!\!+   \bm{\delta}_{:,\kappa(n)} \!\circ\! \mathbf{w}^{(n,:,t\!-\!1)} 
\end{align}
where, $\circ$ represents element wise product.
\begin{figure*}[ht]
\captionsetup{justification=centering}
     \centering
     \begin{subfigure}[p]{0.2\textwidth}
         \centering
         \includegraphics[width=\textwidth]{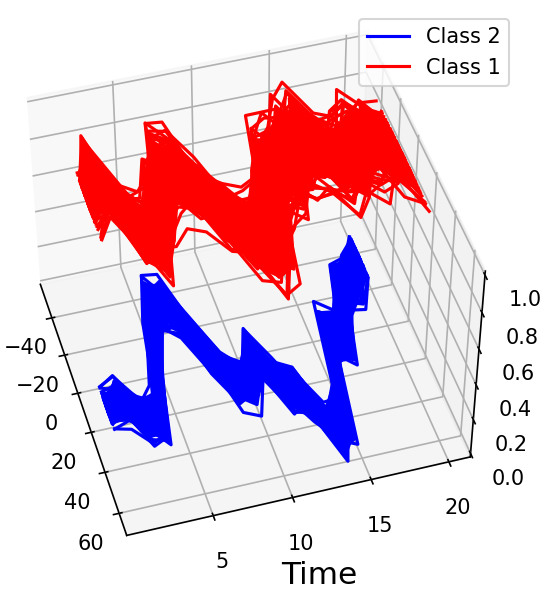}
         \caption{Longitudinal data of $2$ out of $N$ subjects from different classes.}
         \label{fig:DiffCovSameARMA-1}
     \end{subfigure}
     \hspace{0mm}
     \begin{subfigure}[p]{0.29\textwidth}
         \centering
         \includegraphics[width=\textwidth]{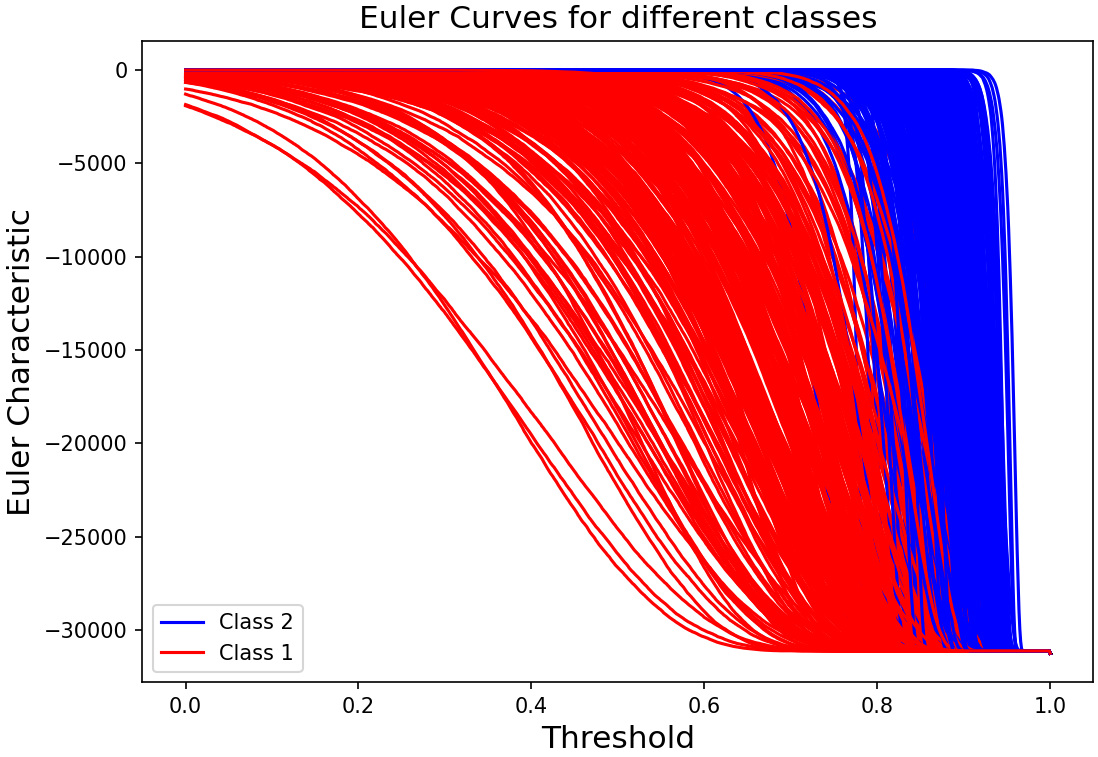}
         \caption{Euler Curves of all the $N$ subjects belonging to either of the two classes.}
         \label{fig:DiffCovSameARMA-2}
     \end{subfigure}
     \hspace{0mm}
     \begin{subfigure}[p]{0.24\textwidth}
         \centering
         \includegraphics[width=\textwidth]{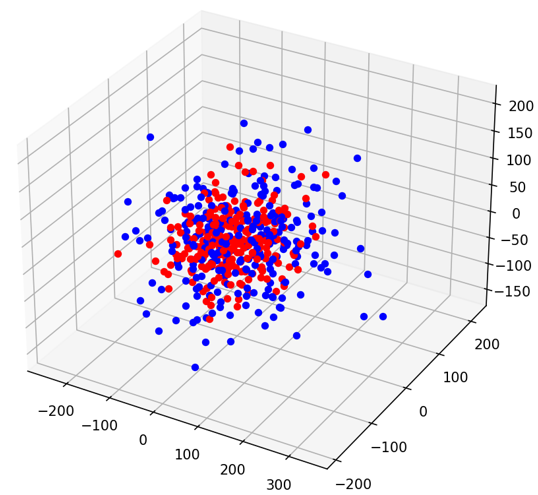}
         \caption{$3$ dimensional FPC scores of one of the variables for all the $N$ subjects.}
         \label{fig:DiffCovSameARMA-3}
     \end{subfigure}
        \caption{Visual comparison between EC and FPCA: Case 1 - The two classes have different covariance matrices (with $\epsilon=0.75$), but the same ARMA parameters (that is, $\eta = 0$). Euler curves can distinguish between the two classes whereas FPCA cannot.}
        \label{fig:DiffCovSameARMA}
\end{figure*}

\begin{figure*}[ht]
\captionsetup{justification=centering}
     \centering
     \begin{subfigure}[p]{0.2\textwidth}
         \centering
         \includegraphics[width=\textwidth]{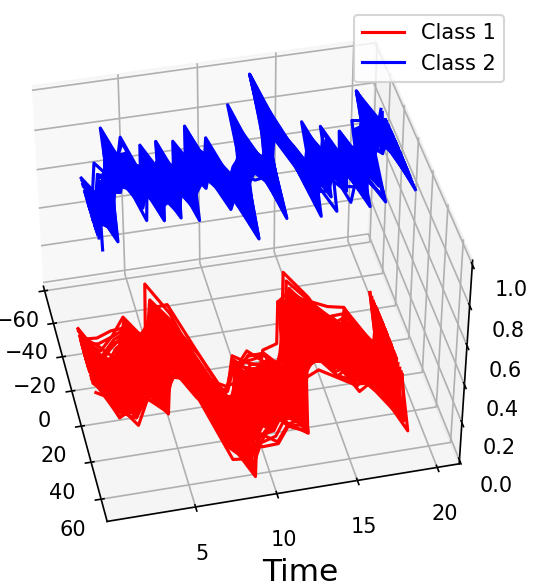}
         \caption{Longitudinal data of $2$ out of $N$ subjects from different classes.}
         \label{fig:SameCovDiffARMA-1}
     \end{subfigure}
     \hspace{0mm}
     \begin{subfigure}[p]{0.29\textwidth}
         \centering
         \includegraphics[width=\textwidth]{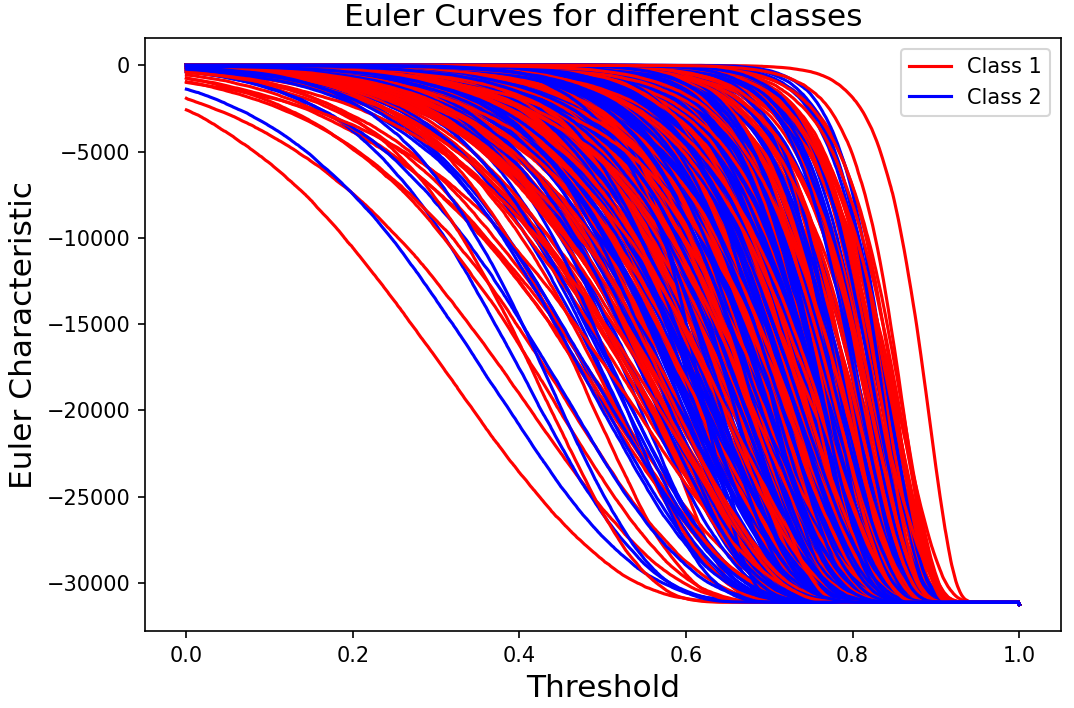}
         \caption{Euler Curves of all the $N$ subjects belonging to either of the two classes.}
         \label{fig:SameCovDiffARMA-2}
     \end{subfigure}
     \hspace{0mm}
     \begin{subfigure}[p]{0.24\textwidth}
         \centering
         \includegraphics[width=\textwidth]{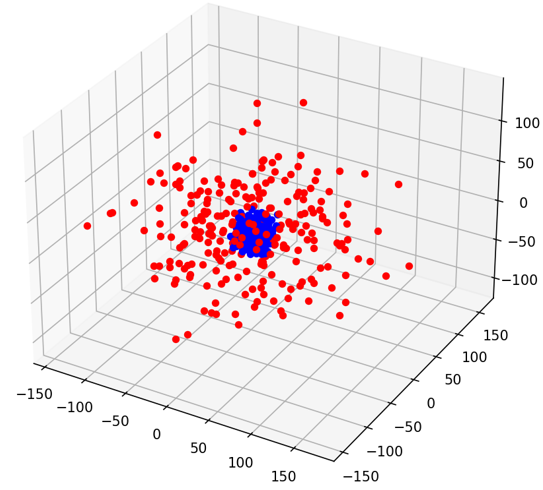}
         \caption{$3$ dimensional FPC scores of one of the variables for all the $N$ subjects.}
         \label{fig:SameCovDiffARMA-3}
     \end{subfigure}
        \caption{Visual comparison between EC and FPCA: Case 2 - The two classes have the same covariance matrix (that is, $\epsilon=0$), but different ARMA parameters (with $\eta = 0.75$). FPCA can distinguish between the two classes whereas EC cannot.}
        \label{fig:SameCovDiffARMA}
\end{figure*}
For any given value of $(\eta, \epsilon)$, a total of $100$ longitudinal multiview datasets are generated randomly according to equation~\eqref{eq:arma}, where in each dataset, there are approximately $50$ percent of the subjects in class $1$ and class $2$ respectively. The three approaches: DeepIDA-GRU, DeepIDA-EC and DeepIDA-FPC are used for the classification task. All the feed-forward networks consist of $3$ layers with $[200,100,20]$ neurons, respectively. All the GRUs contain $3$ layers with $256$ dimensional hidden vector. The synthetic analysis is divided into two cases:
\begin{figure*}[t]
\captionsetup{justification=centering}
         \centering
         \begin{subfigure}[p]{\columnwidth}
         \centering
         \includegraphics[width=0.6\textwidth,height=0.46\textwidth]{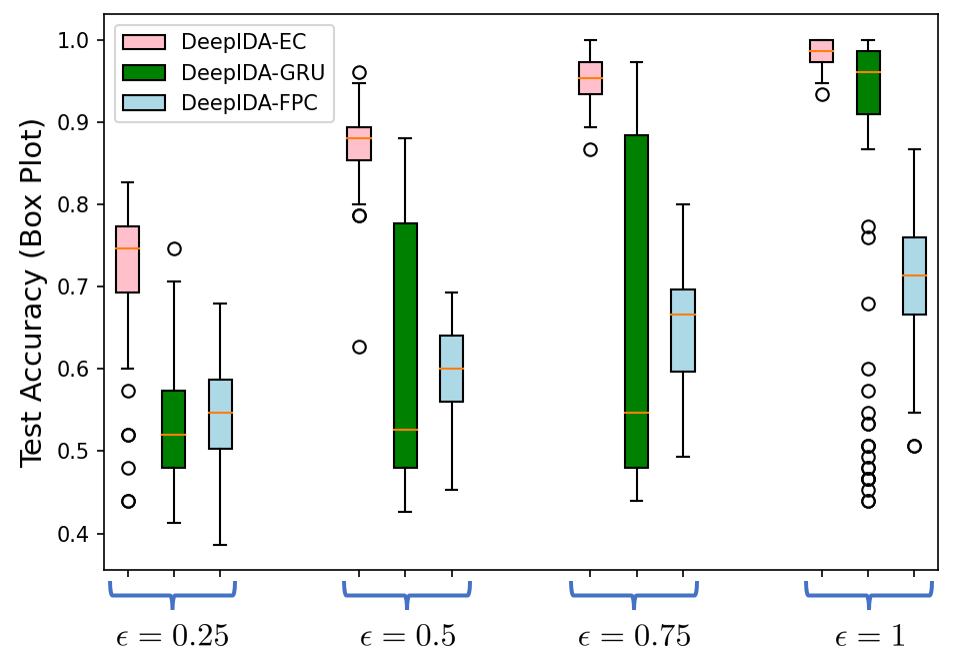}
         \caption{Case 1}
         \label{fig:box_plot_DiffCov}
     \end{subfigure}
     \hspace{0mm}
    \begin{subfigure}[p]{\columnwidth}
         \centering
         \includegraphics[width=0.6\textwidth]{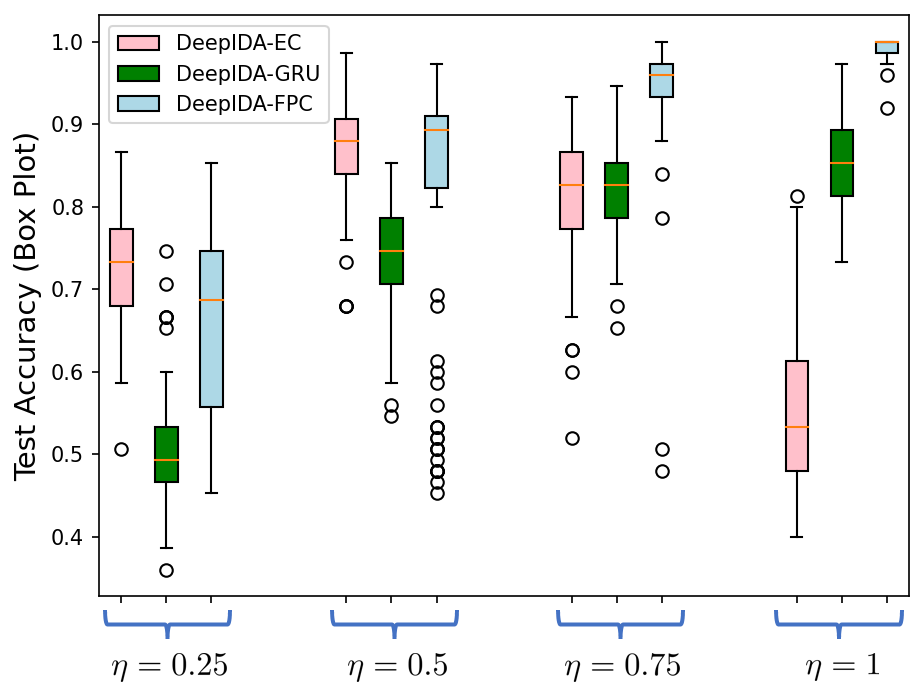}
         \caption{Case 2}
  \label{fig:box_plot_DiffARMA}
     \end{subfigure}
     \caption{Box plots are used to compare the accuracy of the three methods: DeepIDA-GRU, DeepIDA-FPC and DeepIDA-EC in two cases. In Case 1, the covariance structure between the two classes differs, with the difference being proportional to the magnitude of $\epsilon \in \{0.25,0.5,0.75,1\}$, but the ARMA parameters are the same (i.e. $\eta = 0$). In Case 2, the ARMA parameters between the two classes differ, with the difference being proportional to the magnitude of $\eta \in \{0.25,0.5,0.75,1\}$, but the covariance structure is the same (i.e. $\epsilon = 0$). The box plots illustrate the spread of accuracies attained by the three methods for $100$ distinct synthetically generated datasets, for each pair of $(\epsilon,\eta)$ values.}

\end{figure*}
\label{sec:synthetic}

\begin{enumerate}
    \item \textbf{Case 1: Different Covariance Matrices, Same ARMA Parameters: } In this case, $\eta = 0$ and $\epsilon$ assumes the following set of values: $\epsilon \in \{0.25,0.5,0.75,1\}$. Note that larger $\epsilon$ means more difference in the covariance structure of the variables between the two classes and therefore easier for the methods to classify. Figure~\ref{fig:DiffCovSameARMA} provides a visual depiction of the superiority of EC compared to FPCA for this case. In this figure, the synthetic multivariate time-series data (of view $1$), generated with $\epsilon = 0.75$, for one subject of class $1$ and one subject of class $2$ are shown in Figure~\ref{fig:DiffCovSameARMA-1}. The EC curves of all the $N$ subjects are shown in Figure~\ref{fig:DiffCovSameARMA-2} (which shows that the EC curves can clearly distinguish between the two classes). The $3$ dimensional FPC scores are plotted in Figure~\ref{fig:DiffCovSameARMA-3} which shows that FPCA cannot distinguish between the two classes in this case. In Figure~\ref{fig:box_plot_DiffCov}, we compare the performance of the three methods on $100$ randomly generated datasets for each $\epsilon \in \{0.25,0.5,0.75,1\}$. In particular we use box plots to summarize the classification accuracy achieved by the three methods on these $100$ datasets.

    \item \textbf{Case 2: Same Covariance Matrices, Different ARMA Parameters, No Reverse Operation: } In this case, $\epsilon = 0$ and $\eta$ takes values in $\eta \in \{0.25,0.5,0.75,1\}$. Note that larger $\eta$ corresponds to a larger difference in the ARMA parameters between the two classes and therefore easier for the methods to classify. Figure~\ref{fig:SameCovDiffARMA} shows visually that FPCA can clearly distinguish between the two classes, whereas EC cannot. In this figure, the synthetic multivariate time-series data (of view $1$), generated with $\eta = 0.75$, for one subject of class $1$ and one subject of class $2$  are shown in Figure~\ref{fig:SameCovDiffARMA-1}. The EC curves of all the $N$ subjects are shown in Figure~\ref{fig:SameCovDiffARMA-2} (which shows that the EC curves are unable to distinguish between the two classes). The $3$ dimensional FPC scores are plotted in Figure~\ref{fig:SameCovDiffARMA-3} which shows that FPCA performs better at distinguishing between the two classes in this case. In Figure~\ref{fig:box_plot_DiffARMA}, we compare the performance of the three methods on $100$ randomly generated datasets for each $\eta \in \{0.25,0.5,0.75,1\}$. In this figure, we summarize the classification accuracy achieved by the three methods on these $100$ datasets using box plots.
\end{enumerate}

\begin{rem} \label{rem:deepIDA-simulated}
    The box plots of Figures~\ref{fig:box_plot_DiffCov} and~\ref{fig:box_plot_DiffARMA} show that DeepIDA-EC performs better at classifying subjects when the covariance structure of the two classes is different, whereas DeepIDA-FPC performs better when the ARMA parameters of the two classes differ. DeepIDA-GRU does not outperform either of these methods even though it has the potential to handle much more complex tasks. This could be mainly due to the fact that the DeepIDA-GRU network was not fine tuned for each of the randomly generated datasets, which could have led to overfitting/underfitting on many of these datasets.
\end{rem}


\section{Discussion}
\label{sec:conclusion}
Data collected from multiple sources are increasingly being generated in many biomedical research. These data types could be a mix of cross-sectional and longitudinal data. However, literature for integrating cross-sectional and longitudinal data is scarce. This work began to fill in the gaps in existing literature. Motivated by, and used for the analysis of data from, the IBD study of the integrated Human Microbiome Project (iHMP), we have proposed a deep learning pipeline for (i) integrating both cross-sectional and longitudinal data from multiple sources while simultaneously discriminating between disease status; and (ii) identifying key molecular signatures  contributing to the association among the views and separation between classes within a view. Our pipeline combines the strengths in statistical methods, such as the ability to make inference, reduce dimension, and extract longitudinal trends, with the flexibility of deep learning, and consists of variable selection/ranking, feature extraction, and joint integration and classification. For variable selection/ranking, methods applicable to one view at a time (i.e. linear mixed models [LMM]), two longitudinal views (i.e. joint principal trend analysis [JPTA]), and multiple cross-sectional and longitudinal views (i.e. Deep-IDA-GRU with bootstrap [DGB]) were considered. For feature extraction, we considered functional principal component analysis (FPCA) and Euler Characteristics (EC) curves. For integration and classification, we implemented Deep IDA with gated recurrent units [DeepIDA-GRU]. 

When we applied the pipeline to the motivating data, we observed that for variable selection, LMM and DGB achieved slightly better performance metrics than JPTA likely because they are supervised methods-- information on class labels is used in variable selection-- and they are applicable to more than two views.
For feature extraction, the performance of both EC and FPCA was comparable, and both methods performed similar to the direct DeepIDA-GRU approach with no feature extraction. Our work identified multi-omics signatures (genes, metabolites) and microbial pathways discriminating between patients with and without IBD.  Some of the molecules identified have been found to be associated with IBD in the literature, corroborating previous findings, while others have been implicated in other diseases that have been linked to IBD, thus providing likely candidates of molecules to be explored in IBD pathobiology.  We also compared the performance of EC and FPCA using synthetic datasets and found that these methods outperformed one another under different scenarios. EC performed better when the covariance structure of the variables was different between the two classes, while FPCA outperformed EC when there was a difference in the time trends between the two classes.  

Deep learning is typically used with a large sample size to ensure generalizability. The main limitation of this work is the small sample size ($n=90$ subjects) of the IBD data which motivated our work but we attempted to mitigate against potential overfitting/underfitting issues through the use of variable selection, feature extraction and leave-one-out-cross-validation instead of n-fold cross-validation (which would have significantly reduced the sample size for training). Variable selection is widely regarded as an effective technique for high-dimensional small sample size (HDLSS) data and it helps to avoid overfitting and high-variance gradients~\cite{ijcai2017p318, TSAI2020106097}. In this work, we explored both linear methods of variable selection (LMM and JPTA) and non-linear deep learning based methods (DGB). Further analysis is need to see whether the bootstrapping procedure of DGB scales well with increasing data sizes. Future work could consider validating the proposed methodology on multiview data with larger sample sizes. Additionally, for larger and more complicated data, it may be worthwhile  to investigate whether integrating other deep learning networks like transformers and 1D convolutional networks in the DeepIDA pipeline would yield better results for handling longitudinal data than the DeepIDA with GRUs implemented in this work. 

Despite the above limitations, we believe that our pipeline for integrating longitudinal and cross-sectional data from multiple sources that combines statistical and machine learning methods fills an important gap in the literature for data integration and will enable biologically meaningful findings. Our extensive investigation of the scenarios under which FPCA outperforms EC curves and vice versa sheds light on the specific scenarios for using these methods. Further, our real data application has resulted in the identification of molecules and microbial pathways, some implicated in the literature and thus providing evidence to corroborate previous findings, while others are potentially novel and could be explored for their role in IBD pathobiology. 

\begingroup
\let\cleardoublepage\clearpage
 \bibliography{references.bib}
 \bibliographystyle{IEEEtran}
\endgroup
\end{document}


\begin{figure*}[t]
  \centering
  \LARGE\textbf{Supplementary Material}
\end{figure*}

\section{More on Methods}
In this section, we provide more details about some of the methods discussed in the main manuscript. 
\subsection{Gated Recurrent Units (GRUs)} \label{sec:GRUs}

Recurrent Neural Networks (RNNs) are a class of deep learning networks that specialize in processing sequential data. A key feature of an RNN is that it uses something referred to as the hidden state which serves as a memory for the network. In particular, the role of the hidden state is to store and summarize the context from the past data of the sequence. At any given time point, the hidden state is a function of both the previous hidden state and the input at the current time point. This recurrent relationship allows the information to persist over time. The learning algorithm that the RNNs use is called Back Propagation Through Time (BPTT), which propagates gradients backward through time. A major issue of BPTT is the vanishing/exploding gradient problem, where the gradients back propagated through many time steps become extremely small or extremely large. This makes it challenging for the RNNs to learn long term dependencies in the sequence.

\begin{figure}[t]
\captionsetup{justification=centering}
         \centering
    \includegraphics[width=0.6\columnwidth]{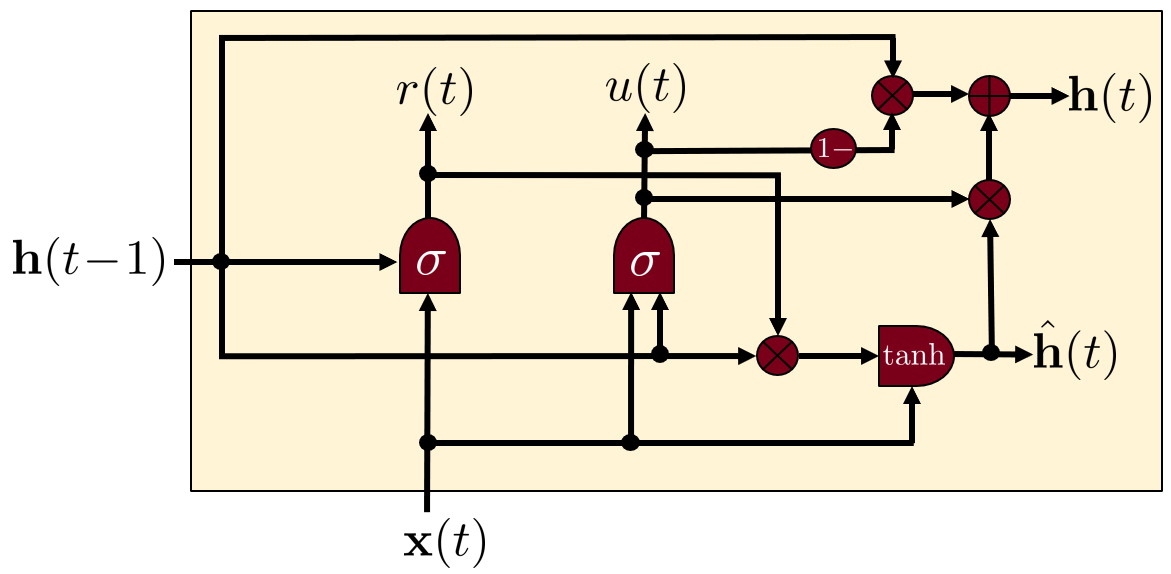}
         \caption{Pictorial Representation of a Gated Recurrent Unit (GRU) described in equation~\eqref{eq:gru}.}
  \label{fig:gru}
\end{figure}

Gated Recurrent Units (GRUs), introduced in 2014 by~\cite{cho-etal-2014-learning}, are a class of RNNs that can learn long term dependencies in sequential data and aid in mitigating vanishing/exploding gradient problem of the vanilla RNNs. GRUs use a gating mechanism which can help them form long term dependencies and forget irrelevant information. In particular, given an input vector $\mathbf{x}(t)$ at time $t$, a GRU can be described using the following set of recursive relations
\begin{align} \label{eq:gru}
    u(t) &= \sigma_u\! \left(\mathbf{W}_{ux} \mathbf{x}(t) + \mathbf{W}_{uh} \mathbf{h}(t-1) +\mathbf{b}_u\right),\nonumber \\
    r(t) &= \sigma_r\! \left(\mathbf{W}_{rx} \mathbf{x}(t) + \mathbf{W}_{rh} \mathbf{h}(t-1) +\mathbf{b}_r\right),\nonumber \\
    \widehat{\mathbf{h}}(t) &= \phi_u \!\left(\mathbf{W}_h \mathbf{x}(t) + \mathbf{W}_{u} (r(t) \odot \mathbf{h}(t-1)) +\mathbf{b}_h\right), \nonumber \\
    \mathbf{h}(t) &= (1-u(t)) \odot \mathbf{h}(t-1) + u(t) \odot \widehat{\mathbf{h}}(t),
\end{align}
where $\odot$ represents the Hadamard product. Moreover, (i) $\mathbf{h}(t-1)$ is the hidden state of time $t-1$ and $\widehat{\mathbf{h}}(t)$ is a candidate for the hidden state at time $t$; (ii) $u(t)$ represents the update probability/weight (if $u(t)=0$, the hidden state at time $t$ is not updated, that is, $\mathbf{h}(t) =\mathbf{h}(t-1)$, whereas if $u(t) = 1$, the hidden state is changed to the candidate state $\widehat{\mathbf{h}}(t)$); (iii) $r(t)$  is the reset probability which determines if the candidate hidden state $\widehat{\mathbf{h}}(t)$ should depend on the previous hidden state $\mathbf{h}(t-1)$ (in other words, if it should use the information from the past) or if it should only depend on the current input (in other words, reset and throw away the past memory); and finally (v) $\mathbf{W}_{ux}, \mathbf{W}_{uh}, \mathbf{W}_{rx}, \mathbf{W}_{rh}, \mathbf{W}_{h}, \mathbf{W}_{u},\mathbf{b}_{u}, \mathbf{b}_{r}, \mathbf{b}_{h}$ are the network's weights and biases (that are learned via training) and $\sigma_u, \sigma_r, \phi_u$ are the activations (typically, $\sigma_u, \sigma_r$ are chosen to be `sigmoid' and $\phi_u$ is chosen to be `tanh'). The recursive equations of GRU are pictorially represented in Figure~\ref{fig:gru}. By using the reset gate, the model learns to determine how much of the previous information should be retained/forgotten. The update gate helps the model learn what proportion of the new candidate hidden state and the previous hidden state should be mixed to determine the current state. Through these mechanisms of reset and update, the model can learn long term dependencies while avoiding the problems of exploding/vanishing gradients. Ever since their inception, GRUs have been extensively used with multivariate time series data for tasks like classification, anomaly detection, regression, forecasting etc. In this work, GRUs are used inside the DeepIDA-GRU network to non-linearly transform data from longitudinal views.

\subsection{DeepIDA-GRU}
\label{sec:deep-ida-gru}

DeepIDA was introduced in~\cite{safo2021} as a method to learn non-linear projections of the different views that simultaneously maximize the separation between classes and the association between views. Moreover, DeepIDA in conjunction with bootstrapping provided variable ranking for interpretability. DeepIDA, in its original form, only supports cross-sectional data. Each of the $D$ views' data is first fed into its own dense feed-forward neural network. The outputs of all these neural networks then go into the IDA optimization setup wherein each view is projected onto a $K-1$ dimensional space and the projection is such that the correlation between the different views and the inter-class separation are simultaneously maximized. Nearest centroid classifiers are then used to classify each subject into one of the $K$ classes. The DeepIDA network is trained end to end: from the input of the neural networks to the loss function of IDA. \par

 In order to expand its ability to accept longitudinal data as well, in this work, DeepIDA is integrated with GRUs. This modified network is referred to as DeepIDA-GRU. 
 In this network, data from each cross-sectional view is fed into its respective dense neural network, whereas data from each longitudinal view is fed into its respective GRU network. The role of the neural networks and the GRUs is to non-linearly transform the data of each view. Following the notation of the main manuscript, let $\mathbf{X}_d \in \mathbb{R}^{N \times p_d \times t_d}$ be a tensor representing the 
longitudinal (if $t_d > 1$) or cross-sectional (if $t_d = 1$) data corresponding to the $d$-th view (for $d \in [1:D]$), for the $N$ subjects. Here, $p_d$ and $t_d$ represent the number of variables and the number of time-points, respectively, in view $d$. Moreover, let $\mathcal{N}_d$ (for $d \in [1:D]$) represent the deep network corresponding to the $d$-th view, where $\mathcal{N}_d$ is a dense neural network if $t_d = 1$ and $\mathcal{N}_d$ is a GRU if $t_d > 1$; and let $\theta_d$ represent all the parameters (weights and biases) of $\mathcal{N}_d$. Assume the output dimension (or the number of hidden units in the final layer) of $\mathcal{N}_d$ to be $o_d$. The input to $\mathcal{N}_d$ is $\mathbf{X}_d$ and the output of $\mathcal{N}_d$ is denoted by $\mathbf{H}_d \in \mathbb{R}^{o_d \times N}$. The outputs for the $D$ views $\{\mathbf{H}_d : d \in [1:D]\}$ are fed into the IDA optimization problem, which finds the projection matrix that maximizes the association between views and separation between classes.

The IDA optimization finds $\ell = \min\{K-1, o_1, \cdots, o_D\}$ projection vectors for each view. These vectors are denoted by $\mathbf{P}_d = [\mathbf{p}_{d,1}, \mathbf{p}_{d,2}, \cdots, \mathbf{p}_{d,\ell}] \in \mathbb{R}^{o_d \times \ell}$. When the non-linearly transformed data $\mathbf{H}_d$ is projected onto $\mathbf{P}_d$ (for $d \in [1:D]$), the
linear association between the views is maximized and the classes within each view are maximally linearly separated. For view $d$, let $\mathbf{H}_d = [\mathbf{H}_{d,1}, \mathbf{H}_{d,2}, \cdots, \mathbf{H}_{d,K}]$, where $\mathbf{H}_{d,k} \in \mathbb{R}^{o_d \times N_k}$ (for $k \in [1:K]$) corresponds to outputs from network $\mathcal{N}_d$, of subjects belonging to the $k$-th class; and $N_k$ is the number of subjects in the $k$-th class (therefore, $\sum_{k=1}^{K} N_k = N$). In particular, $\mathbf{H}_{d,k} = \left[\mathbf{h}_{d,k}^{(1)}, \mathbf{h}_{d,k}^{(2)}, \cdots, \mathbf{h}_{d,k}^{(N_k)}\right]$ where $\mathbf{h}_{d,k}^{(j)} \in \mathbb{R}^{o_d}$ for $j \in [1:N_k]$ and $k \in [1:K]$. Using view $d$'s non-linearly transformed data $\mathbf{H}_d$, the between-class covariance $\mathbf{M}^{(\text{B})}_d$, the total covariance $\mathbf{M}^{(\text{T})}_d$ and the cross covariance $\mathbf{M}_{i-j}$ between views $i$ and $j$ are computed as follows

\begin{align}
    \mathbf{M}^{(\text{B})}_d &= \frac{1}{N-1} \sum_{k=1}^K N_k (\bm{\mu}_{d,k} - \bm{\mu}_{d}) (\bm{\mu}_{d,k} - \bm{\mu}_{d})^T, \nonumber\\
    \mathbf{M}^{(\text{T})}_d &= \frac{1}{N-1} \sum_{k=1}^{K}\sum_{n=1}^{N_k}  (\mathbf{h}_{d,k}^{(n)} - \bm{\mu}_{d}) (\mathbf{h}_{d,k}^{(n)} - \bm{\mu}_{d})^T, \nonumber\\
     \mathbf{M}_{i-j} &= \frac{1}{N-1} \sum_{k=1}^{K}\sum_{n=1}^{N_k}  (\mathbf{h}_{i,k}^{(n)} - \bm{\mu}_{i}) (\mathbf{h}_{j,k}^{(n)} - \bm{\mu}_{j})^T,\nonumber
\end{align}
where $\bm{\mu}_{d,k} = \frac{1}{N_k} \sum_{n=1}^{N_k} \mathbf{h}_{d,k}^{(n)}$ and $\bm{\mu}_d = \frac{1}{K} \sum_{k=1}^K \bm{\mu}_{d,k}$. To obtain the network parameters $\{\theta_d : d \in [1:D]\}$ and the linear projections $\{\mathbf{P}_d : d \in [1:D]\}$, the following optimization problem is solved.

\begin{align}
    \mathbf{M}^{(\text{B})}_d &= \frac{1}{N-1} \sum_{k=1}^K N_k (\bm{\mu}_{d,k} - \bm{\mu}_{d}) (\bm{\mu}_{d,k} - \bm{\mu}_{d})^T, \nonumber\\
    \mathbf{M}^{(\text{T})}_d &= \frac{1}{N-1} \sum_{k=1}^{K}\sum_{n=1}^{N_k}  (\mathbf{h}_{d,k}^{(n)} - \bm{\mu}_{d}) (\mathbf{h}_{d,k}^{(n)} - \bm{\mu}_{d})^T, \nonumber\\
     \mathbf{M}_{i-j} &= \frac{1}{N-1} \sum_{k=1}^{K}\sum_{n=1}^{N_k}  (\mathbf{h}_{i,k}^{(n)} - \bm{\mu}_{i}) (\mathbf{h}_{j,k}^{(n)} - \bm{\mu}_{j})^T,\nonumber
\end{align}
where $\bm{\mu}_{d,k} = \frac{1}{N_k} \sum_{n=1}^{N_k} \mathbf{h}_{d,k}^{(n)}$ and $\bm{\mu}_d = \frac{1}{K} \sum_{k=1}^K \bm{\mu}_{d,k}$. To obtain the network parameters $\{\theta_d : d \in [1:D]\}$ and the linear projections $\{\mathbf{P}_d : d \in [1:D]\}$, the following optimization problem is solved.
\begin{align}
    & \argmax_{\theta_1,\dots, \theta_D, \mathbf{P}_1, \dots \mathbf{P}_D} \mathcal{L} \triangleq \frac{\rho}{D} \sum_{d=1}^{D} \Tr\left[\mathbf{P}_d^{T} \mathbf{M}^{(\text{B})}_d \mathbf{P}_d \right] \nonumber\\
    &+  \frac{2(1-\rho)}{D(D-1)} \sum_{i\in[1:D]} \sum_{j\in [1:D] \setminus \{i\}} \Tr\left[\mathbf{P}_i^T \mathbf{M}_{i-j} \mathbf{P}_j \mathbf{P}_j^T \mathbf{M}_{i-j}^T \mathbf{P}_i\right]\nonumber \\
    &\text{subject to: } \Tr\left[\mathbf{P}_d^T \mathbf{M}^{(\text{T})}_d\mathbf{P}_d \right]= \ell , \forall d \in [1:D], \label{eq:DeepIDA-GRU Loss}
\end{align}
where $\Tr[\cdot]$ represents the trace of the matrix and $\rho$ is a parameter that controls the relative weight of inter-class separation and between-view association in the loss. In this loss function, the first term quantifies the average inter-class separation over the $D$ views and the second term represents the average pairwise squared correlation over all $\binom{D}{2}$ pairs of views. In~\cite{safo2021}, the authors show that for fixed parameters $\{\theta_d: d \in [1:D]\}$, the solution to the above optimization problem can be obtained by solving a system of eigenvalue problems. 
Leveraging this fact, they proposed the following key steps to solve the entire optimization problem.
\begin{enumerate}
    \item[(i) ] For the given network parameters $\{\theta_d : d\in [1:D]\}$, feed forward the inputs $\{\mathbf{X}_d: d \in [1:D]\}$ to obtain the outputs $\{\mathbf{H}_d: d \in [1:D]\}$ and use these outputs to compute the DeepIDA-GRU loss (equation~\eqref{eq:DeepIDA-GRU Loss}).
    \item[(ii) ] The loss function in equation~\eqref{eq:DeepIDA-GRU Loss} is completely determined from $\{\mathbf{H}_d: d \in [1:D]\}$ and $\{\mathbf{P}_d: d \in [1:D]\}$. For the given $\mathbf{H}_d$, first compute the optimal $\mathbf{P}_d$ and then compute the gradients $\frac{\partial \mathcal{L}}{\partial \mathbf{H}_d}$ for all $d \in [1:D]$. 
    \item[(iii) ] Each network $\mathcal{N}_d$ operates independently of other networks $\mathcal{N}_{\widetilde{d}}$ for $d,\widetilde{d} \in [1:D]$ and $\widetilde{d} \neq d$. Compute $\frac{\partial \mathbf{H}_d}{\partial \theta_d}$ for all $d \in [1:D]$ and subsequently use chain rule to obtain $\frac{\partial \mathcal{L}}{\partial \theta_d} = \frac{\partial \mathbf{H}_d}{\partial \theta_d} \frac{\partial \mathcal{L}}{\partial \mathbf{H}_d}$. These gradients are then used to update the network parameters $\{\theta_d : d \in [1:D]\}$.
\end{enumerate}
For a detailed discussion of how the IDA optimization problem~\eqref{eq:DeepIDA-GRU Loss} reduces to solving a system of eigenvalue problem, we direct the readers to~\cite{safo2021}. Similar to the DeepIDA network, the DeepIDA-GRU also uses nearest centroid classifier on the final projections for classification. In summary, because of the inclusion of GRUs, DeepIDA-GRU is capable of handling both cross-sectional views and longitudinal views. It is important to note that there are variants of GRUs in the literature that have the ability to handle missing data as well~\cite{Che2018-nd}. Including these variants of GRU into DeepIDA-GRU would enable DeepIDA-GRU to handle missing data as well.
\subsection{Linear Mixed Models (LMM)}
\begin{figure}
         \centering
    \captionsetup{justification=centering}
    \includegraphics[width=0.6\columnwidth]{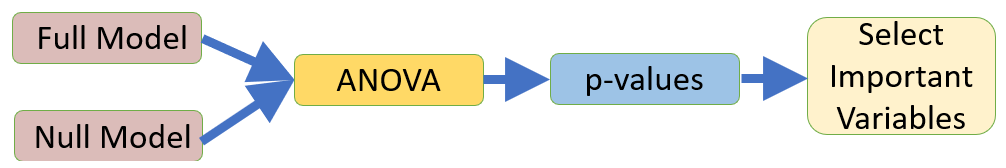}
         \caption{Pictorial representation of variable selection procedure using Linear Mixed Models (LMMs).}
  \label{fig:LMM_flowchart}
\end{figure}
LMMs are generalizations of linear models that allow both fixed and random effects to resolve the non-independence among data samples. They have been used by~\cite{lloyd2019} to perform differential abundance (DA) analysis on the IBD dataset. They can act as important tools to identify key variables in the longitudinal datasets. To determine if a given variable is important, using the LMM method, 
two models are constructed for that variable: (i) null model and (ii) full model. 
\begin{align*}
    \textbf{Null Model: } \text{variable\_value} &\sim  \text{time} + \cdots +\\
    & (1+\text{time}|\text{subject\_name}) + \cdots, \\
    \textbf{Full Model: } \text{variable\_value} &\sim {\text{class}} + \text{time} + \cdots + \\ &(1+\text{time}|\text{subject\_name}) + \cdots.
\end{align*}
Here, `variable\_value' refers to the value of the variable of subject `subject\_name', `class' represents the class to which the `subject\_name' belongs, and `time' represents the time or week the data was collected in. In the notation used above, the fixed effects (for example 
 time, class) are written outside parenthesis and the random effects (for example time$|$subject\_name) are written inside parenthesis. Moreover, the notation $x_1 | x_2$ allows us to model the random effects associated with the variable $x_1$ within the levels of the variable $x_2$ (or, in other words, this tells the model that the random slope/intercept of $x_1$ can vary for different realizations of $x_2$). Finally, the `$1$' in parenthesis indicates the intercept term. The variables corresponding to the fixed and random effects depend on the specific problem at hand. \par
 The null model assumes that `variable\_value' does not depend on the `class' of the subject, while the full model assumes that it does and treats the `class' as a fixed effect. The `time' variable allows us to associate `variable\_value' with its collection time, and makes it possible to use LMMs for variable selection with longitudinal views. The key idea here is to assume that the null model is true and calculate the p-value. In particular, we input both the models into ANOVA and depending on whether the p-value is greater or less than $0.05$, we determine if the null hypothesis is true (or if we should keep or discard the variable). This process is represented in Figure~\ref{fig:LMM_flowchart}.
 Furthermore, the variables can be ranked by arranging them in increasing order of their p-values. 
 It is important to note that this LMM based variable selection takes the classes of the variables into consideration and therefore emphasizes on variables separating the classes. However, it handles each variable separately and does not consider the dependencies between variables in a view, and between variables across views. 
 This could sometimes lead to a suboptimal variable selection because some variables may only be significant in the context of other variables. 
 
 \subsubsection{Functional Principal Component Analysis (FPCA)}
\begin{figure*}\captionsetup{justification=centering}
     \centering
     \begin{subfigure}{0.32\textwidth}
         \centering
\includegraphics[width=\textwidth]{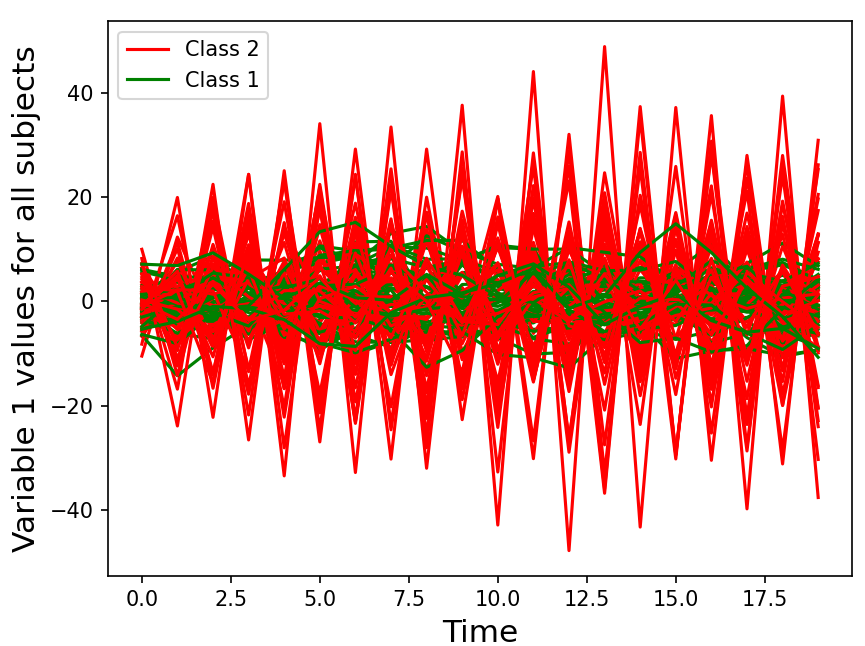}
         \caption{}
         \label{fig:FPC1}
     \end{subfigure}
     \hspace{-2mm}
     \begin{subfigure}{0.32\textwidth}
         \centering
         \includegraphics[width=\textwidth]{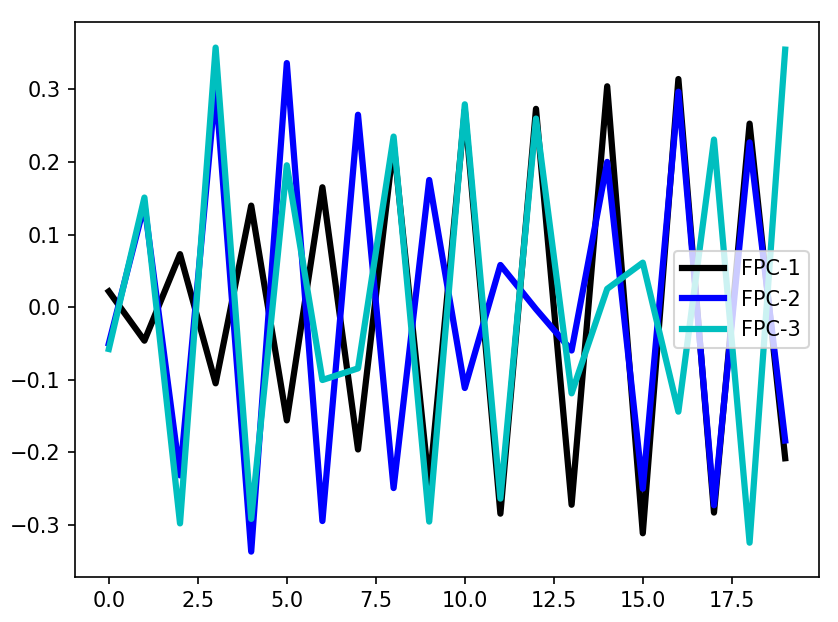}
         \caption{}
         \label{fig:FPC2}
     \end{subfigure}
     \hspace{-2mm}
     \begin{subfigure}{0.25\textwidth}
         \centering
         \includegraphics[width=\textwidth]{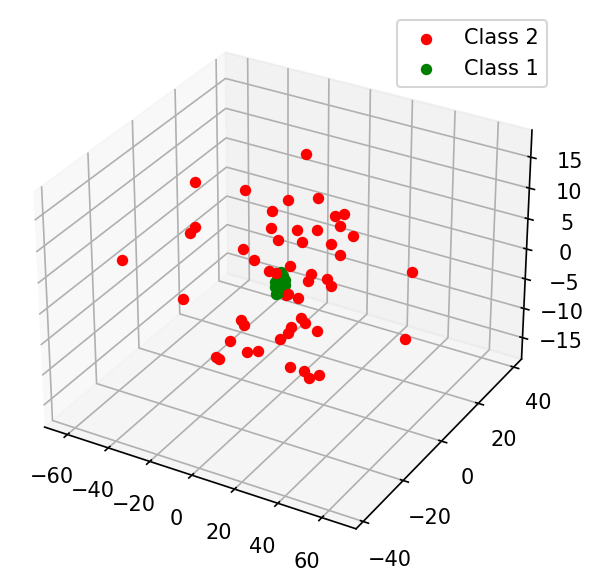}
         \caption{}
         \label{fig:FPC3}
     \end{subfigure}
        \caption{Demonstration of FPCA: (a) Multivariate time series for one of the variables for $N=100$ subjects generated using ARMA parameters that vary between the two classes, (b) First $x\!=\!3$ FPCs of that variable for the $N = 100$ time series and (c) $x\!=\!3$ dimensional FPC scores for the $N\!=\!100$ subjects.}
        \label{fig:FPCA}
\end{figure*}

FPCA is a dimensionality reduction technique similar to PCA that can be used with functional or time-series data. In this work, FPCA is used to convert longitudinal data $\widetilde{\mathbf{X}}_d \in \mathbb{R}^{N \times \widetilde{p}_d \times t_d}$ into one-dimensional form $\widehat{\mathbf{X}}_d \in \mathbb{R}^{N \times (xp_d) \times 1}$, by calculating $x$-dimensional scores for each of the $p_d$ variables, where $x$ is the number of FPC components considered for each variable. Specifically, for any given variable $j$ of view $d$, $\widetilde{\mathbf{X}}_{d}^{:,j,:} \in \mathbb{R}^{N \times t_d}$ is the collection of uni-variate time-series of all the $N$ subjects for that variable $j$ (where $d \in [1:D]$ and $j \in [1:\widetilde{p}_d]$). FPCA first finds the top $x$ functional principle components (FPCs): $f_1(t), f_2(t), \dots, f_x(t) \in \mathbb{R}^{1 \times t_d}$ from the $N$ univariate time series in $\widetilde{\mathbf{X}}_d^{:,j,:}$. These $x$ FPCs represent the top $x$ principal modes in the $N$ univariate time series and are obtained using basis functions like B-splines, wavelets etc. As an example, from a simulated longitudinal dataset, we plot the univariate time series corresponding to one of the variables for all $N=100$ subjects in Figure~\ref{fig:FPC1}. In Figure~\ref{fig:FPC2}, we plot the $x=3$ FPCs of this variable. Note that in this simulated data, the time series for class $1$ is generated using an ARMA(1,1) process with the AR and MA parameters given by $0.5$ and $0.6$ respectively. Moreover, the time series for the second class is generated using an ARMA(1,1) process with AR and MA parameters given by $-0.9$ and $-0.6$, respectively. 

Each of the $N$ uni-variate time-series in $\widetilde{\mathbf{X}}_d^{:,j,:}$ is then projected on each of the $x$ FPCs to get an $x$-dimensional score for each subject $n \in [1:N]$, corresponding to this variable. For example, each of the $N=100$ time-series in  Figure~\ref{fig:FPC1} are projected on the corresponding $x=3$ FPCs in Figure~\ref{fig:FPC2} to obtain $x=3$ dimensional scores, which are plotted in Figure~\ref{fig:FPC3}. The score of each subject (for a given variable) is of dimension  $x$. The scores of all $\widetilde{p}_d$ variables are stacked together to get a $\widetilde{p}_d x$ dimensional vector for that subject. Thus with the FPCA method, we convert a longitudinal data $\widetilde{\mathbf{X}}_d \in \mathbb{R}^{N \times \widetilde{p}_d \times t_d}$ to a cross-sectional type of data $\widehat{\mathbf{X}}_d \in \mathbb{R}^{N \times (x p_d) \times 1}$. We conducted synthetic experiments to compare the benefits and drawbacks of utilizing FPCA for feature extraction to those of Euler curves (EC) based feature extraction. This comparison is further explored in the main paper.
\section{More on Results}
In this section, we provide a detailed description of how the preprocessing steps and the proposed pipeline were applied to the IBD dataset.
\subsection{Prefiltering and Normalization}
For preprocessing the multi-omics data, we utilize some of the established normalization techniques in the literature~\cite{Zhao2021_normalization, Maza2016-qo}. Let us look at the preprocessing steps for each of the three views of the IBD dataset. 
\subsubsection{Metagenomics and Metabolomics Preprocessing}
For the metagenomics view, we preprocessed the path abundances of $22113$ gene pathways of all the subjects, which was gathered over the period of several weeks. Additionally, for the metabolomics view, we have $103$ hilic-negative factors of all the {subjects} collected over multiple weeks. The preprocessing of the metagenomics view is performed as follows (i) keep variables which have less than $90\%$ zeros over all the collected samples, (ii) add a pseudo count of 1 to each data value (this ensures that all entries are non-zero and allows for taking logarithms in the next steps),
(iii) normalize using the “Trimmed Mean of M-values” method~\cite{Maza2016-qo}, (iv) log-transform the data, and (v) {plot the histogram of the variances and filter out variables (pathways) with low variance across all the collected samples (the cutoff variance was chosen as $2.5$).} The prefiltering of the metabolomics view is performed as follows
(i) keep metabolites with less than $5\%$ zeroes over all the collected samples, (ii) add a pseudo count of $1$ to each data value, and (iii) log-transform the data.
After the preprocessing, the number of variables that remain for the metagenomics and the metabolomics data is reduced to $2261$ and $93$, respectively. 

\begin{figure} 
     \centering \captionsetup{justification=centering}
     \begin{subfigure}[b]{0.65\columnwidth}
         \centering
         \includegraphics[width=\textwidth]{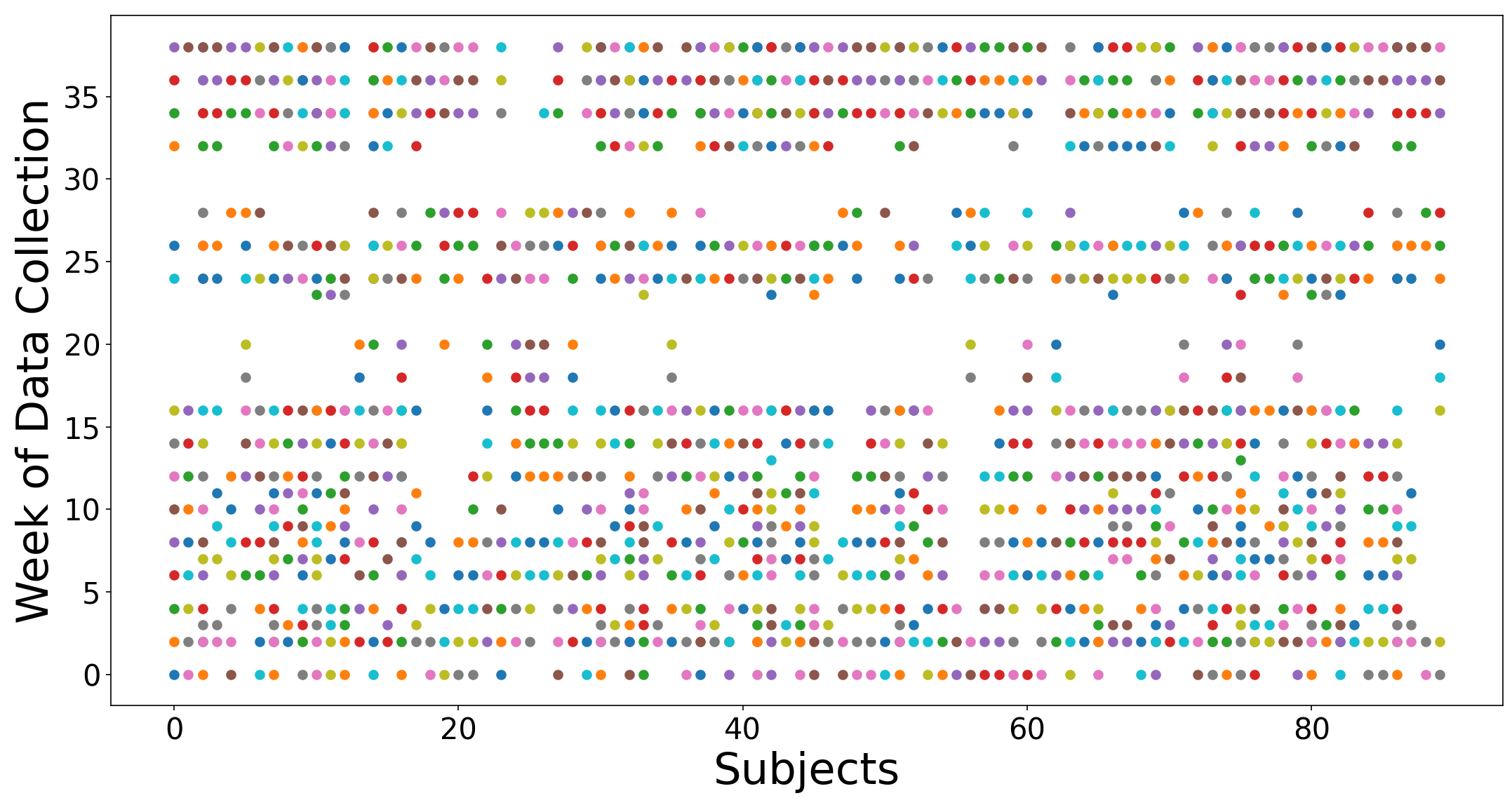}
         \caption{}
         \label{fig:MetagemomicsDataWeeks}
     \end{subfigure}
     \hfill
     \begin{subfigure}[b]{0.65\columnwidth}
         \centering
         \includegraphics[width=\textwidth]{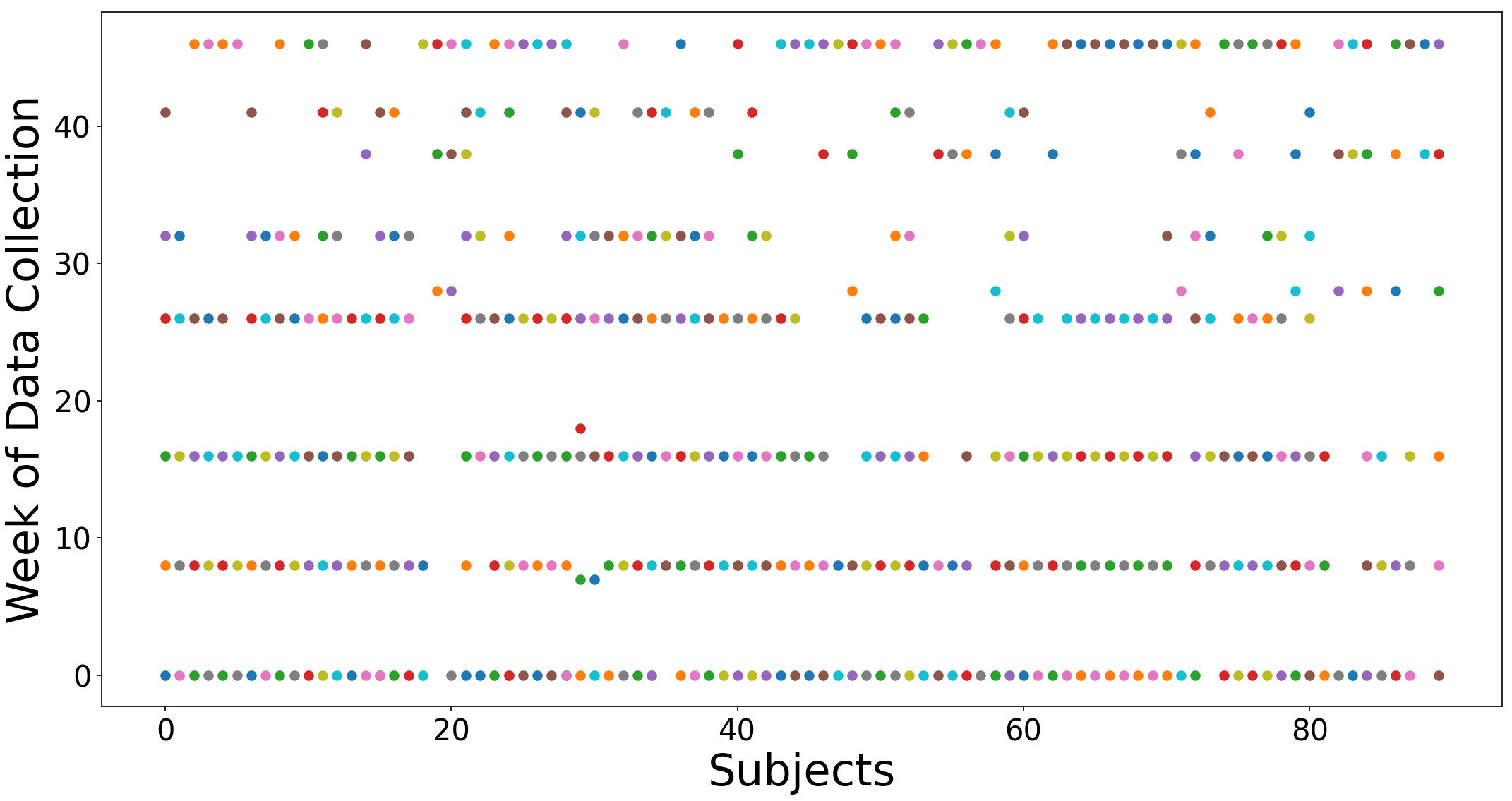}
         \caption{}
         \label{fig:MetabolomicsDataWeeks}
     \end{subfigure}
        \caption{Weeks of data collection of (a) metagenomics, (b) metabolomics data for all $90$ subjects: dots represent that the subjects data was collected during that week, a missing dot indicates missing data. }
        \label{fig:DataCollection}
\end{figure}

Note that after the initial preprocessing, we only retain the data from the $90$ subjects that are common across the three views.
A key challenge in analysing the metagenomics and metabolomics views arises from data missing from some weeks and consequently, unequal number of time-points among different patients. Both the available and the missing data for each of the $90$ subjects is pictorially indicated in Figure~\ref{fig:DataCollection}, where the dots represent that the subject's (x-axis) data was collected in that particular week (y-axis), whereas a missing dot indicates otherwise. In order to have equal number of time-points for each subject $n \in [1:90]$, we perform windowed averaging on the two datasets as follows. For each subject $n \in [1:90]$, we divide their data of the first $50$ weeks into $10$ groups $g \in [1:10]$. The data of group $g$ for subject $n$ is then obtained by averaging subject $n$'s data of weeks $[0:4]+5(g-1)$. Thus for each subject, we get a total of $t_d = 10$ time-points after grouping. Note that if in group $g$ (or the weeks $[0:4]+5(g-1)$), subject $n$ had no data collected, then that subject's data for group $g$ is made equal to their data of group $g-1$. Having equal number of time-points for all the subjects is specifically required by JPTA, but for consistency, we use the windowed dataset for the other methods as well. The preprocessed metagenomics and metabolomics data thus obtained, are denoted using $3$-dimensional real-valued tensors, $\mathbf{X_{\text{metaG, norm}}} \in \mathbb{R}^{90 \times 2261 \times 10}$ and $\mathbf{X_{\text{metaB, norm}}} \in \mathbb{R}^{90 \times 93 \times 10}$, respectively. Both these longitudinal datasets are then passed through the variable selection step. 

\subsubsection{Host Transcriptomics}
This view consists of the host transcriptomics counts of $55765$ gene probes. For each participant, all the samples of their host transcriptomics view are gathered within a single week. Therefore, we consider the host transcriptomics view as a cross-sectional view in this work, and the data for each individual is taken as the mean of all the samples collected from them. The preprocessing steps for this view are: (i) keep genes which have less than $5\%$ zeros across the collected samples, (ii) add a pseudo-count of $1$ to each data value, (iii) normalize using DESeq2-normalized counts (using the median of ratios method~\cite{Maza2013-bp}), 
(iv) log-transform the variables, and 
(v) plot the histogram of the variances and filter out genes which have low variance across the collected samples (the cutoff variance was chosen as $0.5$). After preprocessing, the number of variables are reduced to $9726$. The resulting data is denoted by $\mathbf{X_{\text{hostTx,norm}}} \in \mathbb{R}^{90 \times 9726 \times 1}$. This data is then passed through the variable selection step. 

\subsection{Variable Selection/Ranking} \label{sec:featureselection}
In this subsection, we discuss the application of the three variable selection/ranking methods: LMM, DGB and JPTA 
to the multi-omics IBD dataset. These methods identify the key variables which are likely to be the most important in separating the two classes and/or associating the views. Given the fundamental differences in the three methods' selection approach, we obtain different sets of selected variables, with some overlap between these sets. The utilization of these methods on the IBD dataset is discussed as follows.

\noindent\underline{\textbf{Linear Mixed Models}}: 
Similar to~\cite{IBDMBD}, we construct the following two models for each variable separately:
\begin{align} \label{eq:LMM}
    &\textbf{Null Model: } \text{value} \sim \text{week} + \text{age} + (1+\text{week}|\text{subname}) \nonumber\\
    &\qquad \qquad \qquad \quad + (1|\text{sitename}), \nonumber\\
    &\textbf{Full Model: } \text{value} \sim \text{IBD\_status} + \text{week} + \text{age} \nonumber \\
    &\qquad \qquad \qquad \quad+ (1+\text{week}|\text{subname}) + (1|\text{sitename}),
\end{align}
where `value' is the scalar value of the corresponding variable (gene, pathway or metabolite), `week' is the week number during which that value of the variable was collected, `age' is the age of the subject which this value corresponds to, `subname' is the name of the subject, `sitename' is the name of the site (out of $5$ different hospitals) where this subject's data was collected and `IBD\_status' indicates if the corresponding subject is healthy or has IBD. Note that here `week', `IBD\_status' and `age' are fixed effects, whereas `sitename' and `subname' are random effect grouping variables. The two models are passed through ANOVA and the $p-$values are computed. For each view, the variables are arranged in increasing order of their p-values and the top $1000, 200$ and $50$ variables are selected from the host-transcriptomics, metagenomics and metabolomics datasets, respectively. In this way, we obtain the three datasets of size $\mathbf{\widetilde{X}_{\text{hostTx,norm}}} \in \mathbb{R}^{90 \times 1000 \times 1}$, $\mathbf{\widetilde{X}_{\text{metaB,norm}}} \in \mathbb{R}^{90 \times 50 \times 10}$ and $\mathbf{\widetilde{X}_{\text{metaG,norm}}} \in \mathbb{R}^{90 \times 200 \times 10}$. Note that since LMM uses class labels (IBD\_status) during its selection procedure, it is only used with the training data to perform variable selection.

\noindent\underline{\textbf{DGB}}:
DGB uses dense neural networks for the cross-sectional host-transcriptomics view and GRUs for the longitudinal metabolomics and metagenomics views to non-linearly transform the data before integration. 
 The network architecture used for the three views are as follows: (i) a $3$ layer fully-connected neural network (with $200,100,20$ neurons for the three layers respectively) for the host transcriptomics view and (ii) a GRU (with $2$ layers and $256$ dimensional hidden state) for both the metagenomics and the metabolomics datasets. The outputs of these networks are integrated using IDA. The bootstrapping is performed for $120$ epochs, that is, $M=120$ bootstrapped sets are used, and the top $1000, 50$ and $200$ variables are retained from the host transcriptomics, the metabolomics and the metagenomics views, respectively.  
 Using DGB, we obtain the three datasets of size $\mathbf{\widetilde{X}_{\text{hostTx,norm}}} \in \mathbb{R}^{90 \times 1000 \times 1}$, $\mathbf{\widetilde{X}_{\text{metaB,norm}}} \in \mathbb{R}^{90 \times 50 \times 10}$ and $\mathbf{\widetilde{X}_{\text{metaG,norm}}} \in \mathbb{R}^{90 \times 200 \times 10}$. Note that since DGB also uses class labels (IBD\_status) during variable selection, it only uses the data from the subjects in the training split.
 
\noindent\underline{\textbf{JPTA}}:
Since JPTA can only be used with a pair of longitudinal views, this method can only perform variable selection on metagenomics and metabolomics views. In this work, we use JPTA to extract the top $200$ and $50$ variables from the metagenomics and metabolomics views, respectively. Moreover, since JPTA cannot be used with the host transcriptomics view, all the transcriptomics variables obtained after the preprocessing step, are retained in the JPTA variable selection scenario. Note that we did not associate ranking/scores with variables when using the JPTA method. In other words, the top $50$ and $200$ variables selected from the metabolomics and metagenomics views, respectively, are all assumed to be ranked equally amongst themselves. Using JPTA, we obtain the three datasets of size $\mathbf{\widetilde{X}_{\text{hostTx,norm}}} \in \mathbb{R}^{90 \times 9726 \times 1}$, $\mathbf{\widetilde{X}_{\text{metaB,norm}}} \in \mathbb{R}^{90 \times 50 \times 10}$ and $\mathbf{\widetilde{X}_{\text{metaG,norm}}} \in \mathbb{R}^{90 \times 200 \times 10}$. (Note that the code utilized for implementing JPTA is sourced from their paper~\cite{Zhang2018}.)

\subsection{Feature Extraction}
DeepIDA-GRU can accept a mix of longitudinal and cross-sectional views. This makes the one-dimensional feature extraction step optional. However, feature extraction methods like EC and FPCA are (i) good at capturing important characteristics from the longitudinal datasets~\cite{zavala2021,BERRENDERO20112619}, (ii) easy to compute and computationally efficient, and (iii) more resilient against issues like overfitting, loss divergence, diminishing gradients etc. Therefore, for the multi-omics dataset under consideration, we have explored both EC and FPCA feature extraction methods and compared their results against the direct DeepIDA-GRU approach. These three approaches are discussed as follows.

\begin{itemize}
    \item \textbf{Method 1: DeepIDA-GRU with no Feature Extraction: } In this case, there is no feature extraction. Therefore, we have $\mathbf{\widehat{X}_{\text{hostTx,norm}}}=\mathbf{\widetilde{X}_{\text{hostTx,norm}}}, \mathbf{\widehat{X}_{\text{metaB,norm}}} = \mathbf{\widetilde{X}_{\text{metaB,norm}}}$ and $\mathbf{\widehat{X}_{\text{metaG,norm}}}=\mathbf{\widetilde{X}_{\text{metaG,norm}}}$. The cross-sectional host transcriptomics dataset 
$\mathbf{\widehat{X}_{\text{hostTx,norm}}} \in \mathbb{R}^{90 \times \widetilde{p}_d \times 1}$ (where $\widetilde{p}_d = 1000$ in the case of LMM and DGB variable selection; and $\widetilde{p}_d = 9726$ in the case of JPTA variable selection) is inputted to a fully-connected neural network (with $3$ layers and $200,100,20$ neurons in these three layers). The metagenomics dataset $\mathbf{\widehat{X}_{\text{metaG, norm}}} \in \mathbb{R}^{90 \times 200 \times 10}$ and the metabolomics dataset $\mathbf{\widehat{X}_{\text{metaB, norm}}} \in \mathbb{R}^{90 \times 50 \times 10}$ are each inputted to their respective GRUs (both with $2$ layers and $50$ dimensional hidden unit).
\item \textbf{Method 2: DeepIDA-GRU with EC for Metagenomics View and Mean for Metabolomics View: } In this case, EC (with $100$ threshold values) is used to convert the metagenomics dataset $\mathbf{\widetilde{X}_{\text{metaG, norm}}} \in \mathbb{R}^{90 \times 200 \times 10}$ into $\mathbf{\widehat{X}_{\text{metaG, norm}}} \in \mathbb{R}^{90 \times 100 \times 1}$. The metabolomics dataset $\mathbf{\widetilde{X}_{\text{metaB, norm}}} \in \mathbb{R}^{90 \times 50 \times 10}$ is converted into $\mathbf{\widehat{X}_{\text{metaG, norm}}} \in \mathbb{R}^{90 \times 50 \times 1}$ by taking mean across the time dimension. (The reason for not choosing EC for the metabolomics dataset was the lack of differentiation between the two classes when visualizing the EC curves for this dataset and we achieved a better performance by simply taking the mean.) The cross-sectional host transcriptomics dataset 
$\mathbf{\widetilde{X}_{\text{hostTx,norm}}}$ remains unchanged, that is, $\mathbf{\widehat{X}_{\text{hostTx,norm}}} = \mathbf{\widetilde{X}_{\text{hostTx,norm}}}$. Similar to Method 2, the resulting tensors $\mathbf{\widehat{X}_{\text{hostTx,norm}}}$, $\mathbf{\widehat{X}_{\text{metaB, norm}}}$ and $\mathbf{\widehat{X}_{\text{metaG, norm}}}$ are each fed into their respective $3$-layered dense neural networks with structures $[200,100,20]$, $[20,100,20]$ and $[50,100,20]$, respectively.
    \item \textbf{Method 3: DeepIDA-GRU with FPCA for both the Metabolomics and Metagenomics Views: } In this case, FPCA (with $x=3$ FPCs for each variable) is used to convert the metabolomics dataset $\mathbf{\widetilde{X}_{\text{metaB, norm}}} \in \mathbb{R}^{90 \times 50 \times 10}$ into $\mathbf{\widehat{X}_{\text{metaB, norm}}} \in \mathbb{R}^{90 \times 150 \times 1}$ and the metagenomics dataset $\mathbf{\widetilde{X}_{\text{metaG, norm}}} \in \mathbb{R}^{90 \times 200 \times 10}$ into $\mathbf{\widehat{X}_{\text{metaG, norm}}} \in \mathbb{R}^{90 \times 600 \times 1}$. The cross-sectional host transcriptomics dataset 
$\mathbf{\widetilde{X}_{\text{hostTx,norm}}}$ remains unchanged, that is, $\mathbf{\widehat{X}_{\text{hostTx,norm}}} = \mathbf{\widetilde{X}_{\text{hostTx,norm}}} $. The three resulting tensors $\mathbf{\widehat{X}_{\text{hostTx,norm}}}$, $\mathbf{\widehat{X}_{\text{metaB, norm}}}$ and $\mathbf{\widehat{X}_{\text{metaG, norm}}}$ are each fed into their respective $3$-layered dense neural networks with structures $[200,100,20]$, $[20,100,20]$ and $[50,100,20]$, respectively.
\end{itemize}

We train and test the $9$ possible combinations of the $3$ variable selection methods: LMM, JPTA and DGB and the $3$ feature extraction methods: Method 1, Method 2 and Method 3 for the classification task of classifying each subject into healthy (non-IBD) or diseased (IBD) class. For this performance analysis, we used $N$-fold cross-validation, which is discussed in the main manuscript.

\subsection{Analysis of Significant Variables}

The top variables selected by LMM, JPTA and DGB are analysed in this subsection. As discussed in the main manuscript, in $N$-fold cross validation, the model is trained on $N-1$ subjects (where $N=90$) and tested on the remaining $1$ subject.  This procedure is repeated $N$ times (hence $N$-folds). 
Since LMM and DGB leverage information about the output labels 
while selecting variables, both these methods only use data from the $N-1$ subjects belonging to the training split of each fold.
In $N=90$ folds, since there are $90$ different train-test splits, LMM and DGB methods are repeated $90$ times (once for each fold), resulting in $N$-fold variable selection. 
Each variable can have different scores (p-value in the case of LMM and eff\_prop value in the case of DGB) in these N folds. To associate an overall ranking with each variable, we need to combine these $N$ scores of each variable.
    
    For LMM, an overall rank/score is associated to every variable as follows. First, for each variable, we compute the Fisher combined p-value from the $N=90$ different p-values. The Fisher combined p-values are then normalized across all the variables using the following steps: (i) 
    we add the minimum non-zero Fisher combined p-value to every variable's Fisher combined p-value (this ensures that all values are greater than $0$), (ii) we take negative log of the resulting values (from step (i)) and divide by the maximum. Note that more than $25$ genes from the host-transcriptomics view had Fisher combined p-value of $0$ and following the above normalization steps, all these genes were all assigned the same normalized score (Figure~\ref{fig:LMM_hostTx}). 
Similarly, in Figure~\ref{fig:LMM_metabolomics}, the top $4$ variables and in Figure~\ref{fig:LMM_metagenomics}, the top $3$ variables had Fisher combined p-value equal to $0$, and hence these variables have the same normalized score. 
For the DGB method, the $90$ eff\_prop values of each variable are averaged to get the combined score for that variable. Moreover, JPTA is only performed once on the entire dataset (because it is an unsupervised variable selection method) and we do not assign rank/score to variables while using this method. Hence, all the variables in the figures~\ref{fig:JPTA_metabolomics}~and~\ref{fig:JPTA_metagenomics} have the same scores.  

To highlight the key variables selected by the three methods, in Figure~\ref{fig:top25}, the top $25$ variables selected for each view using the three methods are listed along with their respective combined scores. 
In Figure~\ref{fig:top5}, the values of the top $5$ variables selected by the three methods 
are summarized using violin plots for the cross-sectional host-transcriptomics view and 
mean time-series curves for the longitudinal metabolomics and metagenomics views. These plots show statistically how different the top $5$ variables are between the two classes.
\begin{figure*}[p]
\captionsetup{justification=centering}
     \centering
     \begin{subfigure}{0.94\textwidth}
         \centering
         \includegraphics[width=\textwidth]{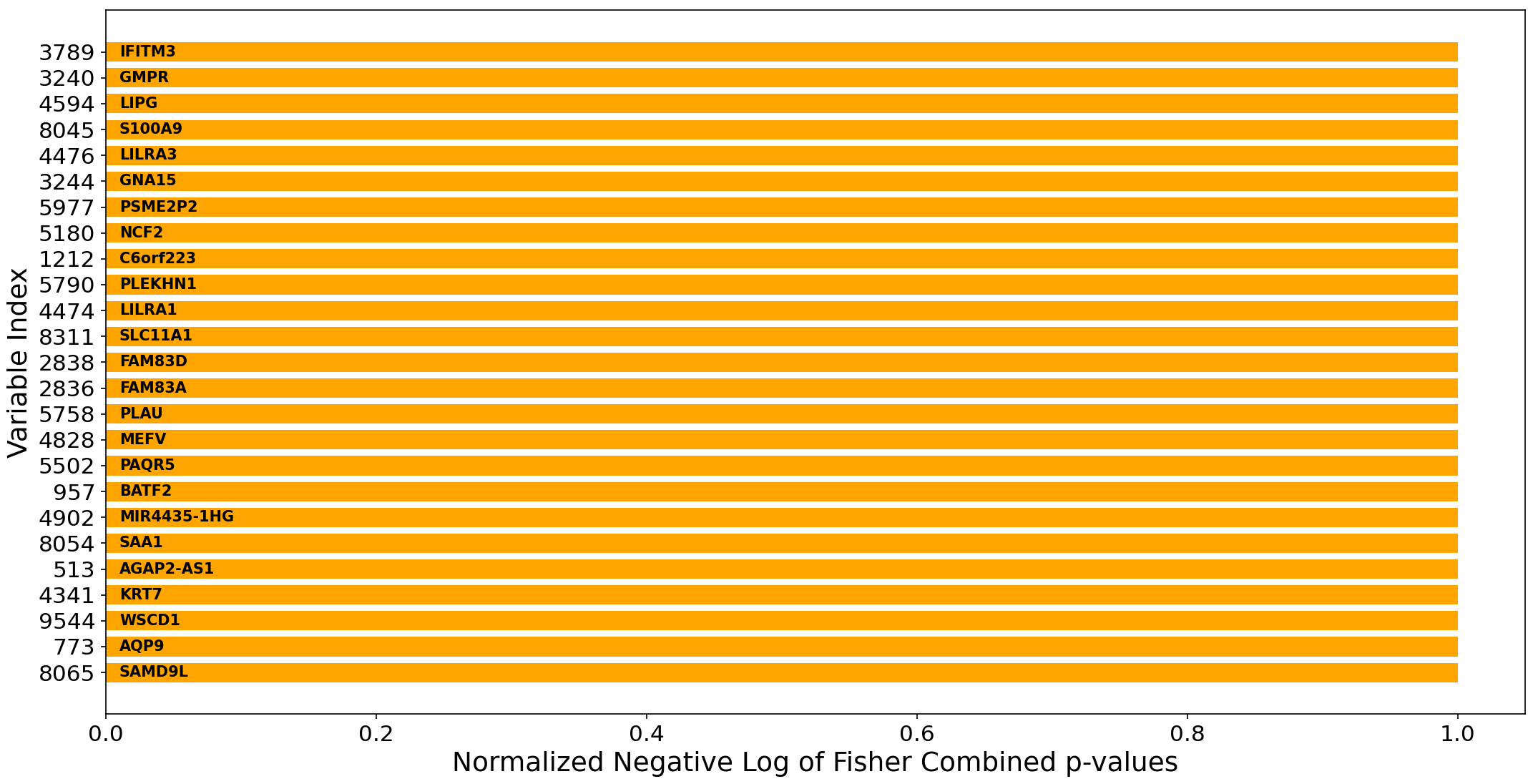}
         \caption{Top $25$ genes selected by LMM from host transcriptomics view. (Since the top $25$ host-transcriptomics genes had Fisher combined p-value equal to $0$, these variables are ranked the same.) }
         \label{fig:LMM_hostTx}
     \end{subfigure}
     \vfill
     \begin{subfigure}{0.94\textwidth}
         \centering
         \includegraphics[width=\textwidth]{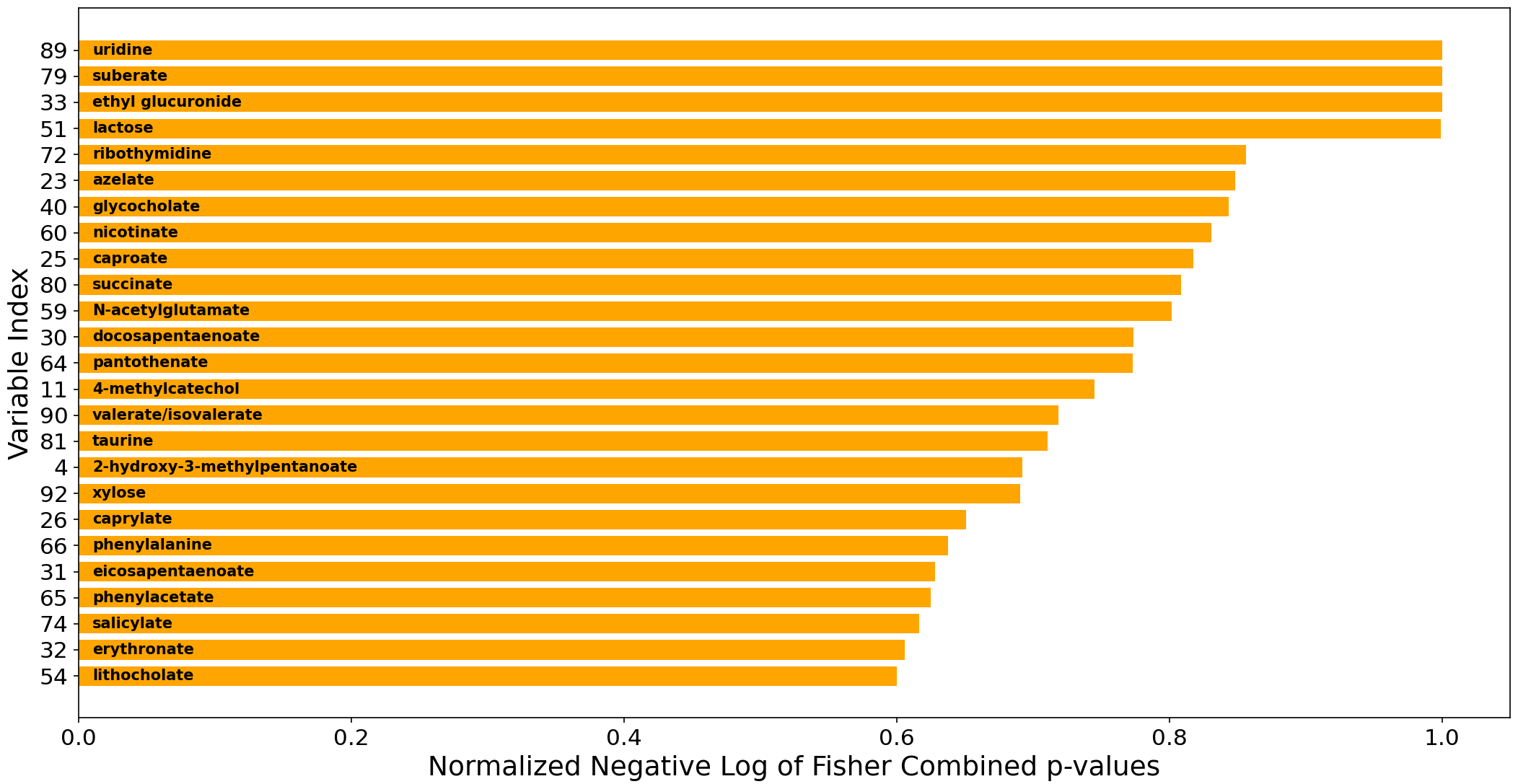}
         \caption{Top $25$ metabolites selected by LMM from metabolomics view.}
         \label{fig:LMM_metabolomics}
         \end{subfigure}
\end{figure*}
\begin{figure*}[p] \ContinuedFloat
\captionsetup{justification=centering}
     \begin{subfigure}{0.98\textwidth}
         \centering
         \includegraphics[width=\textwidth]{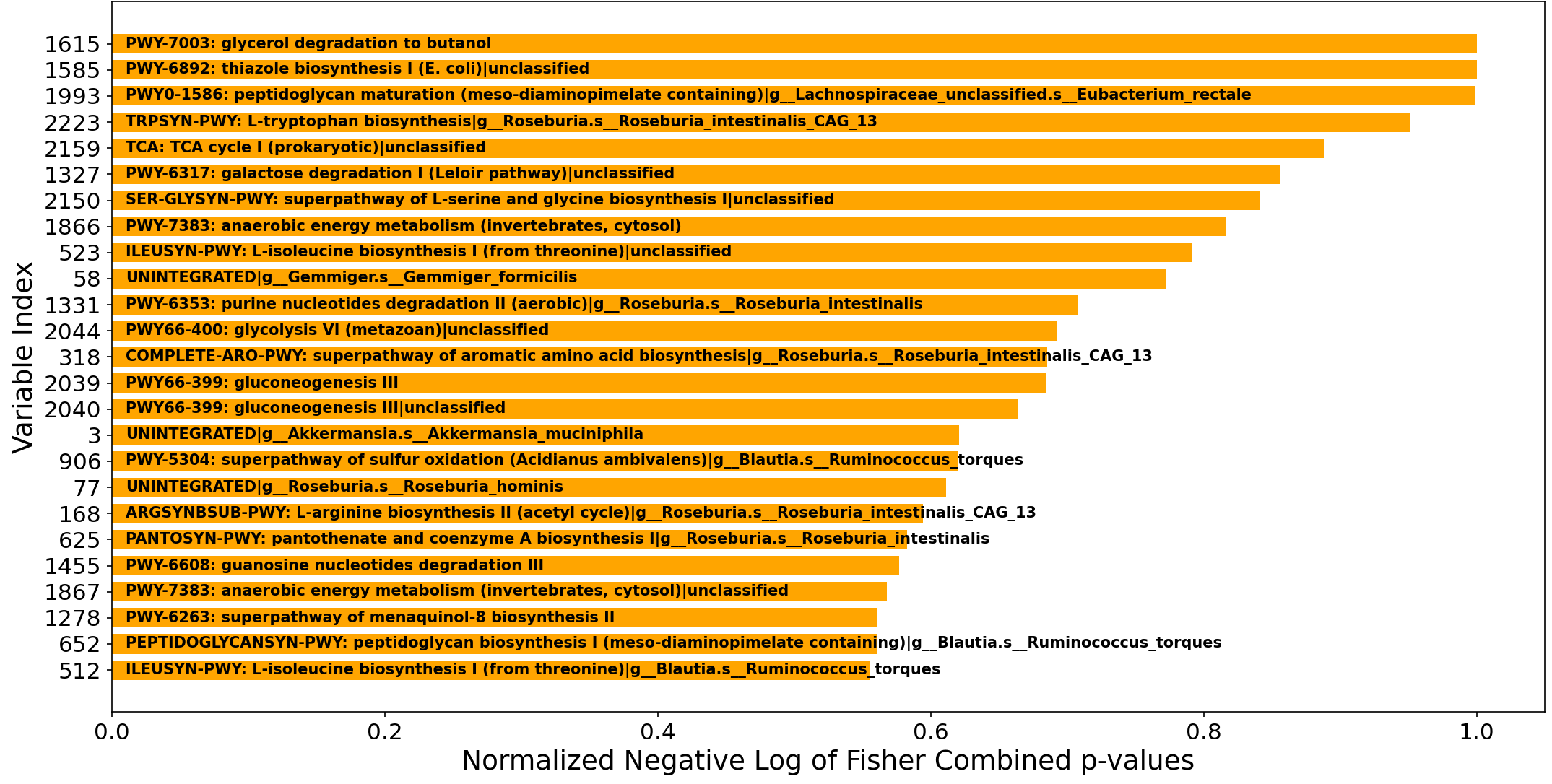}
         \caption{Top $25$ gene pathways selected by LMM from metagenomics view.}
         \label{fig:LMM_metagenomics}
         \end{subfigure}
         \vfill
     \begin{subfigure}{0.98\textwidth}
         \centering
         \includegraphics[width=\textwidth]{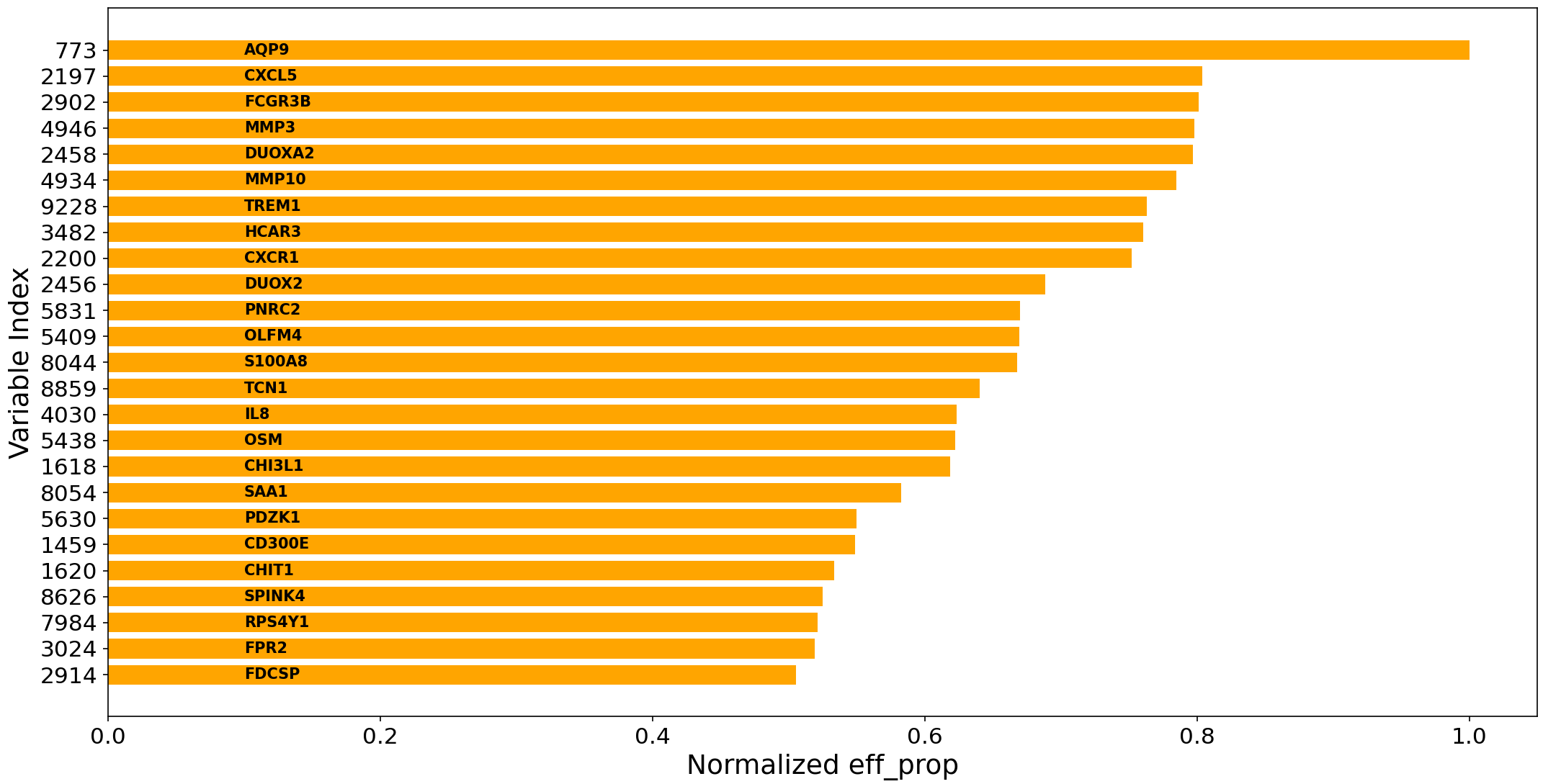}
         \caption{Top $25$ genes selected by DGB from host transcriptomics view.}
         \label{fig:DGB_HostTx}
     \end{subfigure}
    \end{figure*}
     \begin{figure*}[p] \ContinuedFloat  \captionsetup{justification=centering}
     \begin{subfigure}{0.98\textwidth}
         \centering
         \includegraphics[width=\textwidth]{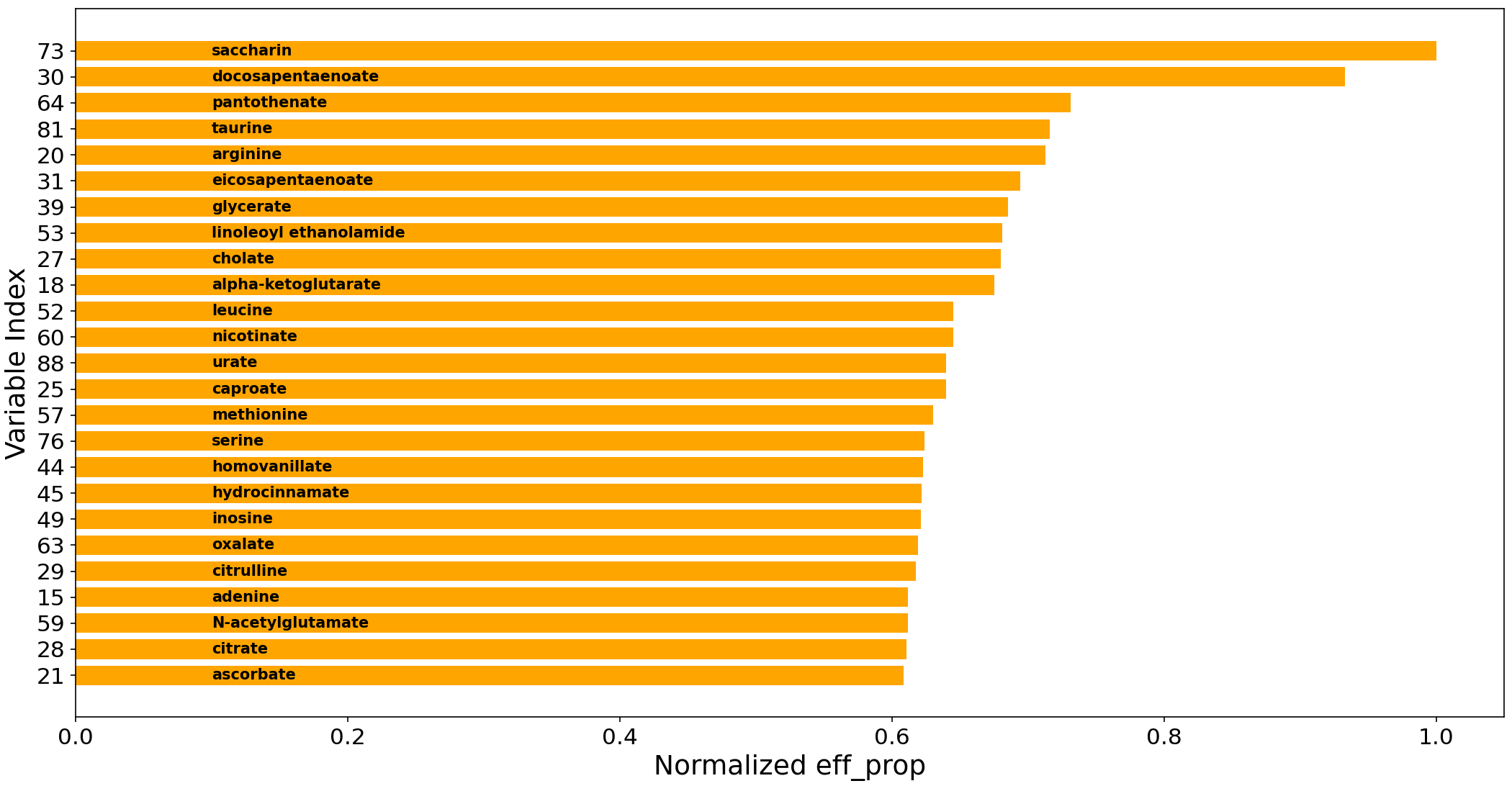}
         \caption{Top $25$ metabolites selected by DGB from metabolomics view.}
         \label{fig:DGB_metabolomics}
         \end{subfigure}
     \begin{subfigure}{0.98\textwidth}
         \centering
         \includegraphics[width=\textwidth]{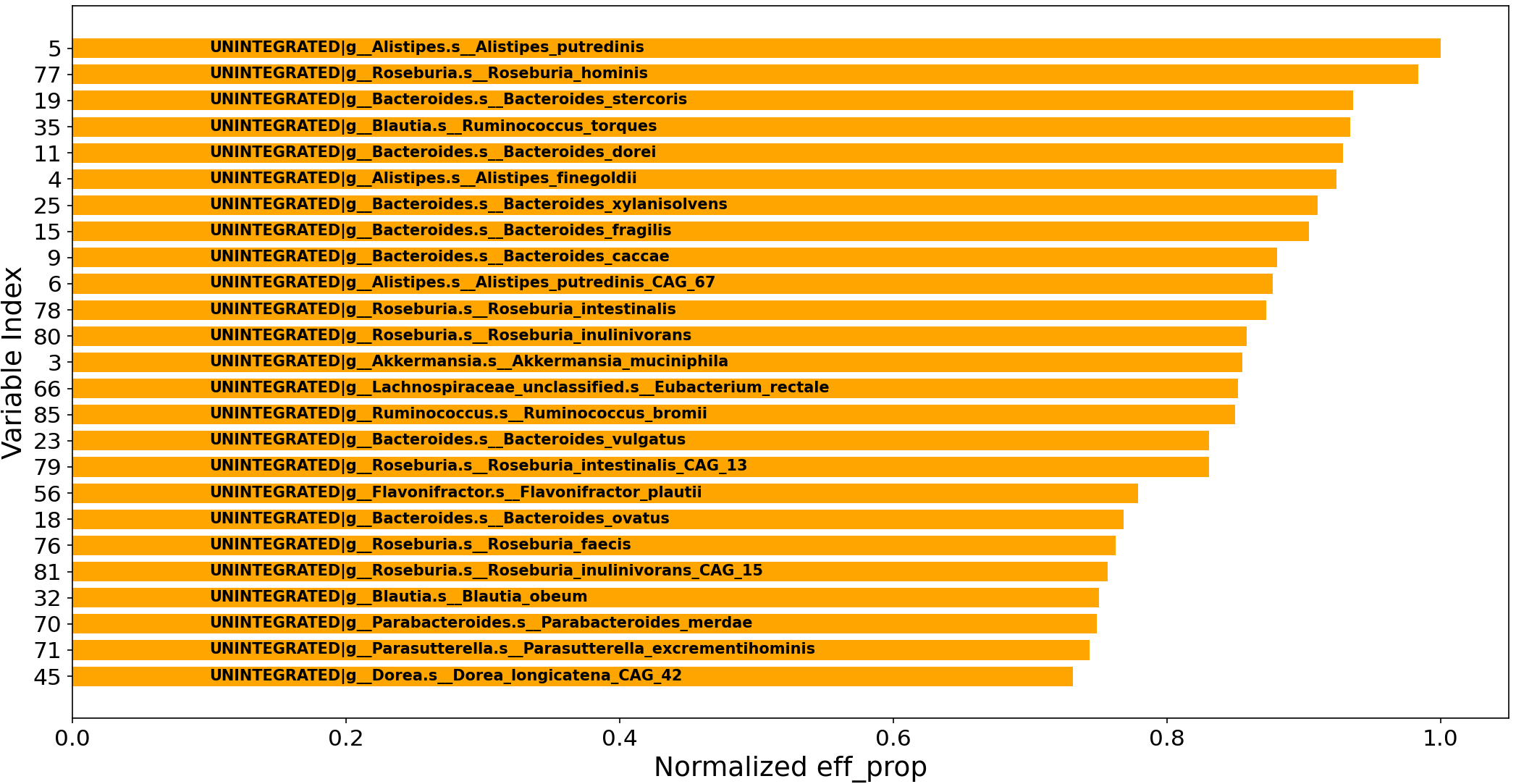}
         \caption{Top $25$ gene pathways selected by DGB from metagenomics view.}
         \label{fig:DGB_metagenomics}
         \end{subfigure}
\end{figure*}
\begin{figure*}[p] \ContinuedFloat
\captionsetup{justification=centering}
     \centering
     \begin{subfigure}{0.93\textwidth}
         \centering
         \includegraphics[width=\textwidth]{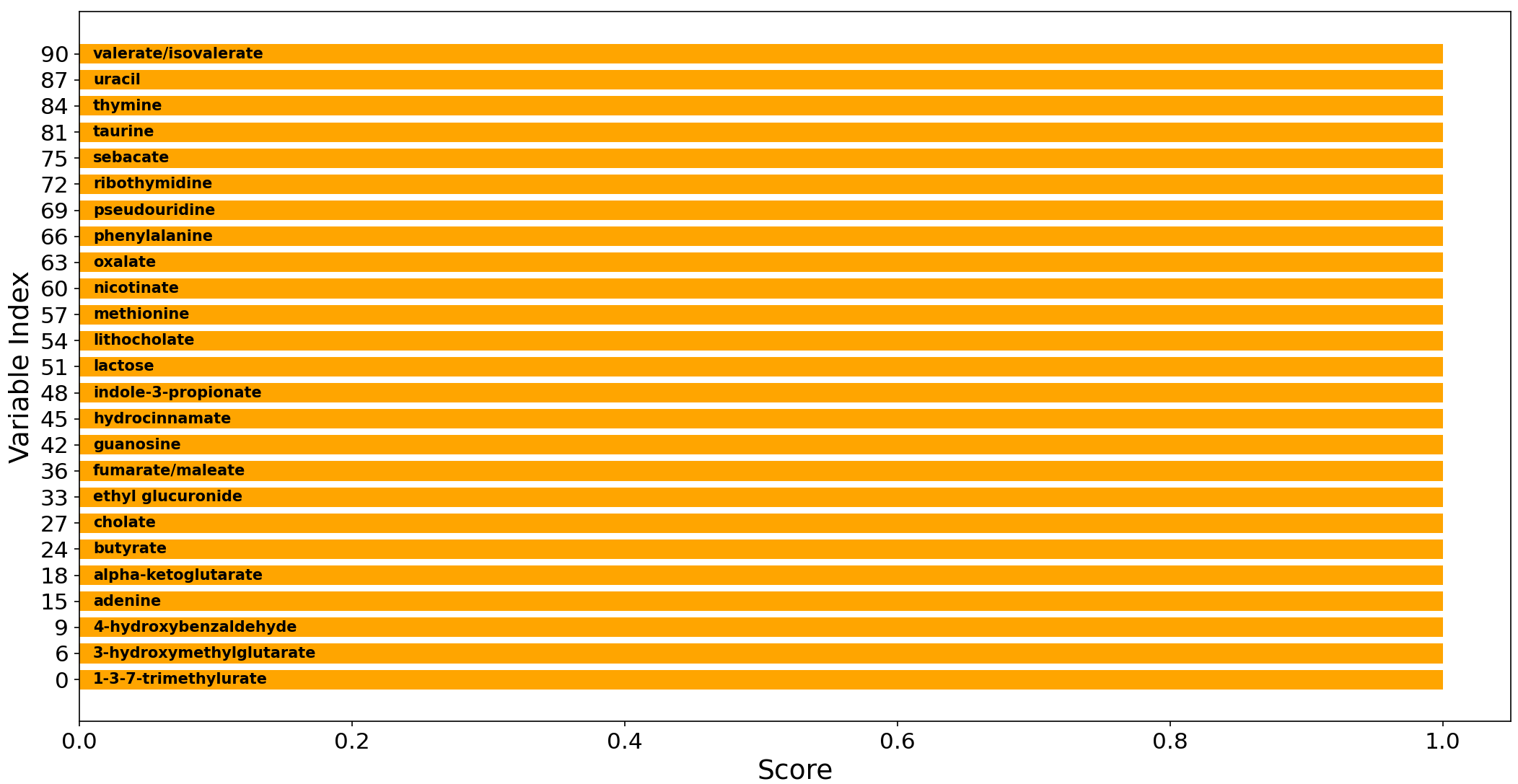}
         \caption{Top $25$ metabolites selected by JPTA from metabolomics view.}
         \label{fig:JPTA_metabolomics}
     \end{subfigure}
     \vfill
     \begin{subfigure}{0.93\textwidth}
         \centering
         \includegraphics[width=\textwidth]{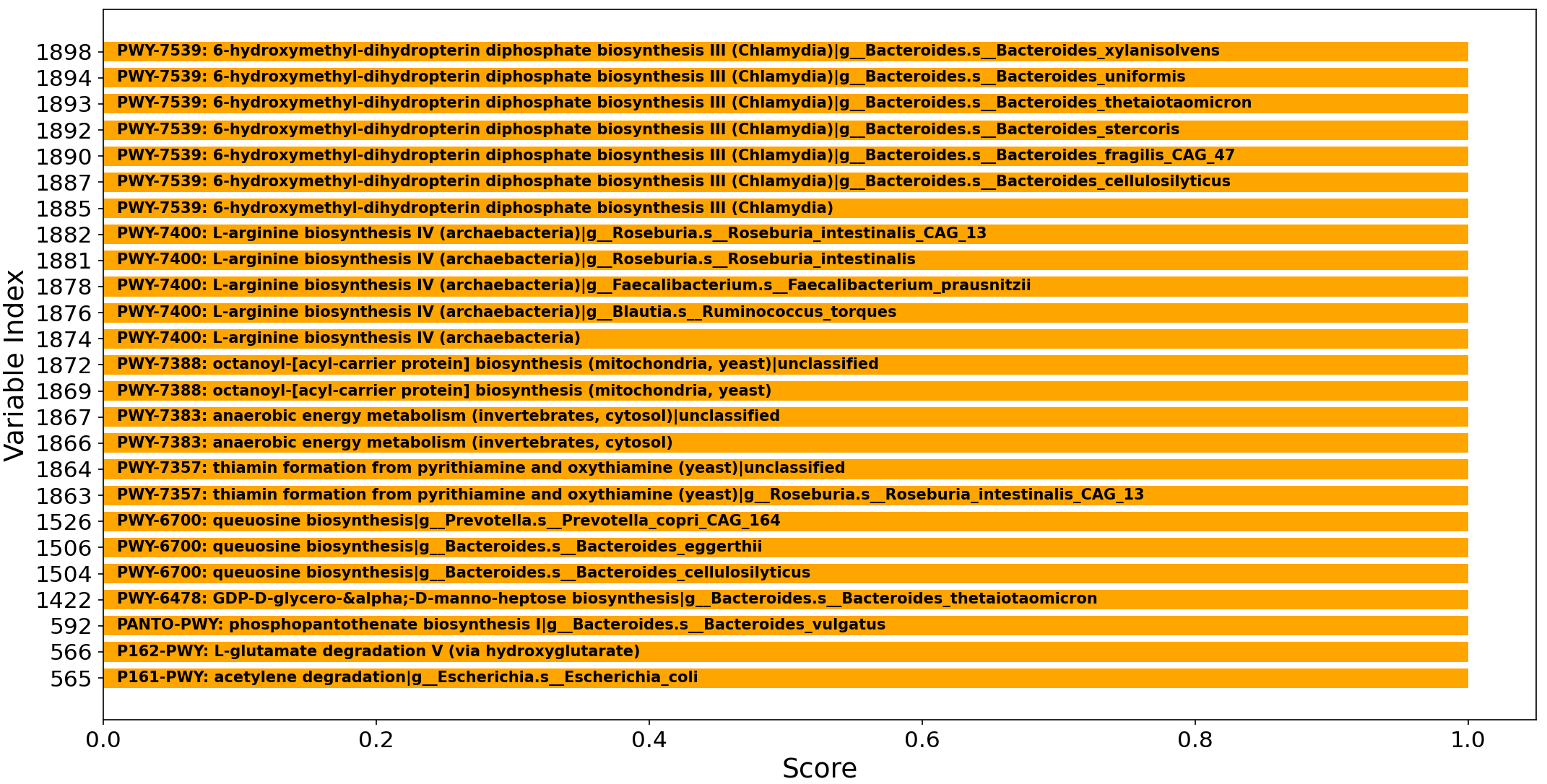}
         \caption{Top $25$ gene pathways selected by JPTA from metagenomics view.}
         \label{fig:JPTA_metagenomics}
         \end{subfigure}
         \caption{Top 25 variables (and their respective scores) selected from the host transcriptomics, metabolomics and metagenomics datasets using LMM, DGB and JPTA.}
         \label{fig:top25}
\end{figure*}
\begin{figure*}[p]
\captionsetup{justification=centering}
     \centering
     \begin{subfigure}{0.86\textwidth}
         \centering
        \includegraphics[width=0.92\textwidth]{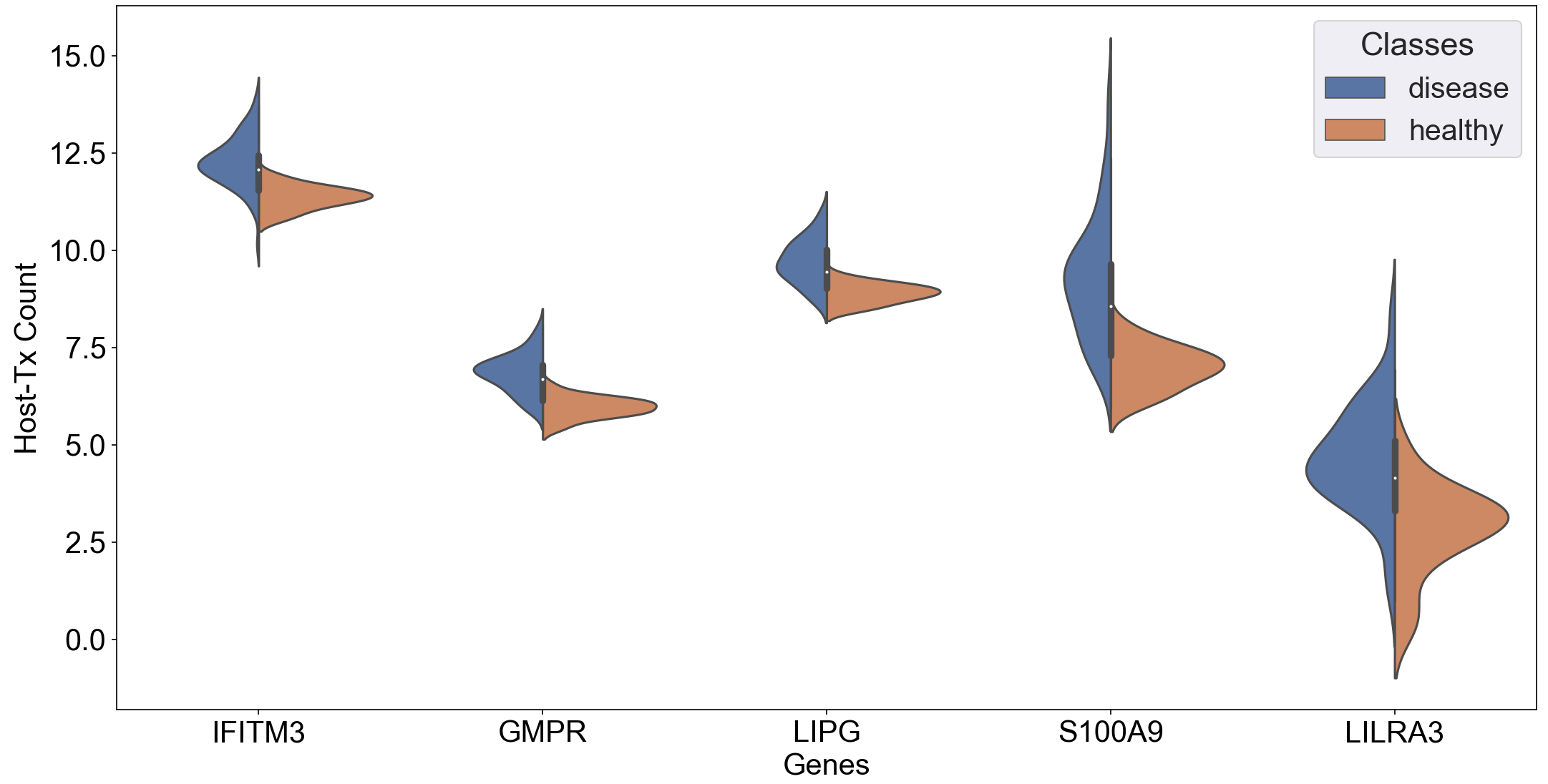}
         \caption{Violin plots for the top $5$ genes selected by LMM from the host transcriptomics view showing the difference in  the distribution of the expression of these genes between the two classes. }
  \label{fig:violin_LMM}
     \end{subfigure}
     \vfill
     \begin{subfigure}{0.86\textwidth}
         \centering
         \includegraphics[width=\textwidth]{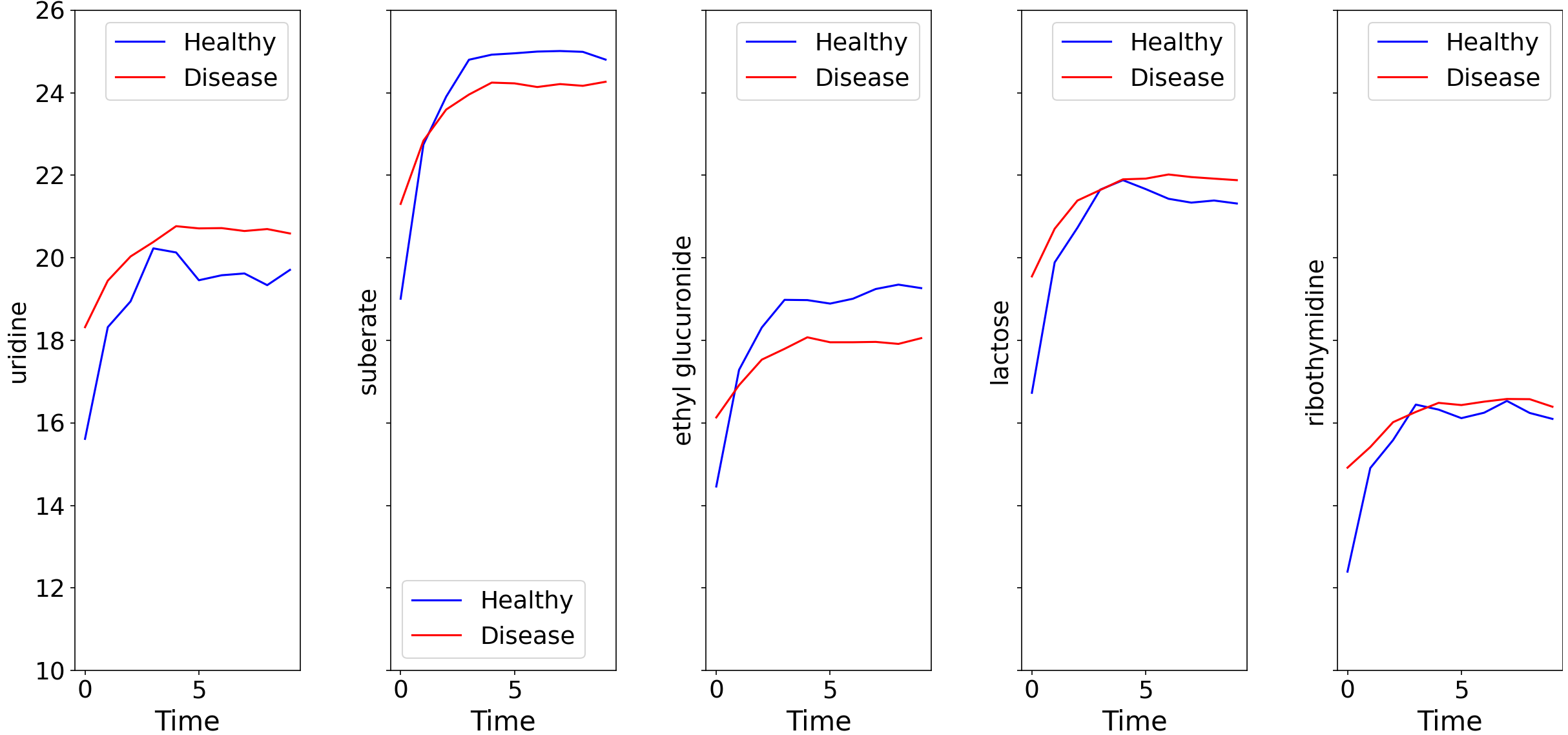}
         \caption{Mean time series plots for the top $5$ metabolites selected by LMM from the metabolomics view, showing the difference in the mean of the time series of these metabolites between the two classes.}
         \label{fig:MeanPlot_LMM_metabolomics}
         \end{subfigure}
\end{figure*}
\begin{figure*}[p] \ContinuedFloat
\captionsetup{justification=centering}
\centering
     \begin{subfigure}{0.86\textwidth}
         \centering
\includegraphics[width=\textwidth]{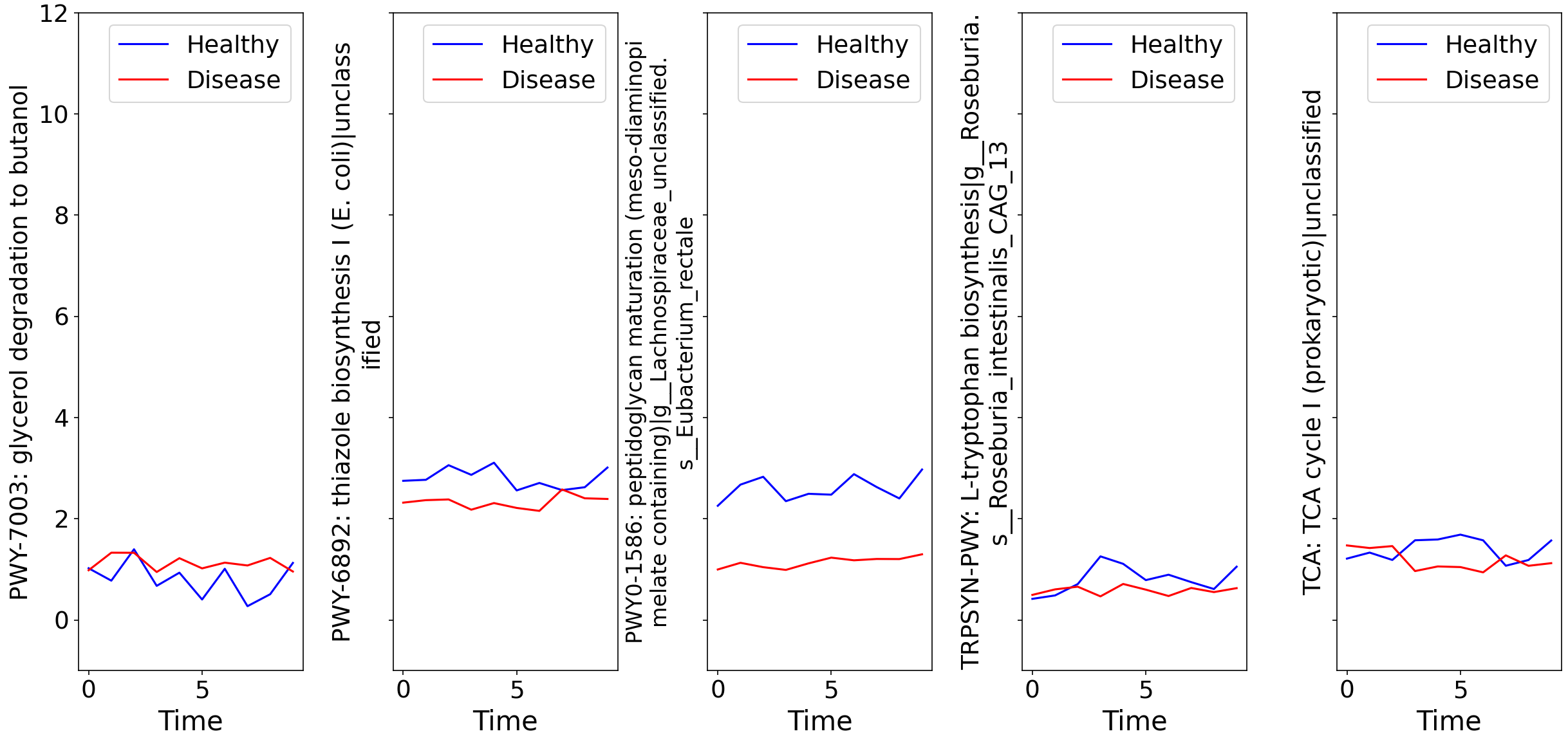}
         \caption{Mean time series plots for the top $5$ gene pathways selected by LMM from the metagenomics view, showing the difference in the mean of the time series of these pathways between the two classes.}
         \label{fig:MeanPlot_LMM_Metagenomics}
         \end{subfigure}
     \begin{subfigure}{0.6\textwidth}
         \centering
        \includegraphics[width=0.92\textwidth]{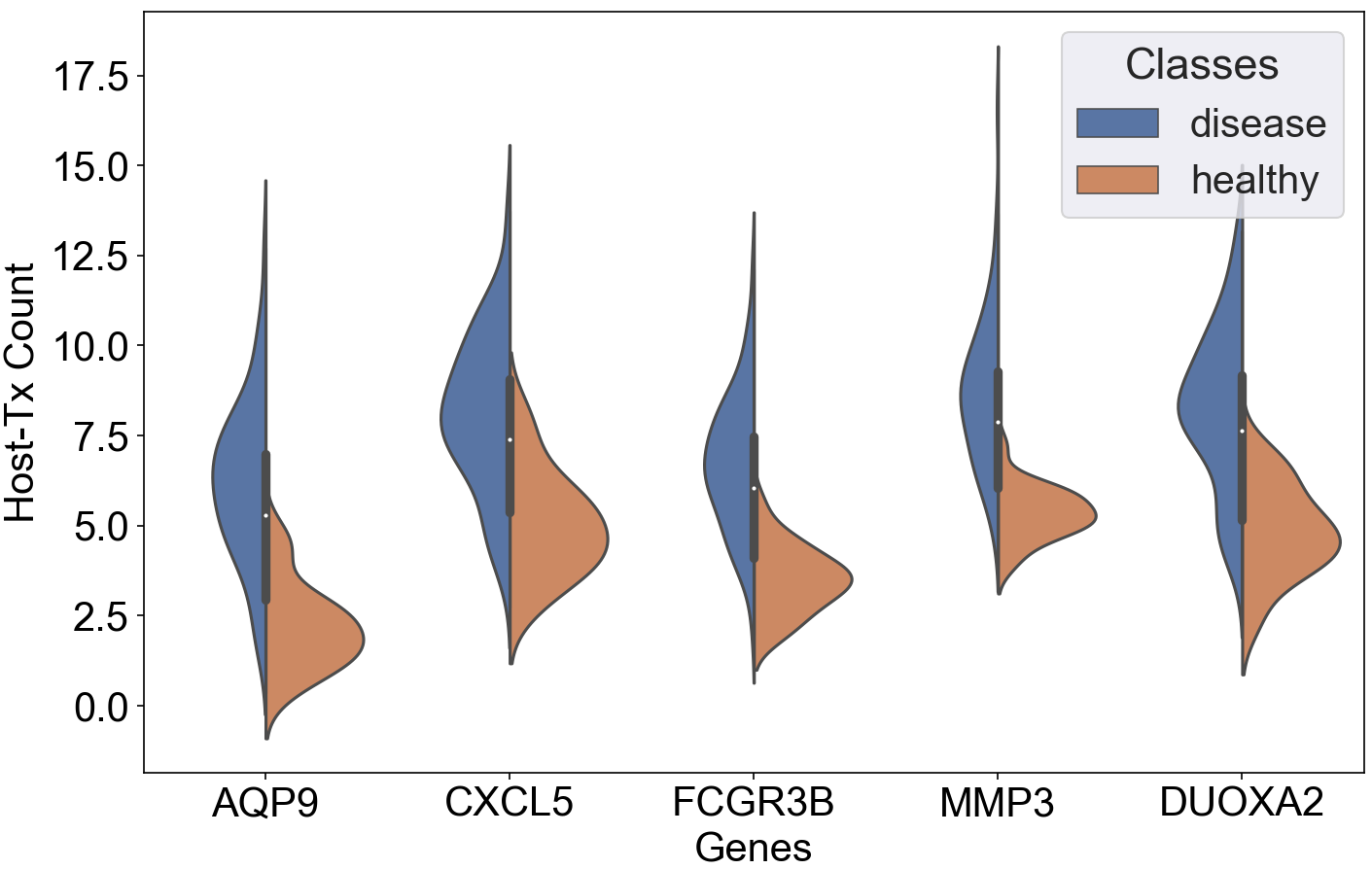}
         \caption{Violin plots for the top $5$ genes selected by DGB from the host transcriptomics view showing the difference in the distribution of the gene expressions between the two classes.}
       \label{fig:violin_DGB}
     \end{subfigure}
\end{figure*}  
\begin{figure*} [p]\ContinuedFloat
\captionsetup{justification=centering}
\centering
     \begin{subfigure}{0.86\textwidth}
         \centering
         \includegraphics[width=\textwidth]{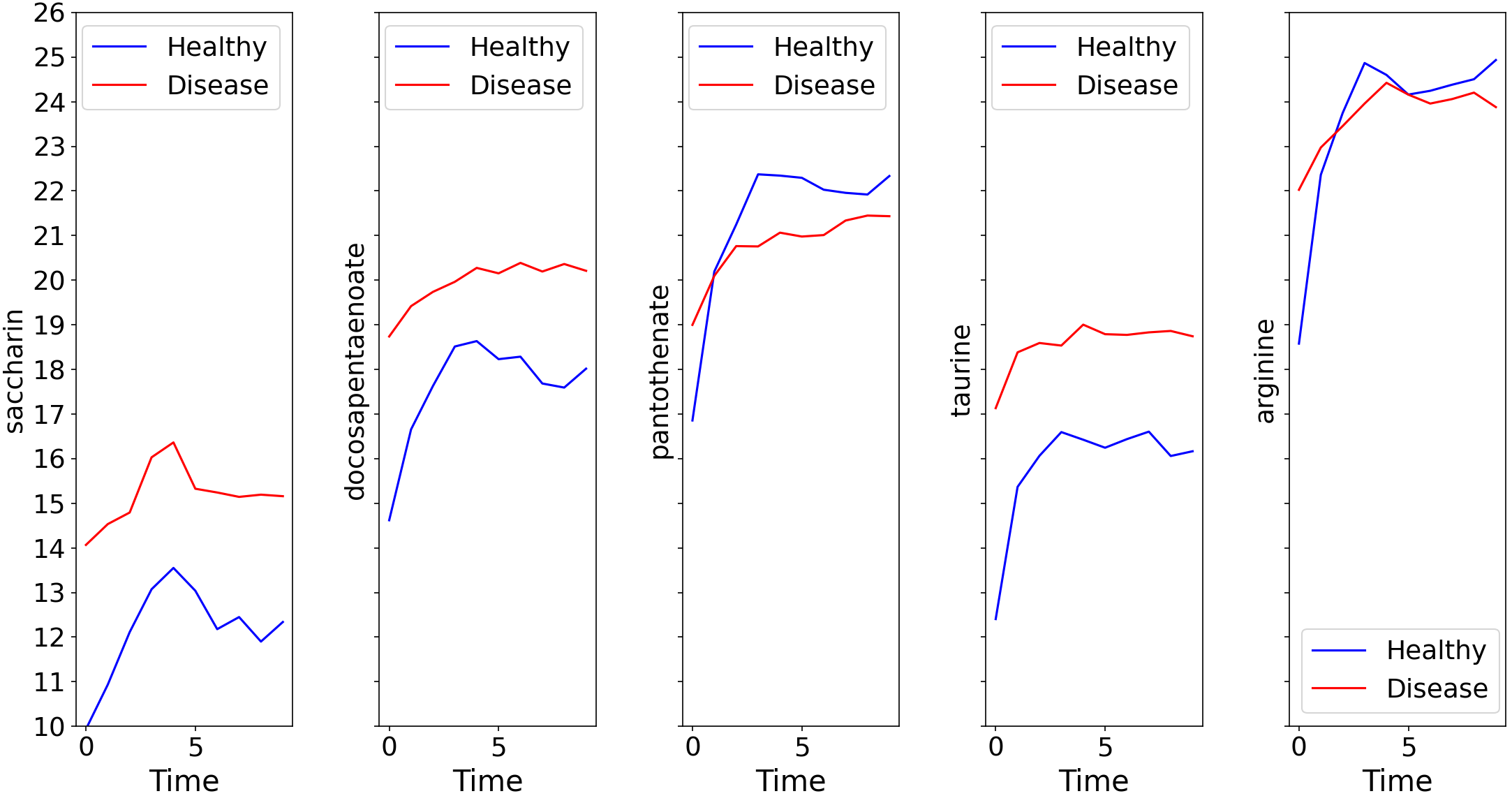}
         \caption{Mean time series plots for the top $5$ metabolites selected by DGB from the metabolomics view, showing the difference in the mean of the time series of these metabolites between the two classes.}
         \label{fig:MeanPlot_DGB_metabolomics}
         \end{subfigure}

     \begin{subfigure}{0.86\textwidth}
         \centering
         \includegraphics[width=\textwidth]{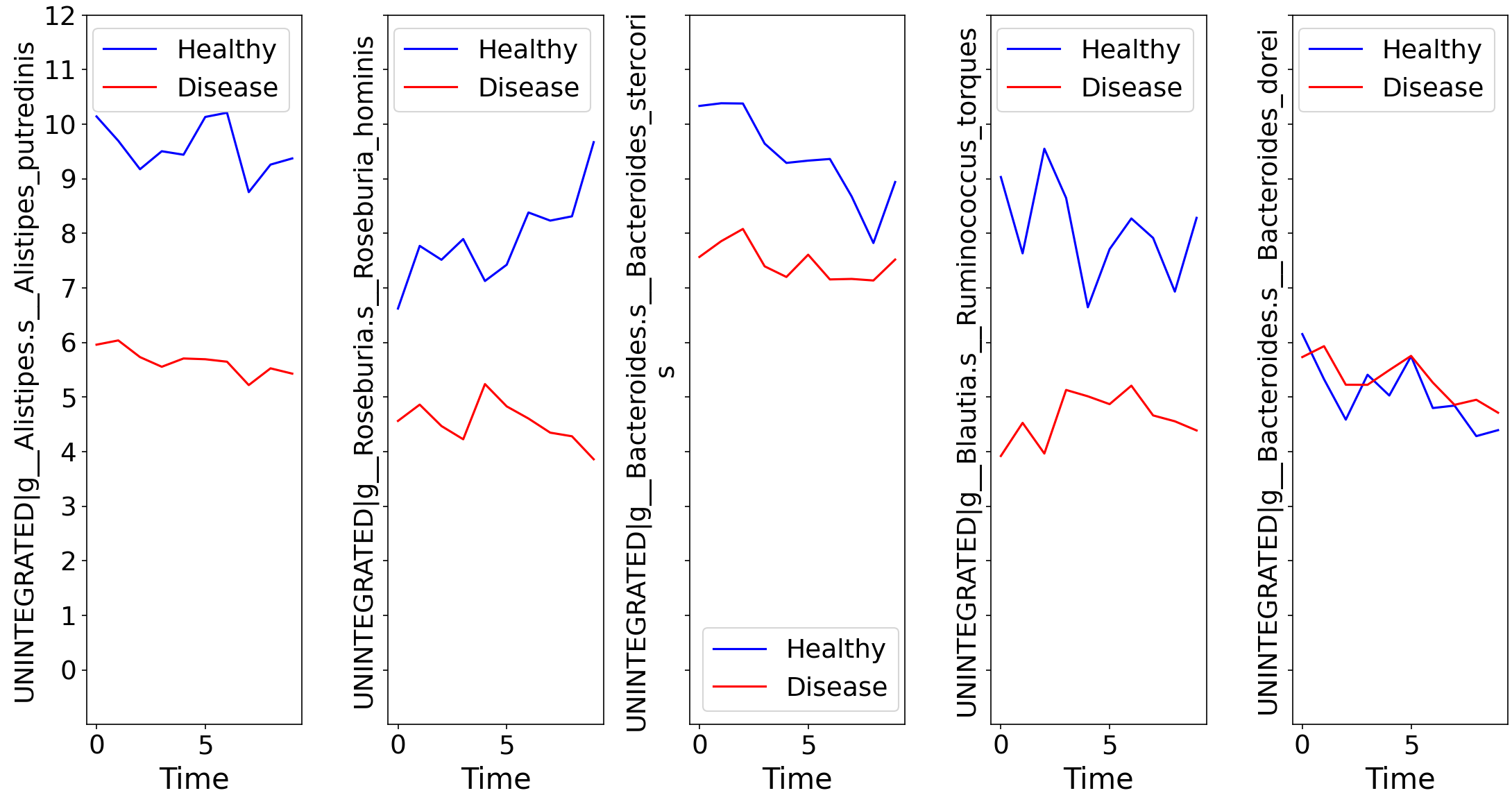}
         \caption{Mean time series plots for the top $5$ gene pathways selected by DGB from the metagenomics view, showing the difference in the mean of the time series of these pathways between the two classes.}
         \label{fig:MeanPlot_DGB_Metagenomics}
         \end{subfigure}
\end{figure*}
\begin{figure*} [p]\ContinuedFloat
\captionsetup{justification=centering}
\centering
    \begin{subfigure}{0.86\textwidth}
         \centering
        \includegraphics[width=\textwidth]{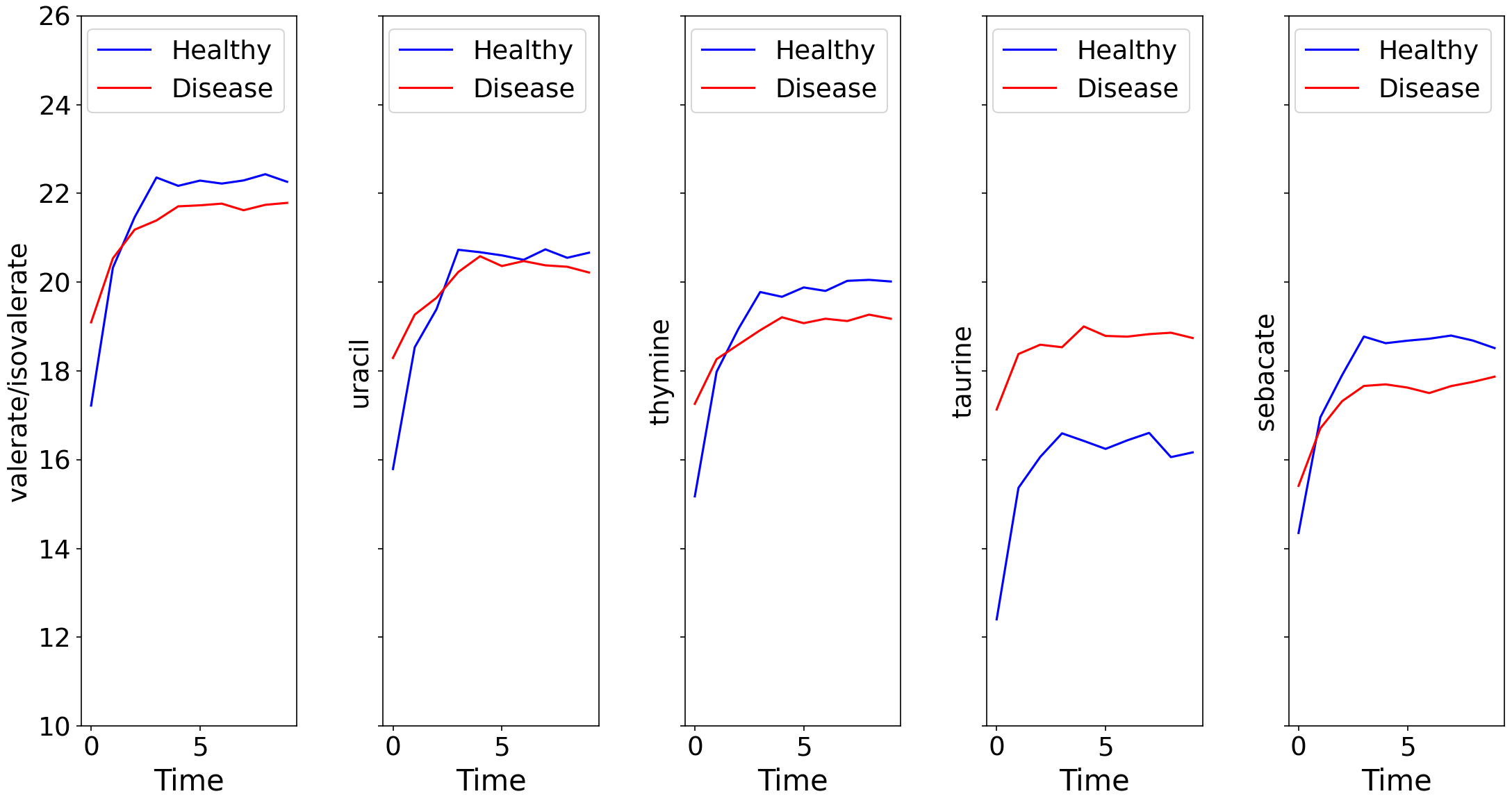}
         \caption{Mean time series plots for the top $5$ metabolites selected by JPTA from the metabolomics view, showing the difference in the mean of the time series of these metabolites between the two classes.}
  \label{fig:MeanPlot_JPTA_metabolomics}
     \end{subfigure}
\begin{subfigure}{0.86\textwidth}
         \centering
         \includegraphics[width=1\textwidth]{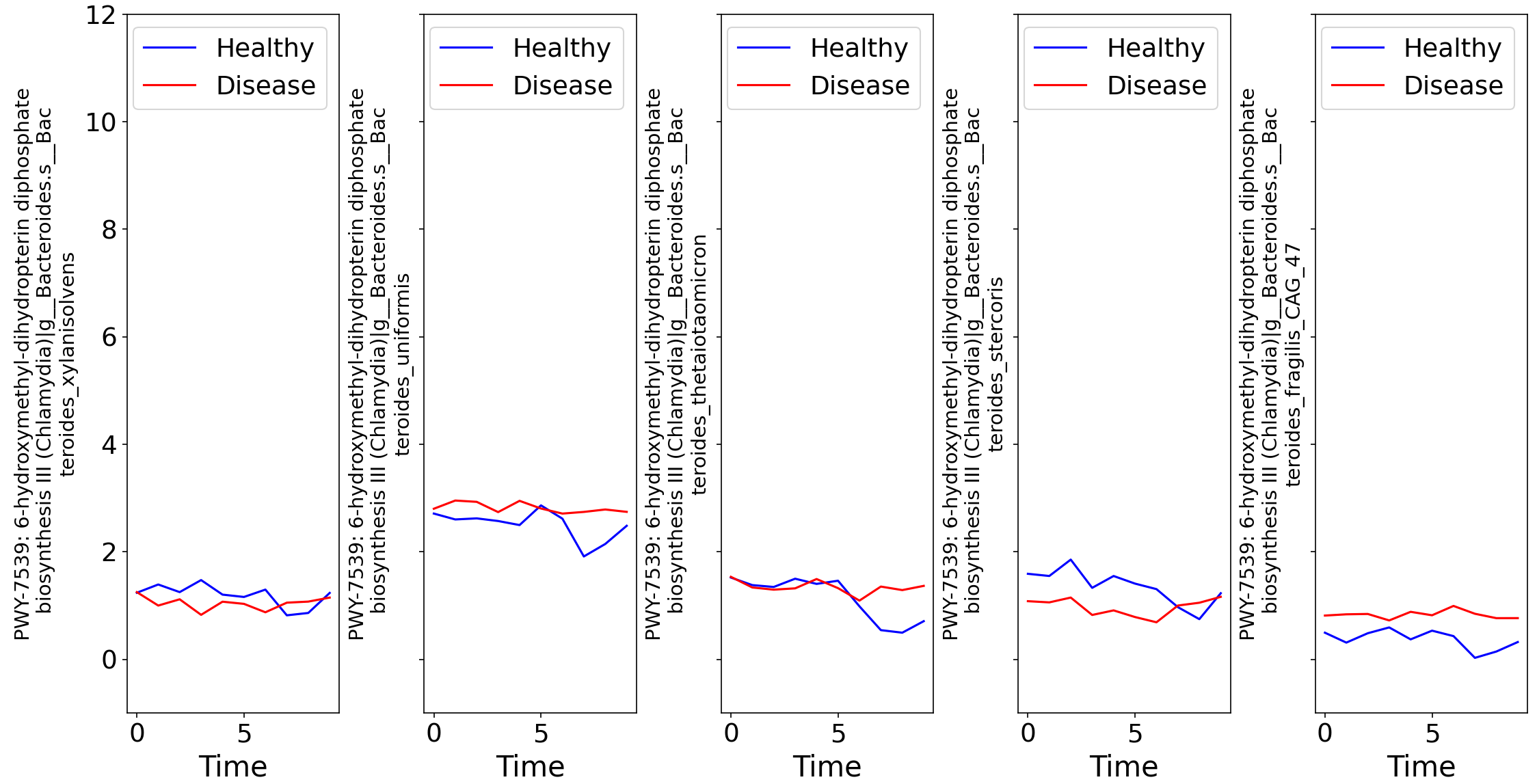}
         \caption{Mean time series plots for the top $5$ gene pathways selected by JPTA from the metagenomics view, showing the difference in the mean of the time series of these pathways between the two classes.}
         \label{fig:MeanPlot_JPTA_metagenomics}
         \end{subfigure}
         \caption{Violin and mean plots for the top $5$ variables selected by LMM, JPTA and DGB from the multi-omics data.}
         \label{fig:top5}
\end{figure*}

\begingroup
\clearpage
 \bibliography{references.bib}
 \bibliographystyle{IEEEtran}
\endgroup